\documentclass[twocolumn]{article}
\usepackage{lipsum} 
\usepackage{titling}
\usepackage{abstract}
\usepackage{caption}
\usepackage{graphicx} 
\usepackage{changepage}
\usepackage{ragged2e}
\usepackage{amsmath} 
\usepackage{float} 
\usepackage{bold-extra} 
\usepackage{listings}
\usepackage[backend=biber, style=numeric, defernumbers=true]{biblatex}
\usepackage{algorithm}
\usepackage{algpseudocode}
\usepackage{hyperref} 
\usepackage{tcolorbox} 

\usepackage{footmisc} 
\DeclareSourcemap{
  \maps[datatype=bibtex]{
    \map{
      \step[fieldsource=url,
        match=\regexp{\\},
        replace=\regexp{}]
    }
  }
}

\usepackage{xcolor}

\title{The Solution of the Zodiac Killer's 340-Character Cipher}
\author{David Oranchak, Sam Blake, Jarl Van Eycke}
\date{\today}

\begin{document}

\begin{titlingpage}
  \maketitle
  \begin{abstract}
    The case of the Zodiac Killer is one of the most widely known unsolved serial killer cases in history.  The unidentified killer murdered five known victims and terrorized the state of California. He also communicated extensively with the press and law enforcement. Besides his murders, Zodiac was known for his use of ciphers. The first Zodiac cipher was solved within a week of its publication, while the second cipher was solved by the authors after 51 years, when it was discovered to be a transposition and homophonic substitution cipher with unusual qualities. In this paper, we detail the historical significance of this cipher and the numerous efforts which culminated in its solution. 
  \end{abstract}
  \begin{figure}[h]
    \centering
    \includegraphics[width=0.35\columnwidth]{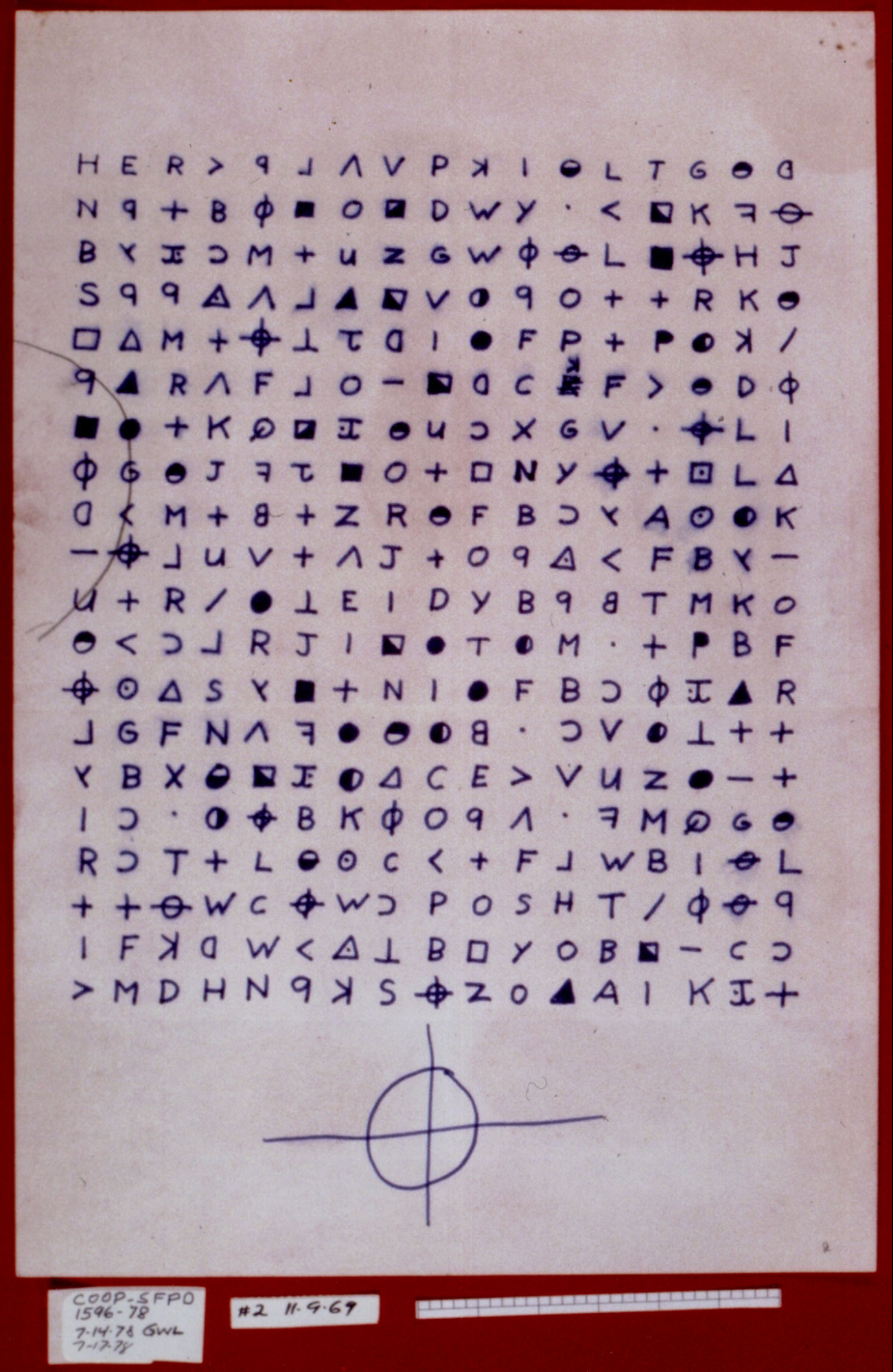} 
    \caption{The Zodiac Killer's 340-character cipher \cite{voigt3402007}}
    \label{fig:Z340}
  \end{figure}  
\end{titlingpage}

\begin{twocolumn}

\raggedright
\section{Background}
\justifying
In the late 1960s, a serial killer known as \textit{The Zodiac Killer} (or \textit{Zodiac}) operated in northern California, killing at least five people \cite{lcwkc1969}.  He targeted mostly young couples in isolated areas, but one victim was a cab driver in San Francisco.  During his crime spree, Zodiac mailed taunting letters to regional newspapers.  The letter writer took credit for the crimes, boasted about his attacks, ridiculed the police for not being able to capture him, threatened to commit more crimes unless newspapers published his correspondences, and mailed four ciphers \cite{drexlerup1969} and cryptic messages.  The first cipher was published in local newspapers on August 1, 1969 and was solved by a puzzle-loving couple within days \cite{murdercode1969}.  Zodiac's second cipher, known as \textit{Z340} \textit{(Figure \ref{fig:Z340})}, appeared in newspapers on November 12, 1969 and is the topic of this paper.  The investigations into the serial killer span multiple San Francisco Bay Area jurisdictions \cite{power1969}, but the case remains unsolved. 

\raggedright
\subsection{Homophonic Substitution Ciphers}
\justifying
Early substitution ciphers used only a single substitution for each plaintext letter.  Because a single substitution fails to conceal letter frequencies, these ciphers were vulnerable to being broken using \textit{frequency analysis} (e.g., counting cipher symbols to see how they might match expected letter frequencies in the target language).  This later evolved into the use of multiple substitutions, known as \textit{homophones} or \textit{variants}, for individual plaintext letters. When the quantity of variants per plaintext letter is selected proportional to the relative frequency of the letter in the plaintext language, the distribution of letter frequencies of the ciphers becomes more uniform, resulting in ciphers that are harder to break. The earliest known example of the use of homophonic substitution in the West is a cipher created by the Duchy of Mantua \cite{kahn1996codebreakers1}.  Prior to this is a system described by 14th century Arab cryptologist `Alī ibn Mu\d{h}ammad Ibn al-Durayhim whereby individual letters are replaced by pairs of letters whose numerical values add up to the original letters \cite{kahn1996codebreakers2}.\\

Today homophonic substitution ciphers are readily broken using sophisticated specialized software such as \textit{AZdecrypt} \cite{azdrepo} and \textit{ZKDecrypto} \cite{zkdrepo}. These programs automatically explore vast spaces of candidate solutions by generating pseudorandom cipher keys, iteratively making small pseudorandom perturbations, and measuring solution fitness using language statistics.  They use an optimization technique known as \textit{hill climbing}, an iterative process that makes incremental changes and improvements to arbitrary solutions to problems.

\raggedright
\subsection{Transposition Ciphers}
\justifying

In contrast with substitution ciphers, which replace letters with other letters or symbols, transposition ciphers involve rearrangements of plaintext messages which rearrange them into seemingly unreadable messages.  In the fifth century BCE, the Greeks were purported to use a device known as the \textit{scytale} \cite{kahn1996codebreakers3} consisting of a strip of parchment wrapped tightly around a wooden dowel or rod.  The plaintext message was written lengthwise along the wrapped rod.  Then, the wrapped parchment was unraveled, and the message became an unreadable rearrangement of the original, until the parchment was wrapped again around a similarly sized rod to restore the original order of the message.\\

Kahn reports \cite{kahn1996codebreakers4} what appears to be the earliest description of \textit{keyed columnar transposition}, by John Falconer, published in 1685.  This form of transposition involves writing out the plaintext message into a grid, then rearranging the column order based on a keyword or phrase.  The transposed message is taken by then reading the message horizontally.  Reversing the process decrypts the message.\\

Early descriptions of transposition ciphers can be found in Giambattista della Porta's 1563 book, \textit{De Furtivis Literarum Notis} \cite{dellafurtivis}.  Porta's book includes descriptions of transposition schemes such as distribution of messages into multiple rows, and a keyed transposition that rearranges plaintext in-place governed by the numerical equivalents, interpreted as linear positions, of letters in the key phrase \cite{tomokiyo2015}.

\raggedright
\subsection{The Zodiac Killer's First Cipher: \textit{Z408}}\label{sec:z408}
\justifying

\begin{figure}[h]
  \centering
  \includegraphics[width=0.8\columnwidth]{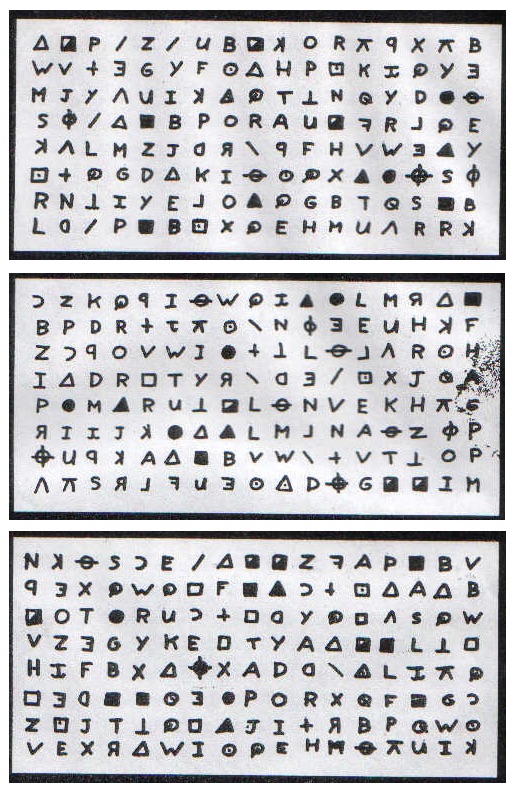} 
  \caption{The Zodiac Killer's 408-character cipher \cite{voigt408}}
  \label{fig:Z408}
\end{figure}

After his attacks on two young couples, Zodiac mailed his first cipher \textit{(Figure \ref{fig:Z408})} on July 31, 1969 \cite{ntokiv1969}.  Denoted \textit{Z408} for its length, the cipher was sent in three parts to three local newspapers:  the \textit{Vallejo Times-Herald}, \textit{San Francisco Examiner} and the \textit{San Francisco Chronicle} \cite{vmmtf1969}.  A handwritten letter accompanied each part.  Each letter essentially said the same thing: that the author of the letter was the murderer of a teenage couple around Christmas of 1968, and was the attacker of another couple around July 4, 1969, which resulted in another death.  The letters included specific details about the case, intended to authenticate the letter writer as the perpetrator.  He also threatened to go on another killing spree if the newspapers failed to publish the ciphers.

\begin{figure}[h]
  \centering
  \includegraphics[width=\columnwidth]{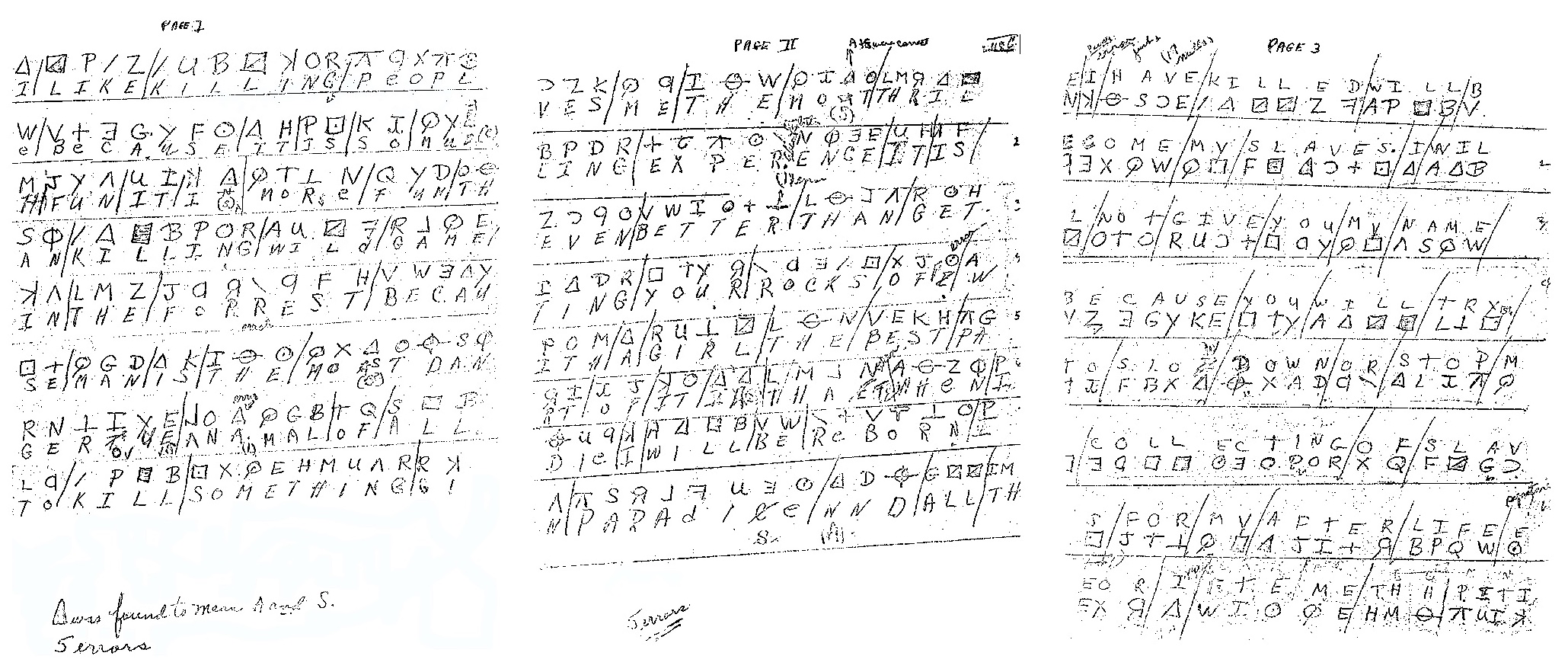} 
  \caption{The Hardens' handwritten solution of \textit{Z408}. (\cite{graysmith2007zodiac} pp. 224--225)}
  \label{fig:Z408_Harden}
\end{figure}

Within a few days, all three local newspapers published their portions of the ciphertext \cite{ntokiv1969} \cite{vmmtf1969} \cite{ccim1969}.  Local police requested help from the US Navy \cite{ccim1969}, the FBI (Federal Bureau of Investigation), the California Bureau of Investigation \cite{cknl1969}, and Donald C. B. Marsh who was the head of the American Cryptogram Association \cite{gcdmszk1969}.\\

On August 8, 1969, eight days after the cipher was mailed, the \textit{San Francisco Chronicle} received a solution \textit{(Figure \ref{fig:Z408_Harden})} from Donald and Bettye Harden, who solved the cipher after seeing it in the newspapers \cite{murdercode1969}.  Owing to Donald's boyhood interest in ciphers, Bettye's insights and persistence, and some trial and error, the Hardens discovered the plaintext message which was encoded using a homophonic substitution cipher (also known as a \textit{monoalphabetic substitution cipher with variants} \cite{friedman1938military}).  The plaintext \cite{murdercode1969} follows (error corrections are in brackets)\footnote{This version of the plaintext is reproduced from \cite{murdercode1969}.  Other articles published slightly different variations of the plaintext.}:\\

\begin{adjustwidth}{0.02\textwidth}{0.02\textwidth}
  \texttt{I like killing people because it is so much fun it is more fun than killing wild game in the \textit{forrest} [forest] because man is the most \textit{hongertue} [dangerous] animal of all to kill something \textit{give eryetheyo a} [gives me the most] thrilling experience it is even better than getting your rocks off with a girl the best part of it \textit{I athae} [is that] when I die I will be reborn in paradice and all the I have killed will become my slaves I will not give you my name because you will \textit{trs} [try] to \textit{sloi} [slow] down or \textit{atop} [stop] my collecting of slaves for my afterlife \textit{ebeo riet emeth hpiti}.} \cite{murdercode1969}
\end{adjustwidth}

\begin{figure}[h]
  \centering
  \includegraphics[width=0.95\columnwidth]{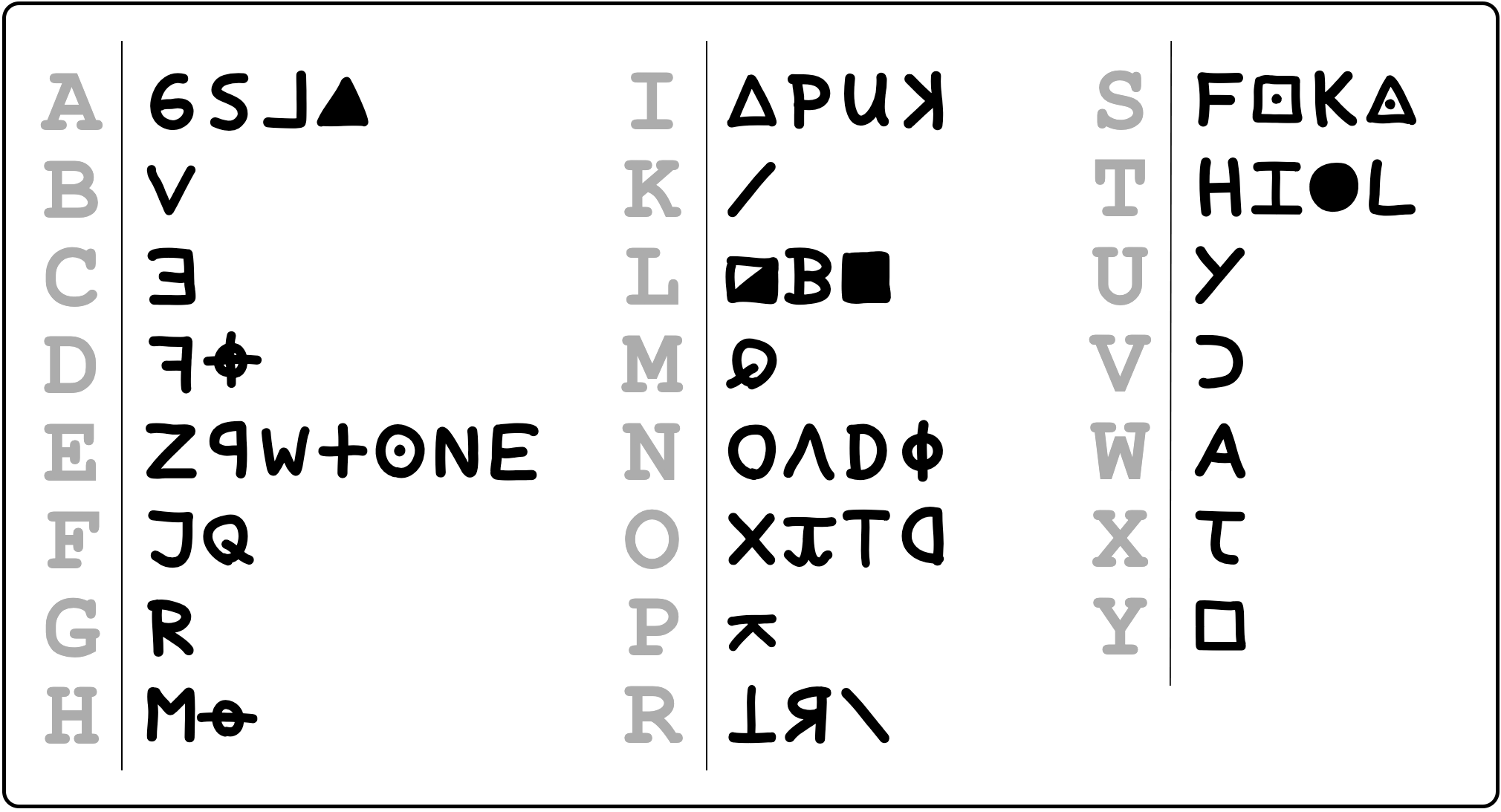} 
  \caption{\textit{Z408} substitution key.  Note: There are no substitutions for the letters J, Q, and Z since none appear in the plaintext.}
  \label{fig:Z408_key}
\end{figure}

The substitution key derived from the Hardens' solution is shown in \textit{Figure \ref{fig:Z408_key}}.\\

The plaintext contained misspellings and/or encipherment mistakes, resulting in some cipher symbols having multiple possible interpretations \cite{oranchak408key}.  It also includes a garbled 18-character portion at the end whose meaning has not been determined.  One conjecture is that symbols in this section were copied from above to serve as filler \textit{(Figure \ref{fig:Z408_filler})}, making the third section the same size as the other two sections \cite{cisco2009}.  Another conjecture is that a second decryption process is required to decode this portion.

\begin{figure}[h]
  \centering
  \includegraphics[width=0.55\columnwidth]{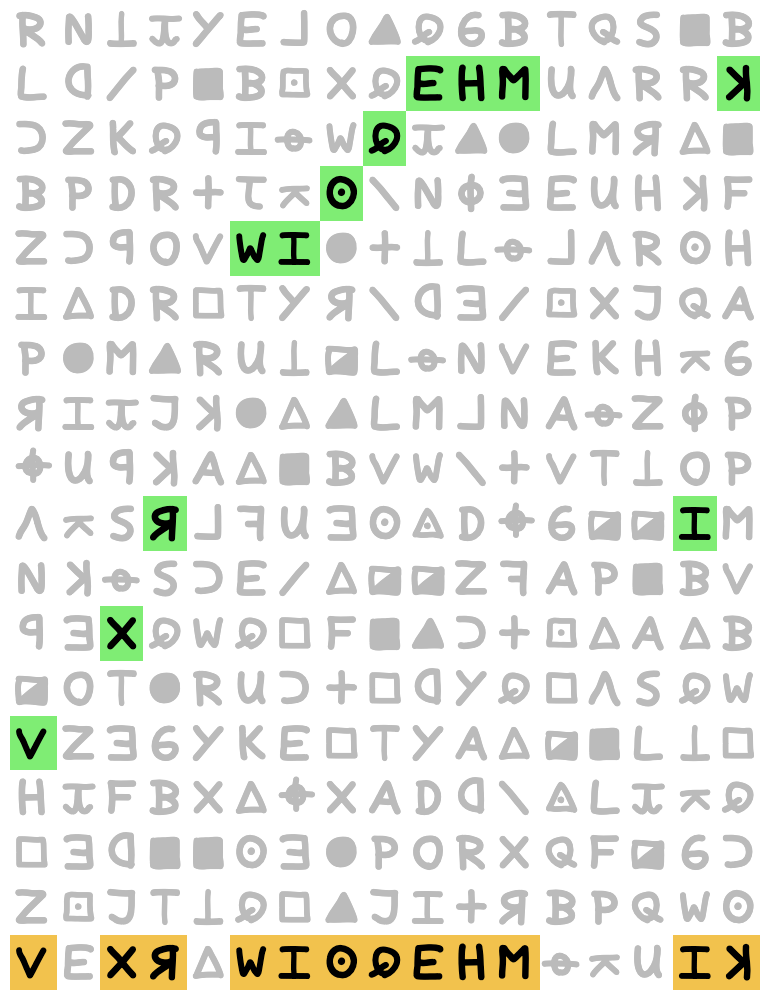} 
  \caption{Filler conjecture.  Symbols along the last line are possibly copied from above. \cite{cisco2009}}
  \label{fig:Z408_filler}
\end{figure}

\raggedright
\subsection{The Zodiac Killer's Second Cipher: \textit{Z340}}
\justifying

\begin{figure}[h]
  \centering
  \includegraphics[width=0.9\columnwidth]{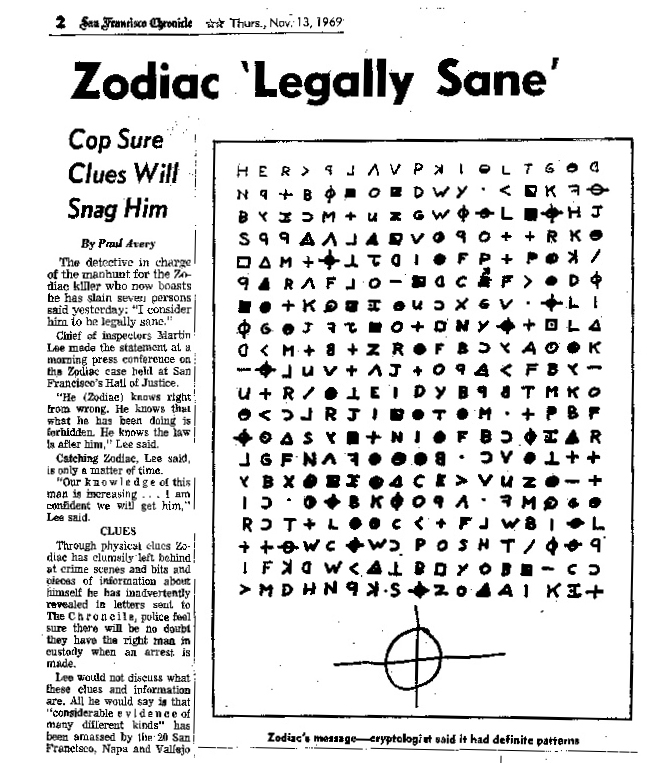} 
  \caption{The Zodiac Killer's 340-character cipher as published by the {\normalfont San Francisco Chronicle} on November 13, 1969.}
  \label{fig:Z340_sf_chronicle}
\end{figure}

On November 8, 1969, Zodiac mailed a second cipher \textit{(Figure \ref{fig:Z340})} to the \textit{San Francisco Chronicle} \textit{(Figure \ref{fig:Z340_sf_chronicle})} \cite{ikszc1969}. \textit{Z340}, so named for its length, resembled Zodiac's first cipher, \textit{Z408}, for its use of a mixture of letters, backwards letters, shapes and other symbols, arranged into an orderly grid.  Accompanying the cipher was a greeting card \textit{(See Appendix \ref{the_appendix}, Figures \ref{fig:Z340_card_1} and \ref{fig:Z340_card_2})}, inside which Zodiac had written:\\[0.5\baselineskip]
\begin{adjustwidth}{0.05\textwidth}{0.05\textwidth}
  \textit{``This is the Zodiac speaking. I though [sic] you would need a good laugh before you hear the bad news.  You won't get the news for a while yet.  PS could you print this new cipher on your frunt [sic] page? I get awfully lonely when I am ignored, so lonely I could do my Thing!!!!!!}

  \textit{Des July Aug Sept Oct = 7''} \cite{graysmith2007zodiac}\\[0.5\baselineskip]
\end{adjustwidth}
As with his first cipher, Zodiac pressured the newspapers to publish his ciphers by threatening to commit more violent acts.  Within a few days, the cipher was published in several local newspapers \cite{carson1969} \cite{bernhard1969} \cite{ikszc1969}.  The \textit{San Francisco Chronicle} reported that homicide detectives believed the cipher may have contained information about Zodiac's additional victims \cite{ikszc1969}.  The detectives sent the cipher to several experts and expected it to be deciphered in a few days.  Graysmith \cite{graysmith2007zodiac2} claimed that the NSA (National Security Agency) said that the cipher definitely contained a message.  Despite more than a half century of attempts to decode the infamous \textit{Z340} cipher, which even ranked first on the FBI's list of top unsolved ciphers \cite{campbell2011}, it remained uncracked.

\raggedright
\subsection{The Zodiac Killer's remaining ciphers}\label{sec:remaining}
\justifying
On April 20, 1970, Zodiac mailed another taunting letter with a cipher to the \textit{San Francisco Chronicle}.  He wrote:\\[0.5\baselineskip]

\begin{adjustwidth}{0.05\textwidth}{0.05\textwidth}
  \textit{``This is the Zodiac speaking.  By the way have you cracked the last cipher I sent you?  My name is \rule[2pt]{30pt}{0.25pt}''.} \cite{zsnlct1969}\\[0.5\baselineskip]
\end{adjustwidth}

That opening was immediately followed by his new, much shorter cipher, known as \textit{Z13} \textit{(Figure \ref{fig:Z13})}.\\

\begin{figure}[h]
  \centering
  \includegraphics[width=0.95\columnwidth]{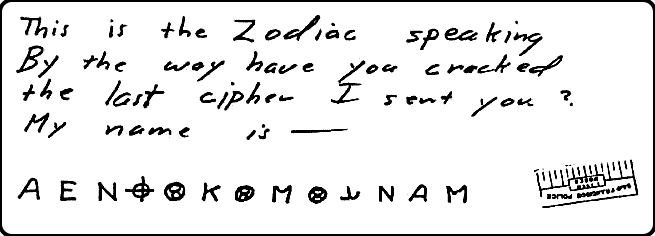} 
  \caption{\textit{Z13}, Zodiac's 13-character cipher \cite{zsnlct1969}}
  \label{fig:Z13}
\end{figure}

Cryptanalysis is all but impossible for such a short ciphertext, because solutions are not guaranteed to be unique, and many thousands have been proposed and generally aren't scientifically falsifiable \cite{oranchakz13sol2023}.
Zodiac mailed another short cipher, \textit{Z32} \textit{(Figures \ref{fig:Z32} and \ref{fig:z32_1})}, on June 26, 1970, along with a map \cite{zshksfo1970} \textit{(Figure \ref{fig:z32_2})}.  In the accompanying letter, Zodiac claimed the solution to the cipher gave the location to a bomb that he had devised and hidden somewhere.  He had written about his bomb design and plans in previous correspondences.  Once again, Zodiac had created a cipher that resists cryptanalysis due to its short length and abundance of unique symbols.

\begin{figure}[h]
  \centering
  \includegraphics[width=0.95\columnwidth]{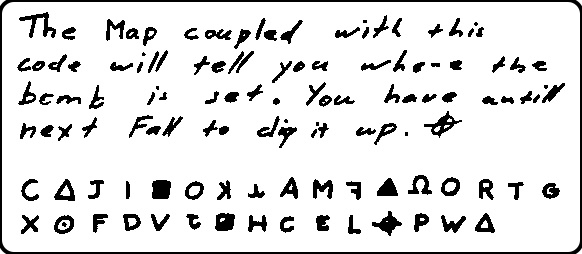} 
  \caption{\textit{Z32}, Zodiac's 32-character cipher \cite{voigtbutton}}
  \label{fig:Z32}
\end{figure}

\raggedright
\section{Historical efforts to break \textit{Z340}}\label{sec:historical}
\justifying

By the time \textit{Z340} appeared in local newspapers, the Zodiac case was becoming well known.  The growing media focus elevated his \textit{Z340} cipher, drawing increased interest in finding its solution.  Shortly after Zodiac mailed it, many people raced to decipher it, and some asserted their decryptions were correct.  But none were definitively endorsed by law enforcement, so these efforts continued for decades.  Hundreds, if not thousands, of attempts and purported solutions exist.  Providing all of them is outside the scope of this paper; however, we will detail several notable attempts. 

\paragraph{Initial months \textit{(1969--1970)}:}  On November 12, 1969, it was reported that \textit{Z408}'s key was tried on \textit{Z340}, but failed to unlock any message \cite{bernhard1969}.  It was also reported that the cipher was turned over to ``professional cryptographers'' \cite{smithhksz1969}.  ``Homicide detectives have forwarded copies of the cryptogram to a number of experts and expect it to be deciphered within a few days'' \cite{ikszc1969}.  At the time, authorities suspected that victims' names might have been concealed in the cipher, owing to the inclusion of August in Zodiac's list of months associated with his attacks, which he had included on the accompanying card \textit{(Figure \ref{fig:Z340_card_2})}.  At that point, no attack that could be definitively associated with Zodiac occurred in August.  Reporter Paul Avery speculated Zodiac might have been taking credit for a recent killing of San Jose girls Deborah Gay Furlong and Kathleen Snoozy in August 1969, despite the doubts of the investigator of that case \cite{ikszc1969}.\\

\textit{San Francisco Examiner} reported that an analysis of \textit{Z340} indicated Zodiac had changed his encipherment method.  ``Cryptographers noted that the man has altered the code used in the earlier messages, in such a way as to indicate he is a sophisticated expert on this sort of puzzle'' \cite{conantzb1969}.  They noted the addition of new symbols that were not present in the first cipher \cite{bernhard1969}.  Using \textit{Z408's} key, \textit{San Francisco Examiner} staff attempted to decipher \textit{Z340's} content.  This resulted in gibberish, even after exploring potential decodings in various directions or skipping letters in a systematic manner.  \textit{Examiner} staff also noted that some symbols that seemed new might simply be the result of sloppy penmanship or hurried construction of the cipher.  \\

On November 13, 1969 the \textit{San Francisco Chronicle} reported \cite{averyzls1969} that amateur cryptographers ``by the hundreds'' were at work trying to decipher \textit{Z340}.  One of them was reported to have said the cipher definitely contained word patterns hidden in the 340 symbols:  ``There is a definite message.  Testing shows it is not just gibberish.''\\

\paragraph{Donald C. B. Marsh:} A November 14, 1969 article \cite{gcdmszk1969} reported Colorado School of Mines math professor Dr. Donald C. B. Marsh was asked by the San Francisco Police Department (SFPD) to assist with the decryption of Zodiac's ciphers.  At the time, Marsh had been a member of the American Cryptogram Association for about 20 years, where he solved thousands of ciphers and also served as its President.\\

Marsh had been asked to decrypt \textit{Z408}, but the Hardens had already solved it before him.  However, he independently verified their solution, and indicated that the unresolved last 18 letters could be nulls, a secondary encipherment not yet discovered, or an anagram.  He thought that while Zodiac made \textit{Z408} more complex by using homophones, it didn't require a cryptography expert to devise, and felt there were aspects that seemed like the work of an amateur.  Further, he issued a challenge to Zodiac, asking him to send him a cipher containing the killer's identity.  Marsh hoped Zodiac would comply, and said he could decrypt such a cipher despite Zodiac's efforts.  Later, Marsh saw \textit{Z340} published in the newspaper, and said he was eager to work on decrypting it to help law enforcement.\\

\paragraph{FBI's cryptanalysis:} On November 12, 1969, the San Francisco Police Department (SFPD) submitted \textit{Z340} to the FBI \cite{fbivault2} and requested a comparison with \textit{Z408} and assistance to decipher the message \textit{(Figure \ref{fig:FBI})}.\\

\begin{figure}[h]
  \centering
  \includegraphics[width=0.95\columnwidth]{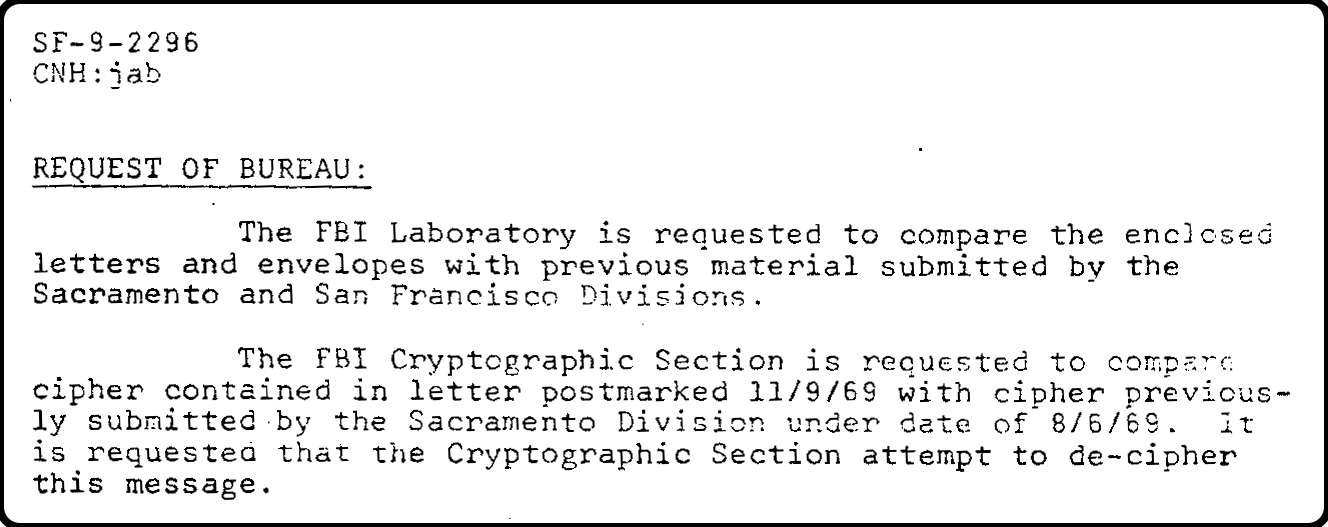} 
  \caption{Request for assistance from the FBI Laboratory and its cryptographic section. \cite{fbivault2}}
  \label{fig:FBI}
\end{figure}

Publicly available Zodiac case files by the FBI confirm their involvement in attempts to break \textit{Z340}.  In a cryptanalysis report dated December 18, 1969, the FBI summarized their findings on \textit{Z340} \textit{(Figure \ref{fig:FBI_340_analysis})}: \cite{fbivaultgd}
\begin{itemize}
  \item \textit{Z408}'s key failed to produce a message.
  \item The FBI tried applying the \textit{Z408} key as part of a ``combination cryptosystem'', including linear and route transposition, with a negative result.
  \item Applying \textit{Z408}'s key produced plaintext letter \textit{n}-gram frequencies that still resembled expected frequencies of English text.
  \item Approximately 20\% of the ciphertext was new (i.e., those symbols did not appear in \textit{Z408}).
  \item They examined the ciphertext for ``cyclic use of variants'', or predictable sequences of alternating substitutions for individual plaintext letters.  The sequences they identified were used in subsequent cryptanalysis.
  \item They applied crib words and phrases used before by Zodiac to try to reveal more of the plaintext message.  This was also done with grouping of various selections of cyclic variants.  They also tried to extract plaintext in different reading directions: backwards, columnar, alternating row-wise directions (a ``snake'' pattern), etc.
  \item They made attempts to anagram portions of the ciphertext with words used as cribs, such as ``Christmass'' (a known misspelling used by Zodiac), with focus on certain symbols, and on ciphertext regions having more repeating symbols.
  \item They concluded: ``No decryption could be affected.''
\end{itemize}  

\paragraph{Attempts from other agencies:} An alleged attempt by the NSA from the late 1960s or early 1970s to solve \textit{Z340} was documented on an online Zodiac forum \cite{voigtnsa}.  The document \textit{(Figure \ref{fig:NSA_z340})} reportedly belonged to Jack Mulanax, a Vallejo Police Department detective who worked the Zodiac case, and shows a printout of the phrase ``ANOTHERSLAVE'' being cribbed or applied to multiple positions in the ciphertext to try to yield more words or phrases from the partial substitution. \\ 

Newspaper articles claimed the National Security Agency \cite{smithztccp1970}, and Navy cryptographers \cite{pruittbk2001}, also attempted to break \textit{Z340}.  The \textit{Los Angeles Times} reported on May 8, 1970 that attempts to solve \textit{Z340} were so far met with failure, including attempts by ``computers at the National Security Agency in Washington'' \cite{smithztccp1970}.  Experts speculated that Zodiac may have deliberately constructed a false code to muddle the investigation.  However, we have not yet found any other confirmation or details of such attempts, perhaps due to classification levels of such information or other information security based impediment.\\

\paragraph{American Cryptogram Association (ACA), 1970:}  ACA publication, \textit{The Cryptogram}, published information about Zodiac and his ciphers in their January--February 1970 edition \cite{acacryptogram1}.  An ACA member prepared a clearer copy of the cipher from the newspapers.  However, \textit{The Cryptogram} reported ``our top solvers have been working on it… but this has not been solved.''  The next issue \cite{acacryptogram2} reported ``there is no word of any solution yet'', and published a newer copy of the cipher with corrections to apparent mistakes in its reproduction.\\

\paragraph{Robert Graysmith \textit{(1986)}:}  The author Robert Graysmith was a cartoonist at the \textit{San Francisco Chronicle} at the time of Zodiac's crime spree.  Years later, in 1986, Graysmith published \textit{Zodiac}, a book about the case, in which he identified suspects and claimed he solved \textit{Z340} \cite{graysmith2007zodiac3} \textit{(Figure \ref{fig:Z340_Graysmith})}.  Further, he claimed his so-called solution was verified by two members of the ACA: Greg Mellen and Eugene Waltz.  We have been unable to find any evidence of such verifications existing outside of Graysmith and his book. Moreover, the FBI conducted a cryptanalysis of Graysmith's claimed solution and concluded that ``the solution has been forced'', ``any random selection of words could be arranged to be as ‘logical' [as Graysmith's transposed plaintext]'', and the ``sense of rightness is completely absent in the proposed solution'' \cite{fbivault5}.  Our own analysis led us to the same conclusion \cite{oranchaklcz1}.

\begin{figure}[h]
  \centering
  \includegraphics[width=0.95\columnwidth]{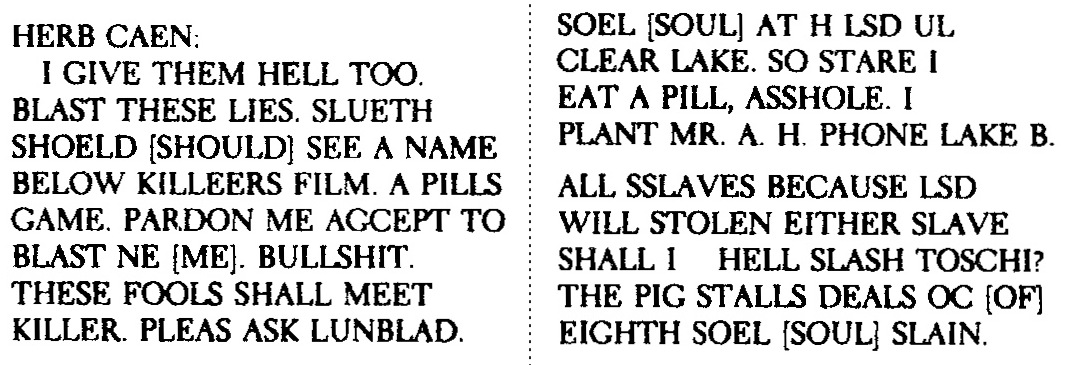} 
  \caption{Graysmith's claimed solution to \textit{Z340} \cite{graysmith2007zodiac3}.}
  \label{fig:Z340_Graysmith}
\end{figure}

\paragraph{American Cryptogram Association (ACA), 1988--1989:}  In a 1988 issue of \textit{The Cryptogram} \cite{acacryptogram3}, a non-member claimed to have an accurate Zodiac cipher solution, and requested assistance from ACA members to verify it.  We don't know if such assistance was provided, or of the outcome of any attempts to verify the solution.  Then in 1989, \textit{The Cryptogram} published a computer program \cite{acacryptogram4} that allowed ACA members to experiment with \textit{Z340}, and reported a member, who went by the pseudonym \textit{TRIODE}, found a possible decryption of the last two lines \textit{(Figure \ref{fig:Z340_ACA})}.

\begin{figure}[h]
  \centering
  \includegraphics[width=0.8\columnwidth]{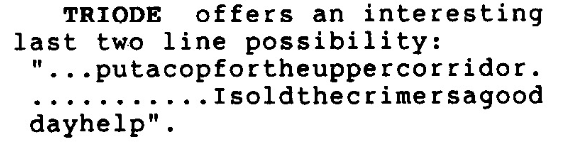} 
  \caption{Claimed decipherment of last two lines of \textit{Z340} \cite{acacryptogram4}}
  \label{fig:Z340_ACA}
\end{figure}

\paragraph{Gareth Penn, 1981--1985:}  Gareth Penn (pen name \textit{George Oakes}), an author who wrote frequently about the Zodiac case, accused a University of California, Berkeley professor of being the killer, and developed a large body of writings containing theories, many mathematical and pseudo-cryptographical in nature, in support of his claims.  Many of them appear in his self-published books \textit{Times 17} \cite{penn1987times} and its follow-up \textit{The Second Power} \cite{pennsecond}.  On several occasions, the FBI analyzed his cryptographic ideas \cite{fbivault5b} and gave negative conclusions:

\begin{itemize}
  \item ``The bulk of [Penn's] theory… is based on speculation and a multitude of assumptions.''
  \item ``It is possible that some of the assumptions are correct. Many, if not most, appear to be forced with results being used selectively if they are in keeping with the overall theme of the solution.''
\end{itemize}

\subsection{Solution claims in popular media \textit{(2011--present)}}

The popularity of the case and noteworthy nature of \textit{Z340} led many amateur codebreakers to join the efforts to solve the cipher.  Many so-called solutions were developed that ``went viral'' and received significant attention in the news media.  Examples include Corey Starliper's arbitrary decipherment system in 2011 \cite{schillemat2011}, Daryll Lathers' anagrammed solution in 2012 \cite{cleveland2013}, and Gary Stewart's 2014 production of his father's name from Zodiac's ciphers \cite{green2014}.  Proposed solutions to \textit{Z340} were frequently posted at online forums specializing in discussions on the Zodiac case. Many alleged solutions were also sent to the present authors to solicit analysis and feedback.  A diverse range of decryption methods were used, with a significant portion failing to meet standards of testing for correctness, such as having sufficient \textit{unicity distance} (minimum ciphertext length to guarantee a single correct key), and demonstrating uniqueness of the deciphered message, such that no other equally plausible messages can be generated with the same approach. \\

\paragraph{Corey Starliper \textit{(2011)}:}  An example of a solution claim in popular media that attracted a lot of attention was one made by Corey Starliper in 2011.  His claim was reported in the \textit{Tewksbury Patch} with the headline, ``Tewksbury Native: I've Cracked The Code Of The Zodiac Killer'' \cite{schillemat2011}.  The headline suggested Zodiac's cipher was solved, causing the story to go viral multiple times \cite{starliper2017} and spread rapidly on social media.  David Oranchak developed an analysis pointing out the many flaws in the solution claim \cite{oranchak2011cs}, concluding that it was a hoax.  The original news source published a follow-up article detailing criticisms from cryptographers, researchers, and investigators about Starliper's claims \cite{schillemat2011fire}.  The story exemplified a trend in Zodiac-related news that uncritically reports individuals' claims\footnote{Examples include: \cite{murhy2008}, \cite{abc7perez}, \cite{cbs13dk}, and \cite{gorner2011}}. 

\paragraph{Gary Stewart \textit{(2014)}:}  In 2014, authors Susan Mustafa and Gary Stewart published \textit{The Most Dangerous Animal of All: Searching for My Father… and Finding the Zodiac Killer} \cite{stewart2014most}.  The book chronicled Stewart's search for his biological father, Earl Van Best Jr., whom he identified as the Zodiac Killer based on circumstantial evidence.  The book was adapted for production into a documentary series on \textit{FX Network} in 2020 \cite{gajanan2020}.  During the production, producers tried to validate Stewart's claims with a private investigator, who ultimately uncovered evidence contradicting and discrediting Stewart.  Stewart and Mustafa were confronted with these findings on camera for the docuseries.  \textit{Vulture} detailed \cite{fernandez2020} how the series disproved the book's claims, and included Mustafa's contrite admission:  ``The lesson to be learned here, for me, is setting out to prove something as opposed to disproving something is a big mistake.  That’s a big mistake for a journalist to make. I have to own up to it. I have to take responsibility for it.''  Her admission is a reminder of \textit{confirmation bias} \cite{nickerson1998confirmation}, a tendency which can cause people to focus on supportive evidence for their claims and beliefs while ignoring contrary information.

\paragraph{Craig Bauer \textit{(2017)}:}  In 2017, the \textit{History} channel released \textit{The Hunt for the Zodiac Killer} \cite{history2017}, a 5-part television series in which a codebreaking team consisting of Kevin Knight, Ryan Garlick, Craig Bauer, and David Oranchak was formed to explore the Zodiac case and attempt to solve \textit{Z340}. In the final episode an alleged partial solution (\textit{Figures \ref{fig:Z340_bauer_soln_1}} and \textit{\ref{fig:Z340_bauer_soln_2}}) was presented by Craig Bauer, a professor of mathematics at York College of Pennsylvania, former Scholar-in-Residence at NSA's Center for Cryptologic History, and Editor-in-Chief of the journal \textit{Cryptologia}.  It reads: \\

\begin{adjustwidth}{0.02\textwidth}{0.02\textwidth}
  \texttt{HERE IT IS I KILL BOTH NIGHT AND DAY. I LIVE BY THE GUN BAREL AIM SO QUIT WISHING FOR GAME TO BE OVER PIGS IS MI WRIST NI LOCKS? NOW ANGRY DANGEROS. I WON'T CHANGE ANY OF GAME}\\
  \texttt{RICHERD M NIKSON} \cite{bauer2018}\\
\end{adjustwidth}

Producers of the show opted to excise the codebreaking team's objections to Bauer's so-called solution, and it was aired as if it were supported by the team's efforts and endorsement. Bill Briere, a classical cryptanalyst, wrote a critical review of the production, including: \textit{``Unlike History’s portrayal, cryptanalysis is not a creative-writing exercise, a game of Scrabble, or a magic trick. It is a science. If the damage that's been done is to be turned around, we cryptanalysts must call out the fakes when we see them and not allow cryptologic history to be rewritten by entertainers, charlatans, and profiteers.'' \cite{briere2018}}\\

Nick Pelling, author of the \textit{Cipher Mysteries} blog, gave a sardonic and negative review of Bauer's solution: 
\textit{%
``I suspect what most people would agree on about this `solution' are:\\
\begin{itemize}
\item it’s primarily intuitive, and not really ‘cryptological’ in any useful sense of the word.\\
\item it’s either really brilliant or really foolish, and almost certainly nowhere inbetween.'' \cite{pelling2017}
\end{itemize}
}

\begin{figure}[h]
  \centering
  \includegraphics[width=0.95\columnwidth]{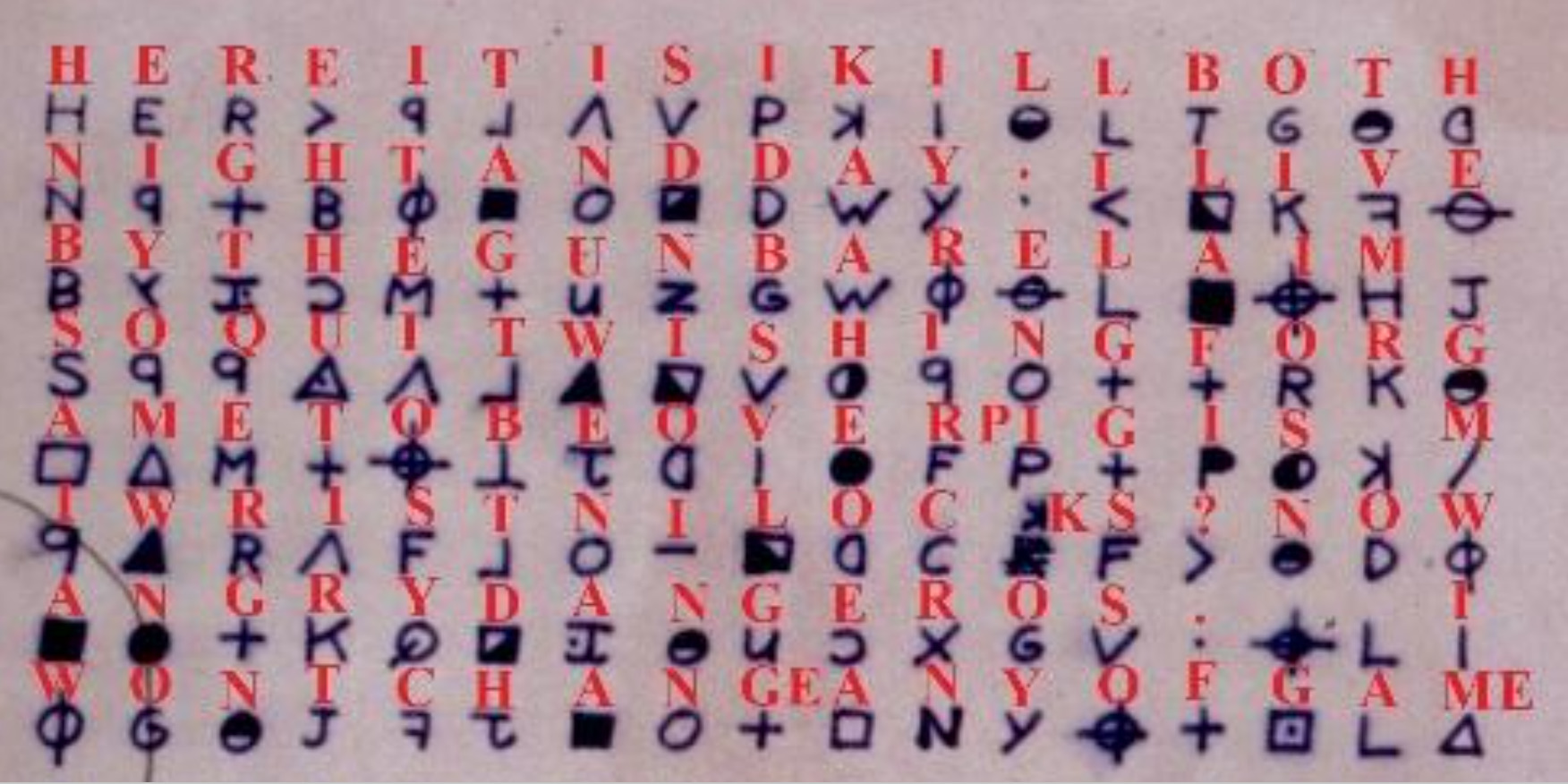} 
  \caption{The Craig Bauer alleged solution to the first 8 lines of \textit{Z340}. \cite{bauer2018}}
  \label{fig:Z340_bauer_soln_1}
\end{figure}

\begin{figure}[h]
  \centering
  \includegraphics[width=0.85\columnwidth]{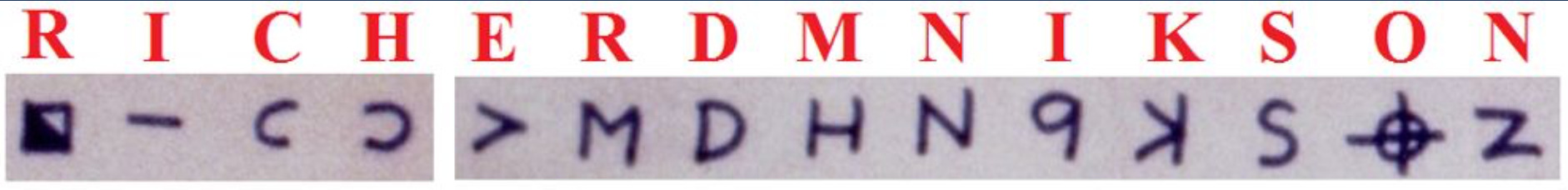} 
  \caption{The Craig Bauer alleged partial solution to lines 19 and 20 of \textit{Z340}. \cite{bauer2018}}
  \label{fig:Z340_bauer_soln_2}
\end{figure}

\raggedright
\subsection{Present authors' prior efforts (2012--2017)}
\justifying

The decryption success reported herein was preceded by the authors' many years of failed experiments, dead-end ideas, and efforts to summarize what was known about the Zodiac case and ciphers.  Some examples include:

\begin{itemize}
  \item Oranchak \cite{oranchakttbah1} ran an experiment using a large corpus of text to apply cribs to \textit{Z408} and \textit{Z340}, in an attempt to reveal possible plaintext in non-cribbed portions of the ciphers.
  \item In a 2012 article \cite{oranchaktcl2012}, Oranchak summarized reasons for and against belief in a real plaintext message in \textit{Z340}.
  \item Around 2013, Jarl Van Eycke began participating in Zodiac forums \cite{morford2010} and exploring hypotheses for \textit{Z340} using the \textit{ZKDecrypto} \cite{zkdrepo} software written by Brax Sisco, Michael Eaton, and Wesley Hopper.  This eventually led to his development of \textit{AZdecrypt} to achieve better performance and more cryptanalytic capabilities.  He also worked on novel statistical analyses, which helped guide various explorations of hypotheses and experiments.
  \item At the 2015 \textit{Symposium on Cryptologic History} \cite{oranchaktzc2015} and 2018 ACA conference \cite{oranchakwiz2018}, Oranchak gave a summary of Zodiac's ciphers, what was known about \textit{Z340}, and the progress to date towards any solution.  In 2017 he presented the challenges of separating true and false clues observed in \textit{Z340} \cite{oranchaktuzc2017}.
\end{itemize}

More examples involving academic work and computational methods are included in subsequent sections.

\raggedright
\subsection{Prior academic efforts (1993--2019)}
\justifying

A wide variety of academic studies explored codebreaking methods for the Zodiac ciphers in particular, and homophonic ciphers in general.  A sampling of these studies follows.\\

King and Bahler (1993) \cite{king1993algorithmic} described an algorithm that automatically detected and collapsed cyclic homophones (also known as cyclic variants).  Their approach was able to partially detect cyclic homophones in \textit{Z408} but failed to confirm them in \textit{Z340}.  Their previous paper \cite{king1993framework} described a measurement called \textit{multiplicity}, computed as the ratio of the cipher alphabet size to the ciphertext length.  This measurement combined with unicity distances provided a way to characterize the difficulty in arriving at solutions for homophonic ciphers.\\

Dao (2008) \cite{dao2008analysis} attempted to determine if the method used to produce \textit{Z340} was homophonic substitution, and produced hill climbing software for solving such ciphers.\\

Oranchak (2008) \cite{oranchak2008evolutionary} developed a dictionary-based attack using a genetic algorithm (an optimization method inspired by natural selection) that reframed homophonic substitution as a constraint satisfaction problem.  The technique effectively solved known ciphers such as \textit{Z408} but \textit{Z340} remained unbroken.\\

Basavaraju (2009) \cite{basavaraju2009heuristic} produced a homophonic cipher solver via genetic algorithm with crossover variations.  It was able to produce partial decrypts of \textit{Z408} but no solution to \textit{Z340} was found.\\

Raddum and Sýs (2010) \cite{raddum2010zodiac} explored the hypothesis that \textit{Z340} was not just a faked cipher (i.e., a random stream of meaningless symbols), by showing that the amount of orderly cycles of repeating symbols in the ciphertext was high compared to random sequences.  They explained a strategy for decryption of homophonic ciphers that worked for \textit{Z408} but failed to produce a solution for \textit{Z340}.\\

Ravi and Knight (2011) \cite{ravi2011bayesian} developed a Bayesian approach (updating probabilities and making predictions), using dictionaries and letter \textit{n}-gram language models to crack simple and homophonic substitution ciphers.  They demonstrated the approach's ability to automatically crack \textit{Z408}.\\

Berg-Kirkpatrick and Klein (2013) \cite{berg2013decipherment} improved \textit{Expectation-Maximization} (an iterative statistical algorithm) using random restarts, and applied the algorithm to homophonic ciphers, including \textit{Z340}, which failed to yield a solution.  Also, by successfully cracking many ciphers made to match properties of the real \textit{Z340}, the authors argued that \textit{Z340} was likely not a homophonic cipher as assumed.\\

Dhavare, Low, and Stamp (2013) \cite{dhavare2013efficient} developed a nested hill climber attack for homophonic ciphers that they also applied to \textit{Z340}.  They reported putative plaintexts but none are correct.\\

Nuhn et al. (2013) \cite{nuhn2013beam} used \textit{beam search}, a heuristic algorithm that narrows down a search space, to crack homophonic substitution ciphers, including \textit{Z408}.  They did not report the presumably negative results for \textit{Z340}.\\

Yi (2014) \cite{yicryptanalysis} described an attack based on the combination of homophonic substitution and columnar transposition.  When applied to \textit{Z340}, no legible plaintext was discovered.\\

Serengil and Akin (2011) \cite{serengil2011attacking} attacked Turkish homophonic ciphers using vulnerabilities of the Turkish language.\\

Zhong (2016) \cite{zhong2016cryptanalysis} used \textit{Hidden Markov Models (HMM)} to break homophonic substitution ciphers.  The attack was tested on \textit{Z340} but failed, even when assuming \textit{Z340} was a combination of homophonic substitution and a type of column transposition.\\

Vobbilisetty et al. (2017) \cite{vobbilisetty2017classic} applied \textit{Hidden Markov Models} to solve simple substitution ciphers, then expanded the technique to work on homophonic substitution ciphers.\\

Kopal (2019) \cite{kopal2019cryptanalysis} developed a codebreaking method for homophonic ciphers using simulated annealing (an optimization method inspired by metallurgical annealing) with fixed temperatures.\\

Juzek (2019) \cite{juzek2019using} introduced a cipher classification method using information theory and support vector machines (a supervised learning algorithm for classification and regression tasks).  The method correctly classified \textit{Z408} as a substitution cipher.  It classified \textit{Z340} as an ``advanced cipher'' or ``pseudo-cipher''.

\raggedright
\subsection{Computational tools and methods (1969--2020)}
\justifying

In addition to the computational methods described in the academic papers listed earlier, other computer based tools for cryptanalysis of homophonic ciphers have been developed over the years.  Some examples are summarized below.\\

In the Zodiac case files made available by the FBI, a cryptanalysis report dated December 18, 1969 appears \cite{fbivaultgd} \textit{(Figure \ref{fig:FBI_340_analysis})}.  The report describes examinations on \textit{Z340} and the application of \textit{EDP} (electronic data processing) \textit{runs}, wherein crib words and phrases were applied with computer assistance into the ciphertext at different positions, resulting in printouts that could be examined for potentially legible results from other regions of the ciphertext.\\

In 1989 the American Cryptogram Association's periodical \textit{The Cryptogram} featured an article \cite{acacryptogram4} about the Zodiac ciphers and included an interactive BASIC program which prompts the user to enter cipher-to-plaintext letter substitutions, and displays the resulting plaintext decryptions for examination.\\

\textit{CrypTool} \cite{esslinger2009cryptool} \cite{kneb2022} is open source cryptography e-learning software which supports numerous cryptographic algorithms.  Development began in 1998, and by 2012 its successor \textit{CrypTool 2} was released.  The software has evolved into an extensible, modular platform for configuring workflows and experiments with cryptographic algorithms, and has automated cryptanalysis and codebreaking features.  It also includes a homophonic substitution analyzer, which can break \textit{Z408}.\\

Software called \textit{ZKDecrypto} \cite{zkdrepo} was one of the first (if not the first) that could automatically and accurately decrypt the \textit{Z408} cipher.  It uses a \textit{tabu} hill climber search to explore random modifications to candidate substitution keys.  Recent modifications are stored in a \textit{tabu list}, a memory structure that stores recently explored solutions, which helps guide the search into novel territory in the search landscape.  \textit{ZKDecrypto} was a group programming effort started in 2006 by Brax Sisco, Michael Eaton, and Wesley Hopper, with some contributions from David Campbell.  Sisco started with a command line version that could automatically solve \textit{Z408}.  Eaton and Hopper joined the project later to add a graphical user interface.\\

In 2007 Oranchak developed a web-based interactive tool for cracking \textit{Z340} \cite{oranchakwt2007}.  Users could visit the site and experiment with their own symbol substitutions to try to solve the cipher.  The tool assumed \textit{Z340} was constructed as a simple homophonic substitution, so in retrospect the users were always doomed to failure.\\

In 2007 University of North Texas professor Ryan Garlick led an effort \cite{unt2007} to implement a heuristic search for solutions to \textit{Z340}, including a server-based architecture that received candidate keys from distributed clients, as a way to parallelize and scale up the codebreaking task.\\

Oranchak developed a limited dictionary-based substitution solver in 2008 \cite{oranchak2008evolutionary}. It uses a genetic algorithm to slide a non-conflicting set of words into candidate locations within a highly-constrained substring of a ciphertext.  The quality of solutions is measured using letter \textit{n}-gram statistics.\\

In 2011 Oranchak developed \textit{CryptoScope} \cite{oranchakzcc2011}, another web-based interactive tool for performing cryptanalysis on \textit{Z340} and other ciphers.  It was designed to produce statistical analyses and search for meaningful patterns in ciphertexts. \\

\textit{CryptoCrack} \cite{philcrow2011}, initially released in 2011 and regularly updated, is interactive codebreaking software developed by Phil Pilcrow that can automatically crack over 60 different cipher types.\\

Mike Cole developed a cipher generator in 2012 \cite{colecg2012}, designed to produce Zodiac-like homophonic substitutions, which is especially useful for testing hypotheses and automated solvers.\\

In 2013 Umanovskis \cite{uman2013} developed \textit{lgp-decrypto}, a linear genetic algorithm to attack \textit{Z340}.  He also developed \textit{zkdecrypto-lite}, a lightweight version of \textit{ZKDecrypto} designed for batched experiments \cite{uman2012}. \\

Oranchak in 2013 released an interactive word search tool \cite{oranchakwsg2013} that interprets cipher symbols directly as plaintext, and conducts searches for words entered by users.\\

In 2017, Heiko Kalista developed several tools related to cryptanalysis of Zodiac ciphers.  One was \textit{Peek-A-Boo} \cite{kalistapab2017}, a tool for visualizing transposition schemes.  Kalista designed it to improve the exploration of different transpositions of \textit{Z340} in a more visual manner.  He also created an automatic solver called \textit{Your Secret Pal} \cite{kalistamac2018} which uses a hill climber to crack homophonic substitution ciphers quickly.  Additionally, he created a homophonic cipher generation tool called \textit{Cipher Factory} \cite{kalistayet2017}, which allows users to produce test ciphers with adjustable parameters.\\

Oranchak released another homophonic cipher cryptanalysis tool called \textit{Cipher Explorer} in 2017 \cite{oranchakce2017}.  It was designed to display the various Zodiac ciphers symbolically, to more resemble the original ciphertexts and layouts, and gives users many tools for marking up ciphertexts for presentational needs.\\

In 2019 Fritz Reichmann developed \textit{cDecryptor} and \textit{jDecryptor} \cite{reichmann2020}, a hill climbing algorithm written in C++ and Java, respectively, for solving homophonic substitution ciphers.  \\

George Belden released \textit{Project Zenith} \cite{belden2019}, another hill climber for homophonic substitution ciphers, in 2019.  Inspired by \textit{AZdecrypt}, it started as a standalone program which evolved into a web-based version.\\

In 2020 Jonathan Block created an interactive web-based solver \cite{block2020} that lets users supply substitutions for cipher symbols to produce plaintext solution candidates for \textit{Z340}.  It also supports rearranging \textit{Z340} with a few transposition variants.  \\

Other studies of computational techniques used to cryptanalyze and solve substitution ciphers in general include Delman's application of genetic algorithms (2004) \cite{delman2004genetic}, Forsyth and Safavi-Naini's use of simulated annealing (1993) \cite{forsyth1993automated}, Jakobsen's iterations of matrix manipulations (1995) \cite{jakobsen1995fast}, King and Bahler's algorithm for reducing sets of homophones (1993) \cite{king1993algorithmic}, Lasry's broad application of search metaheuristics (2018) \cite{lasry2018methodology}, and Spillman et al.'s use of genetic algorithms (1993) \cite{spillman1993use}.\\

Studies more focused on transposition ciphers in general include Al-Kazaz et al's use of a compression model to recognize correct decryptions \cite{al2016automatic}, Delman's application of genetic algorithms \cite{delman2004genetic}, Dimovski and Gligoroski's use of three optimization heuristics \cite{dimovski2003attacks}, Giddy and Safavi-Naini's application of simulated annealing to combinatorial optimization \cite{giddy1994automated}, Kullback's general solution for double transposition \cite{kullback1980general}, Lasry et al's divide-and-conquer approach to solving a double transposition challenge \cite{lasry2014solving}, Lasry's broad application of search metaheuristics \cite{lasry2018methodology}, Matthews' use of genetic algorithms \cite{matthews1993use}, and Song et al.'s development of a simulated annealing genetic algorithm \cite{song2008cryptanalysis}.\\

\paragraph{AZdecrypt:}  \textit{AZdecrypt} \cite{azdrepo} is a highly specialized cipher solver created by one of the present authors, Jarl Van Eycke, in 2014.  He developed the program because he wished to join the effort to break \textit{Z340}, and was inspired by \textit{ZKDecrypto} which became the basis for his program.  As Van Eycke explored different hypotheses about \textit{Z340}, he incorporated experimental efforts and ideas into \textit{AZdecrypt}.  It uses a highly tuned, optimized, and specialized simulated annealing algorithm.  On a modern 8-core CPU \textit{AZdecrypt} can solve up to 200 homophonic substitution ciphers per second with a 99\%+ solve rate \cite{eyckekeynote2021}.  Because of this speed and accuracy, \textit{AZdecrypt} set the stage for running tests on millions of transposition variations and test ciphers per day on personal computers, which accelerated the exploration of experimental hypotheses.  It can also solve a wide variety of transposition and polyalphabetic ciphers directly, and in various combinations.  \\

When testing candidate plaintexts, \textit{AZdecrypt} normally measures letter \textit{n}-gram statistics crossing the row boundaries of the cipher grid.  Geoffrey LaTurner introduced the idea of adapting \textit{AZdecrypt} so these measurements avoid crossing row boundaries instead \cite{oranchakblakeeycke2021}.  This feature was important for exploring the possibility that \textit{Z340} contained fragments of plaintext that were not in the correct order, or that such fragments resulted from errors in the original encipherment process.\\

\textit{AZdecrypt} uses letter \textit{n}-gram statistics to measure candidate solutions when its hill climber explores candidate keys.  Most letter \textit{n}-gram based solvers are limited to 5- or 6-grams, but because of its optimizations, \textit{AZdecrypt} can support up to 8-grams, which are ideal for codebreaking problems with longer keys or shorter available ciphertext.  Such large \textit{n}-grams impose huge memory requirements on personal computers (up to 256GB), but in 2019 programmer and researcher Louie Helm developed a modular method to load 8-gram stats into much less RAM (16GB and less) \cite{helm2019}.  In the two years leading up to \textit{Z340's} solution, Helm made many contributions to improve \textit{AZdecrypt's} code and handling of \textit{n}-gram statistics.\\

While many of the aforementioned tools and studies were effective for specific encipherment systems and configurations, we will see in subsequent sections that \textit{Z340} resisted cryptanalysis for so long because of its nature as a mixed or blended cryptosystem.

\raggedright
\section{Observations and measurements}\label{sec:observe}
\justifying

Online collaboration was an important component in the search for a solution to \textit{Z340}.  Many participants in this search made online contributions such as forum posts, blog entries, emails, social media posts, and videos.  A significant challenge in this effort was organizing the information that was generated from such a diverse community, which is composed of contributors with a wide range of skills, both applicable and non-applicable to cryptanalysis.  Much of the content generated by the community is speculative in nature, or otherwise does not directly apply to formal cryptanalysis of \textit{Z340}.\\

In 2012 Oranchak started a web page to collect factual observations made on the Zodiac ciphers over the years \cite{oranchakeoo2012}.  The primary goal in maintaining such a summary was to create a substantial factual basis from which to draw ideas and plans for further experimentation on \textit{Z340}.  Because the encipherment method was unknown, the sea of possibilities was vast, so it was hoped these observations would help guide the way.\\

Some of the noteworthy observations made since \textit{Z340} was published were as follows:\\

\begin{figure}[h]
  \centering
  \includegraphics[width=1.0\columnwidth]{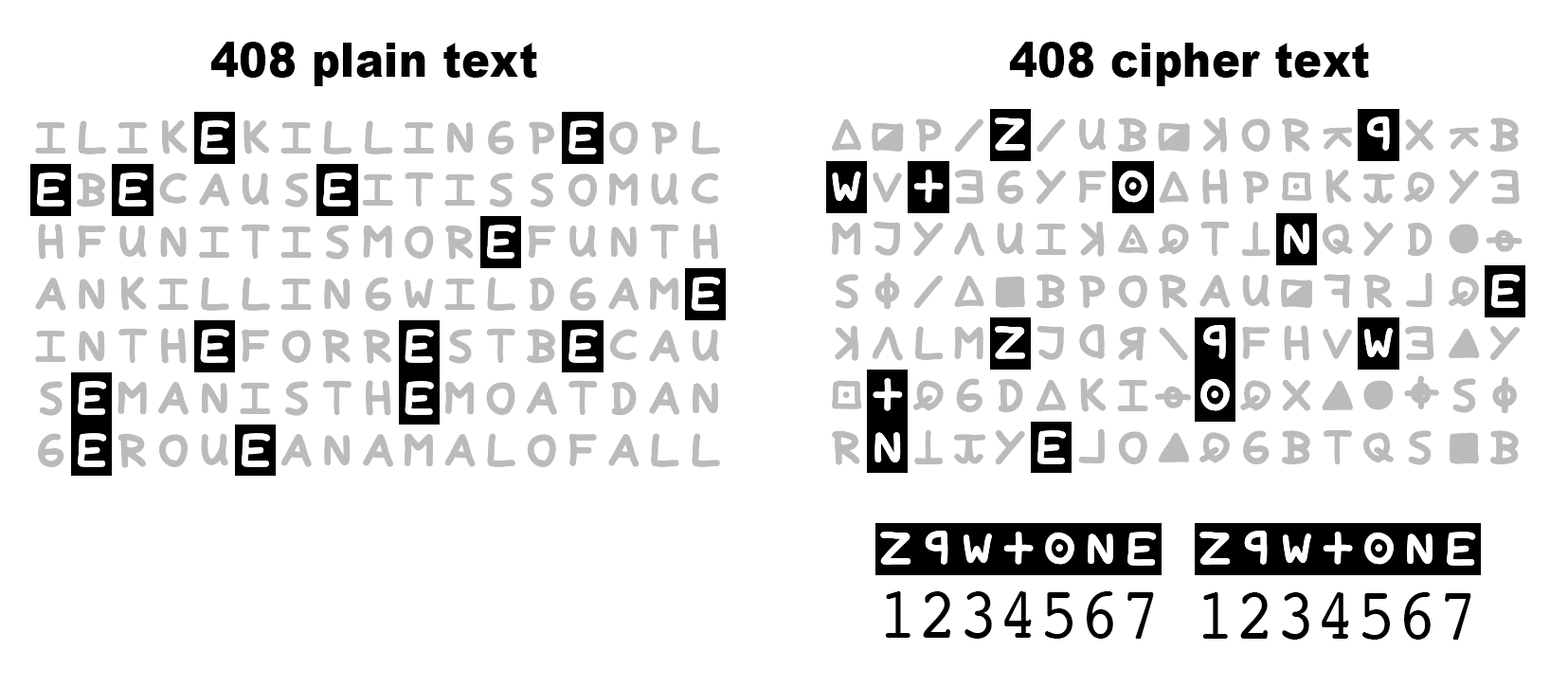} 
  \caption{\textit{Z408} cycling of homophones.  Fourteen plaintext Es are substituted with two identical seven-symbol sequences.}
  \label{fig:Z408_homophones}
\end{figure}

\begin{figure}[h]
  \centering
  \includegraphics[width=1.0\columnwidth]{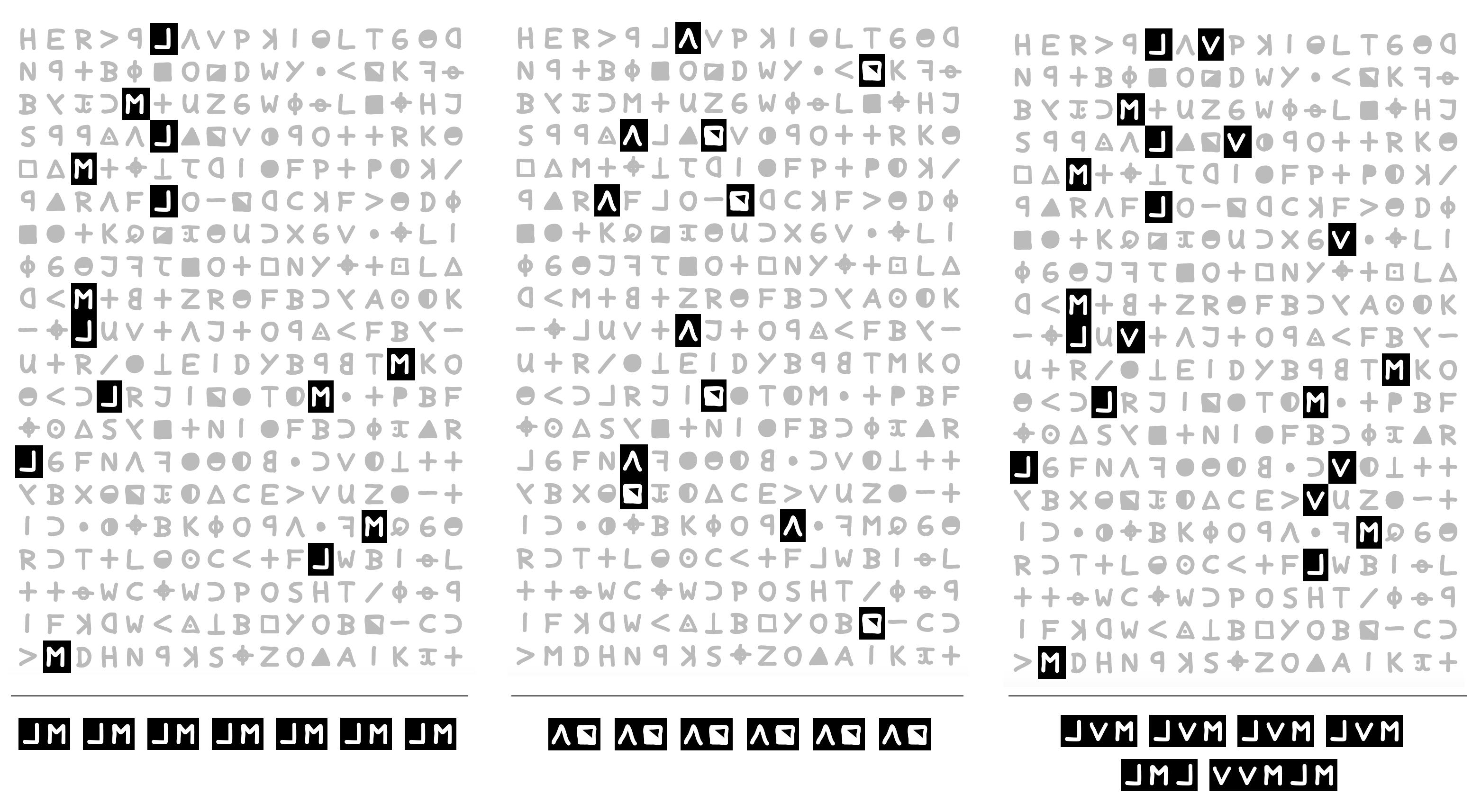} 
  \caption{Cycling of symbol groups in \textit{Z340}}
  \label{fig:Z340_homophones}
\end{figure}

\textit{Z408} shows a very strong homophone cycling behavior.  That is, for a given plaintext letter, substitution variants are sometimes selected from an orderly repeating sequence of multiple cipher symbols \textit{(example shown in Figure \ref{fig:Z408_homophones})}.  However, while \textit{Z340} showed a similar cycling behavior \textit{(Figure \ref{fig:Z340_homophones})}, it was not as significant, and was perhaps disrupted by the encipherment method.  \\

\textit{Z340} exhibits an unusual absence of repetitions of symbols appearing within the same horizontal window of length 17 (the width of the cipher grid is also 17).  This was taken to be evidence of some purposeful process Zodiac applied when creating the cipher.  A related observation is that there are nine rows that have zero repeating symbols \textit{(Figure \ref{fig:Z340_9lines})}, which is statistically significant compared to random scrambles of the cipher.  We speculate Zodiac may have intentionally obscured cycling homophones with this approach, and that it suggests the symbols were assigned in a normal horizontal reading order.\\

\begin{figure}[h]
  \centering
  \includegraphics[width=0.8\columnwidth]{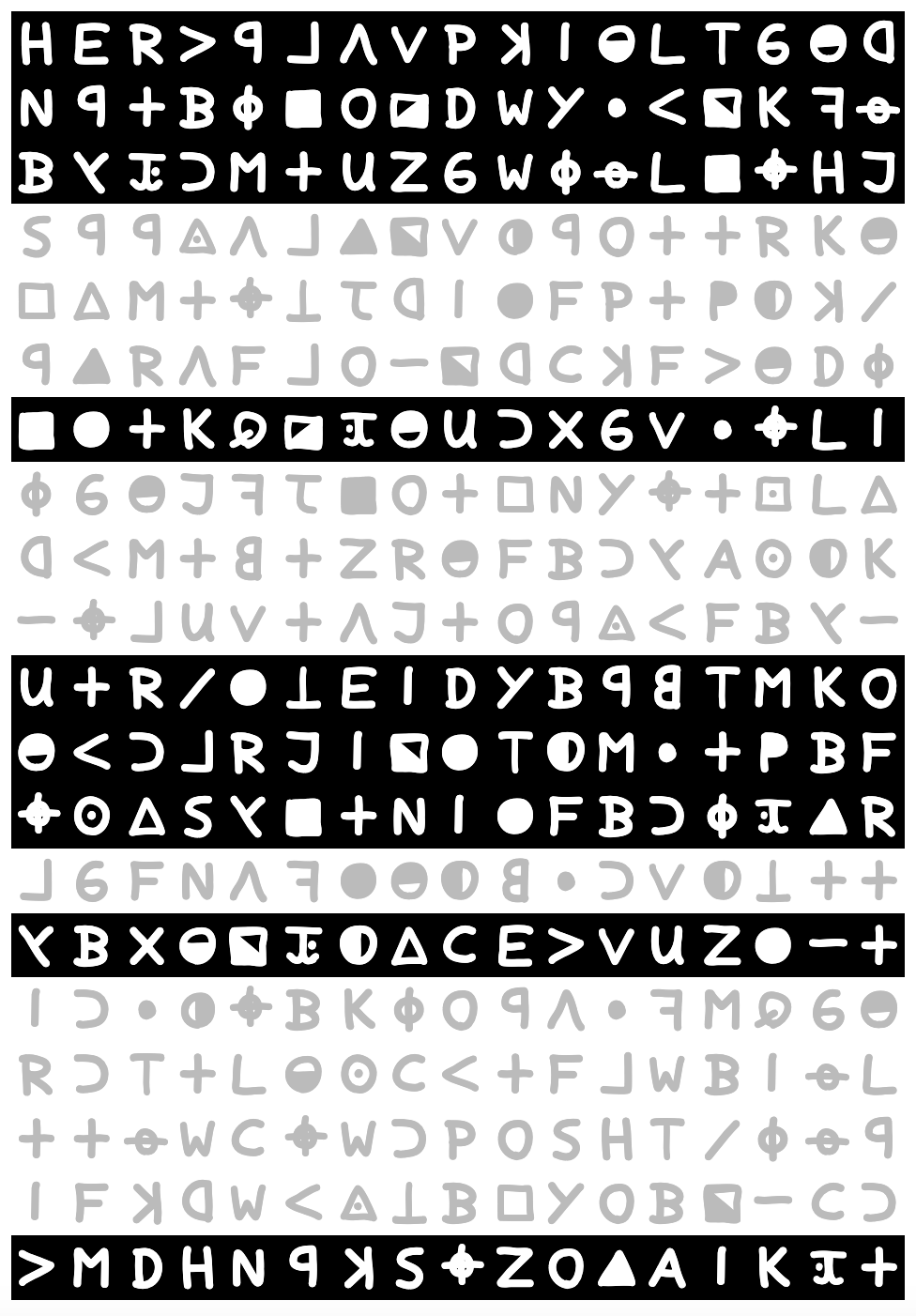} 
  \caption{Nine lines of \textit{Z340} have no repeating symbols.}
  \label{fig:Z340_9lines}
\end{figure}

\textit{Z408} has a significant amount of errors, which include spelling errors, encipherment mistakes, and an undeciphered section at the end (possible nulls or filler).  Whether these were intentional, or reflections of hurried or amateur effort, is unknown.\\

Twenty of the symbols in \textit{Z340} were not used in \textit{Z408}, and comprise nearly a fourth of the entire \textit{Z340} ciphertext.  The additional symbols compared to \textit{Z408} increases the keyspace size for \textit{Z340} by a factor of $26^9$, and combined with the shorter cipher length of \textit{Z340} compared to \textit{Z408}, makes \textit{Z340} a harder problem to solve.\\

\textit{Z340} has an unusual pattern of two ``backwards L'' shapes, composed of trigrams (three cipher symbols) that repeat in vertical and horizontal directions and are joined by a shared fourth cipher symbol \textit{(Figure \ref{fig:Z340_pivots})}.  The orientations of the two shapes are identical, making them more conspicuous.  One discovery of the shapes was posted in 2010 on an online Zodiac forum \cite{bentley2010}.  It was suspected that the shapes were indications of some purposeful encipherment process beyond homophonic substitution, particularly because of the rarity of such patterns appearing in both normal substitution cipher grids and random shuffles of \textit{Z340}.  Oranchak observed them appearing at a rate of approximately once per 237,000 random shuffles of \textit{Z340} \cite{oranchakeoo2012}.  It was also observed that such patterns are more likely to occur if the underlying plaintext is very repetitive.\\

\begin{figure}[h]
  \centering
  \includegraphics[width=0.75\columnwidth]{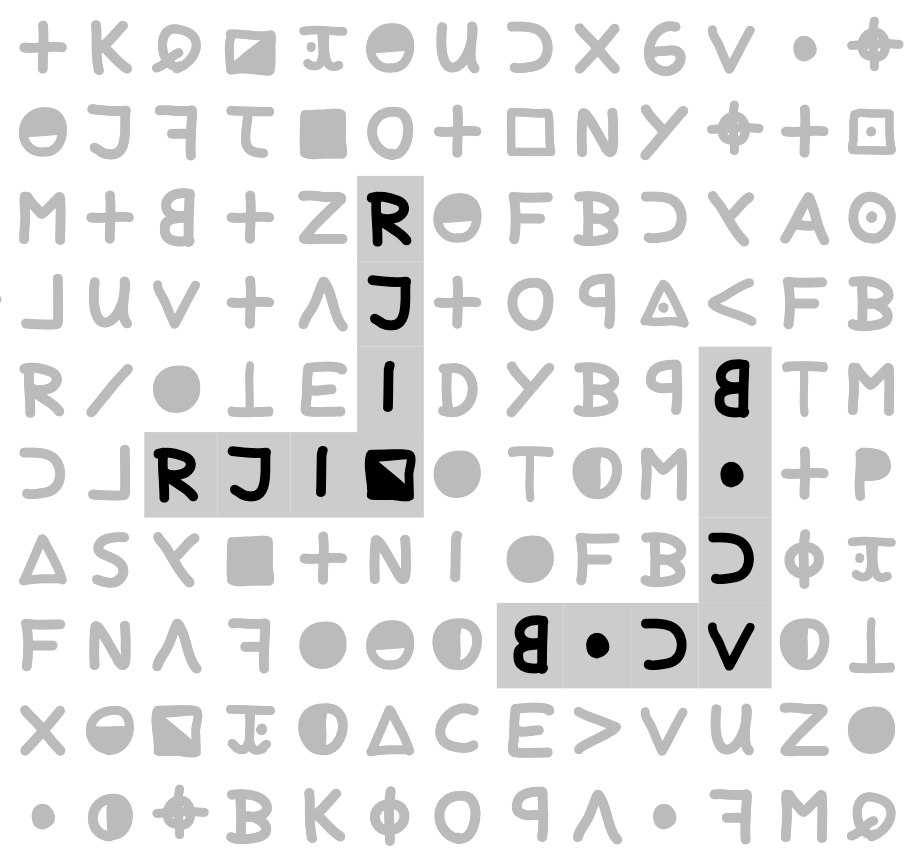} 
  \caption{Repeating backwards L-shaped patterns of intersecting trigrams in \textit{Z340}.  These patterns are sometimes referred to as ``pivots''.}
  \label{fig:Z340_pivots}
\end{figure}

\textit{Z340} has a set of symbols that are unusually clustered, or are unexpectedly absent from certain regions of the cipher grid \textit{(Figure \ref{fig:Z340_clustering})}.  This was interpreted as additional evidence of a real encipherment process and message, since the behavior deviates significantly from a random process.\\

\begin{figure}[h]
  \centering
  \includegraphics[width=0.8\columnwidth]{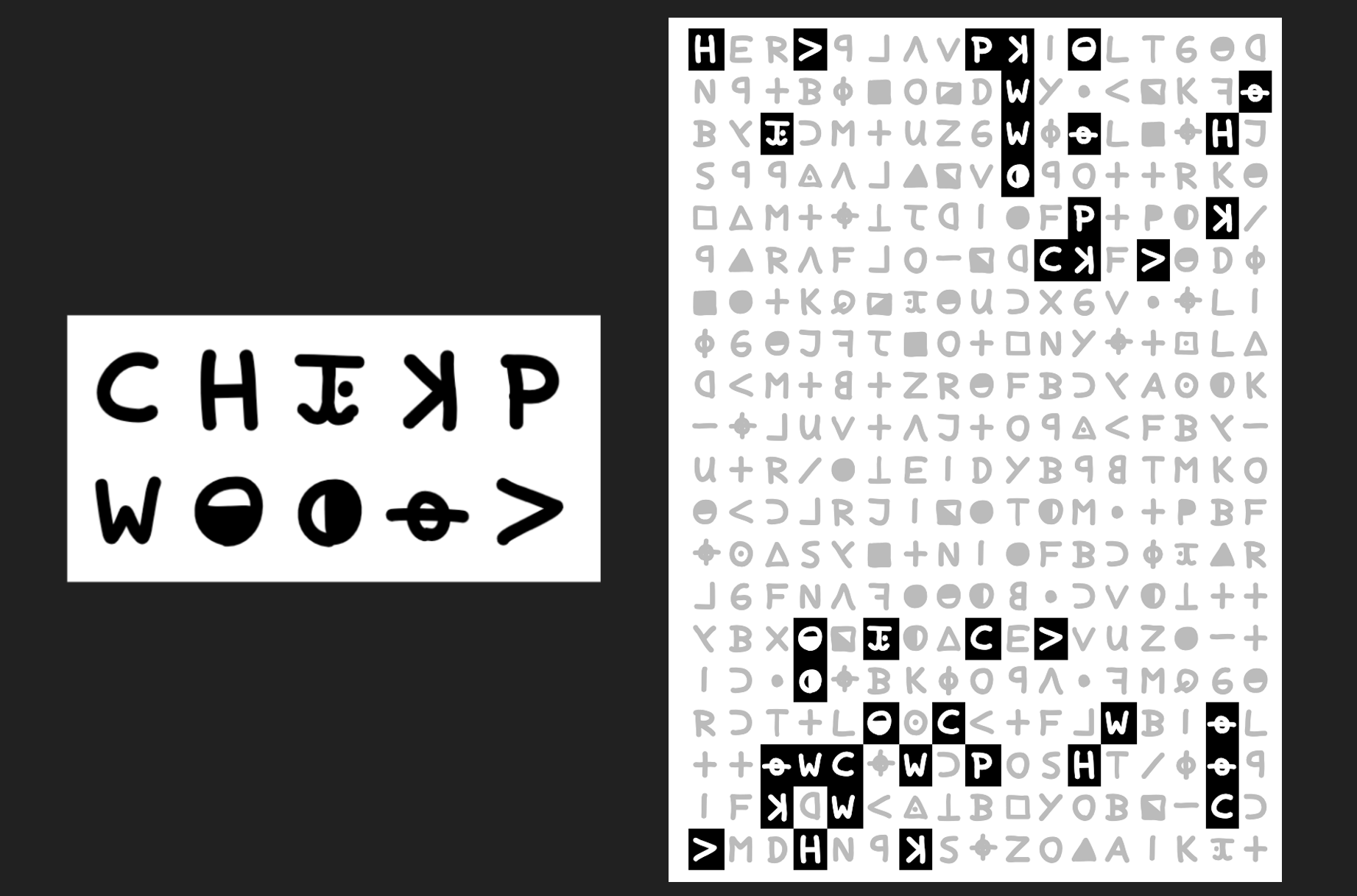} 
  \caption{Unusual clustering.  The symbols from the box on the left only appear in the highlighted positions in \textit{Z340}, avoiding a large section in the middle of the ciphertext.}
  \label{fig:Z340_clustering}
\end{figure}

Repeating trigrams are present in both \textit{Z408} and \textit{Z340}.  Their appearance in the latter \textit{(Figure \ref{fig:Z340_trigrams})} is suggestive of repetitive underlying plaintext, much as it is for the former \textit{(Figure \ref{fig:Z408_trigrams})}.  Additionally, one such repetition occurs in the same column positions but on different rows of the cipher grid.\\

\begin{figure}[h]
  \centering
  \includegraphics[width=0.8\columnwidth]{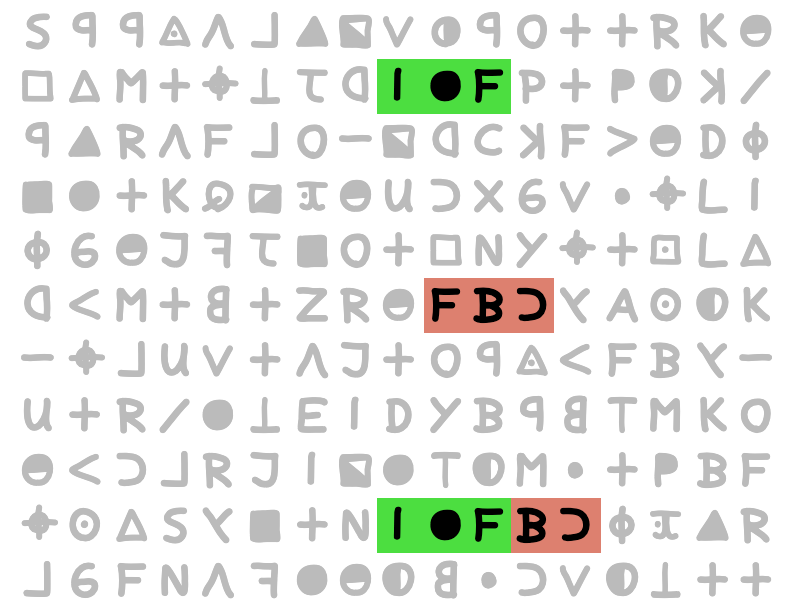} 
  \caption{Repeating trigrams in \textit{Z340}.  Note one pair appears in the same columns.}
  \label{fig:Z340_trigrams}
\end{figure}

\begin{figure}[h]
  \centering
  \includegraphics[width=0.6\columnwidth]{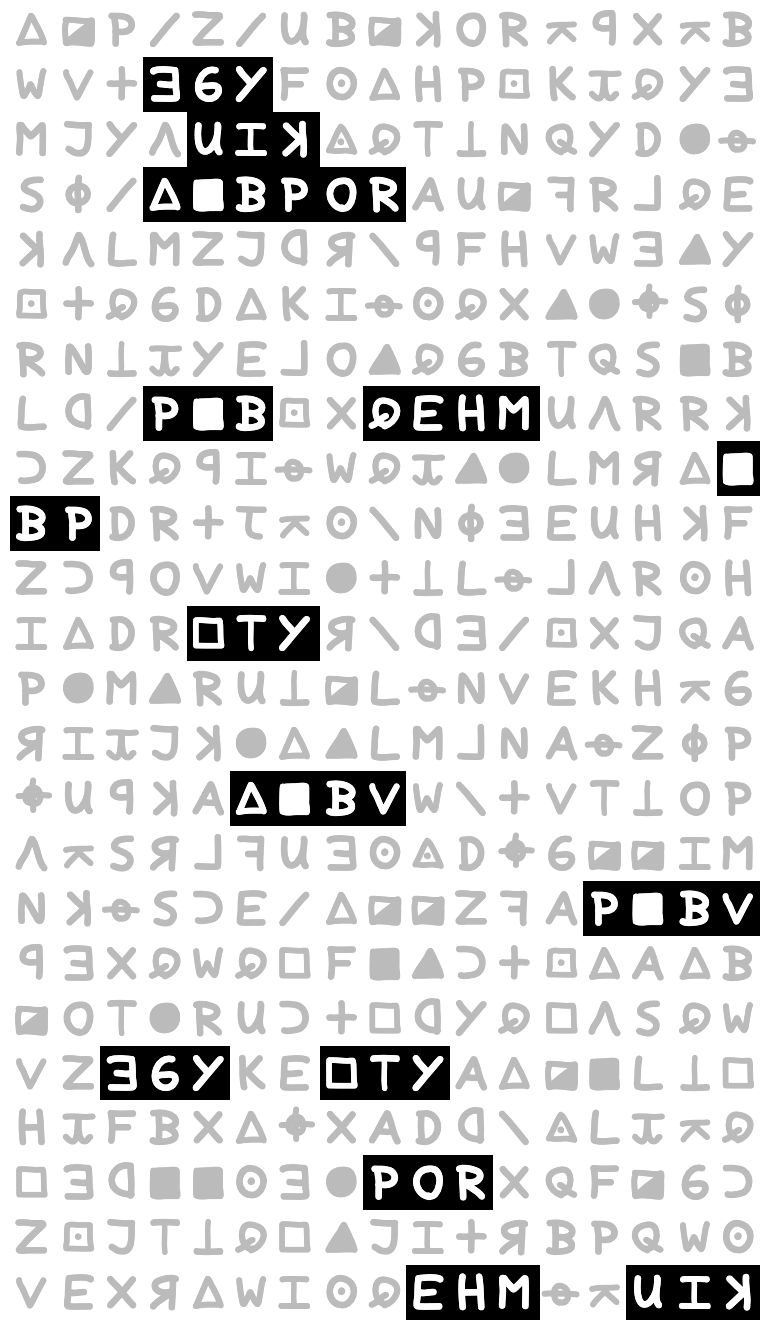} 
  \caption{Coverage of repeating trigrams in \textit{Z408}.}
  \label{fig:Z408_trigrams}
\end{figure}

A few readable words are visible in the ciphertext directly, such as ``HER'' at the very beginning, ``FBI'' read diagonally, and ``ZODIAC'' (but spelled ZODAIK with a solid triangle standing for the D) at the end.  Those and other examples are shown in \textit{Figure \ref{fig:Z340_words}}.\\

\begin{figure}[h]
  \centering
  \includegraphics[width=0.6\columnwidth]{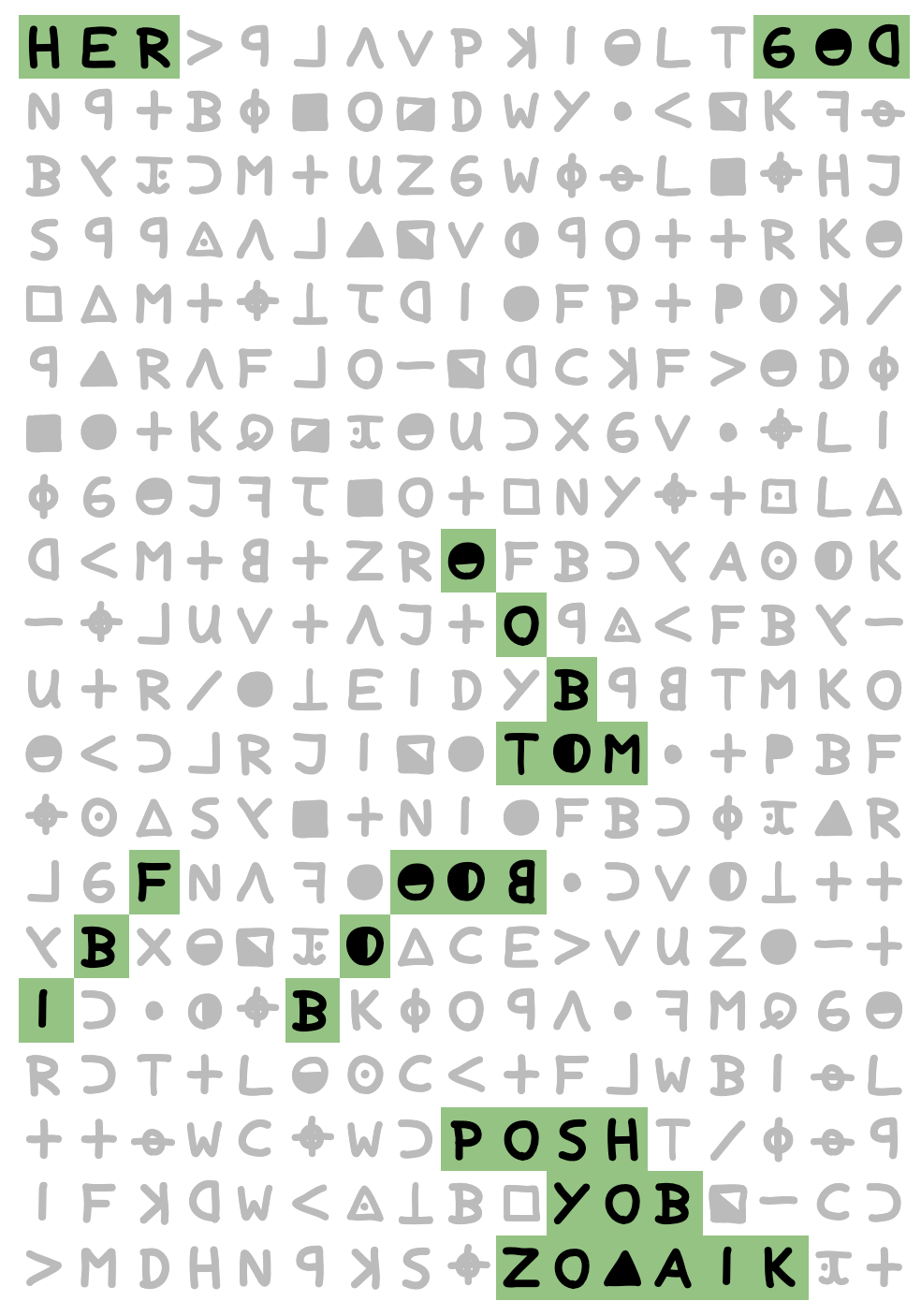} 
  \caption{Examples of ``words'' appearing directly in the ciphertext: HER, GOD, BOO, TOM, FBI, POSH, BOY, and ZODIAC (``ZODAIK'')}
  \label{fig:Z340_words}
\end{figure}

\textit{Z408} has a significant quantity of repeating bigrams, whereas \textit{Z340} has a quantity that does not significantly deviate from the expected result of a random process.\\

\begin{figure}[h]
  \centering
  \includegraphics[width=0.8\columnwidth]{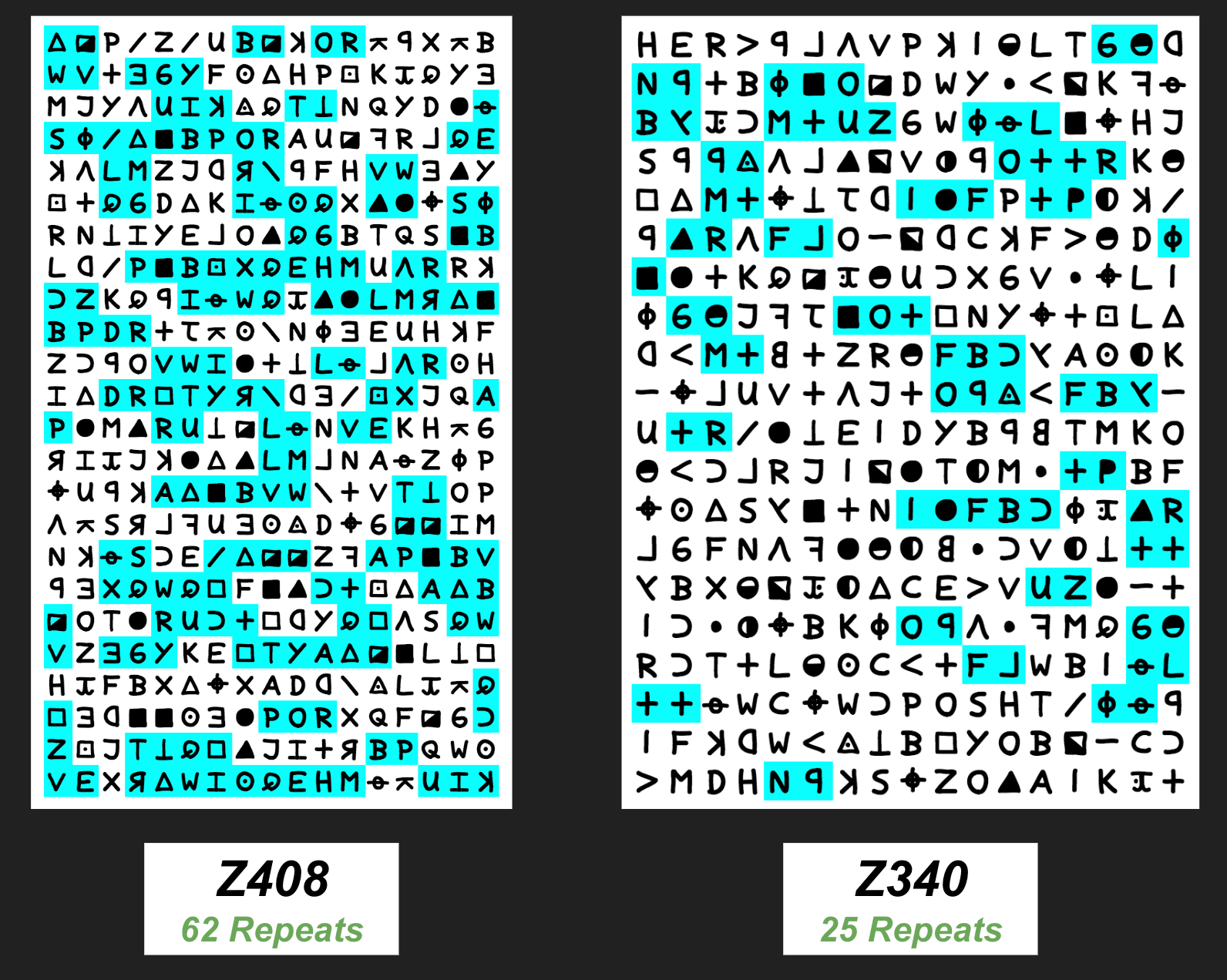} 
  \caption{Coverage of repeating bigrams in \textit{Z408} and \textit{Z340}.}
  \label{fig:bigrams}
\end{figure}

However, when repeating bigrams are measured periodically (that is, with fixed intervals between the symbols comprising the bigram), \textit{Z340} shows a dramatic increase in repeating bigrams at period 19, whereas \textit{Z408} shows no increases at any period other than 1.  This observation was suggestive of the possible use of transposition in \textit{Z340}, and became one of the most important guiding factors in the search for its solution.  This is discussed in more detail in \textit{Section~\ref{sec:bigrams}}.\\

\begin{figure}[h]
  \centering
  \includegraphics[width=0.8\columnwidth]{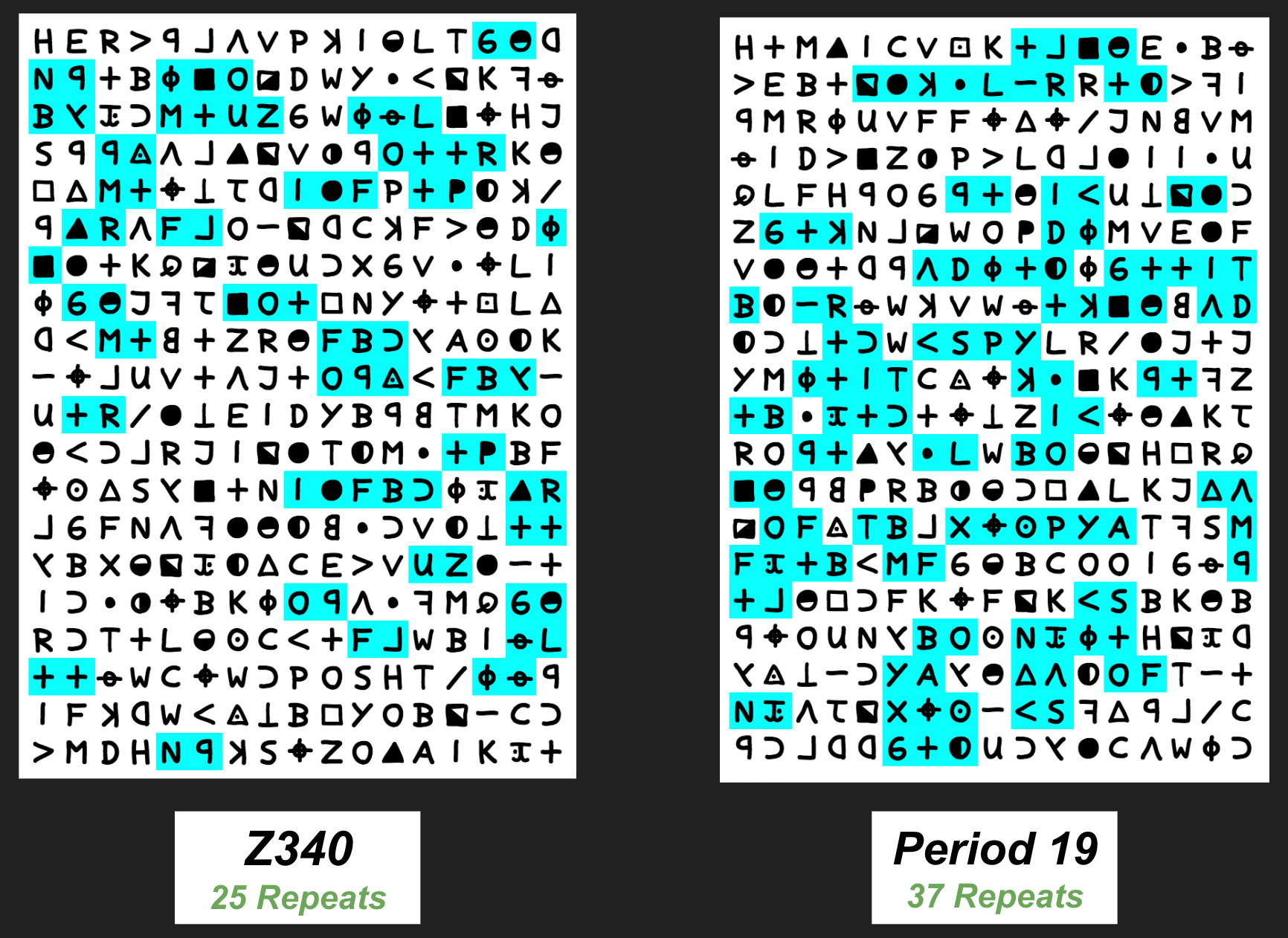} 
  \caption{Re-writing \textit{Z340} using Period 19 transposition dramatically increases the count of repeating bigrams.}
  \label{fig:Z340_period19}
\end{figure}

\begin{figure}[h]
  \centering
  \includegraphics[width=1.0\columnwidth]{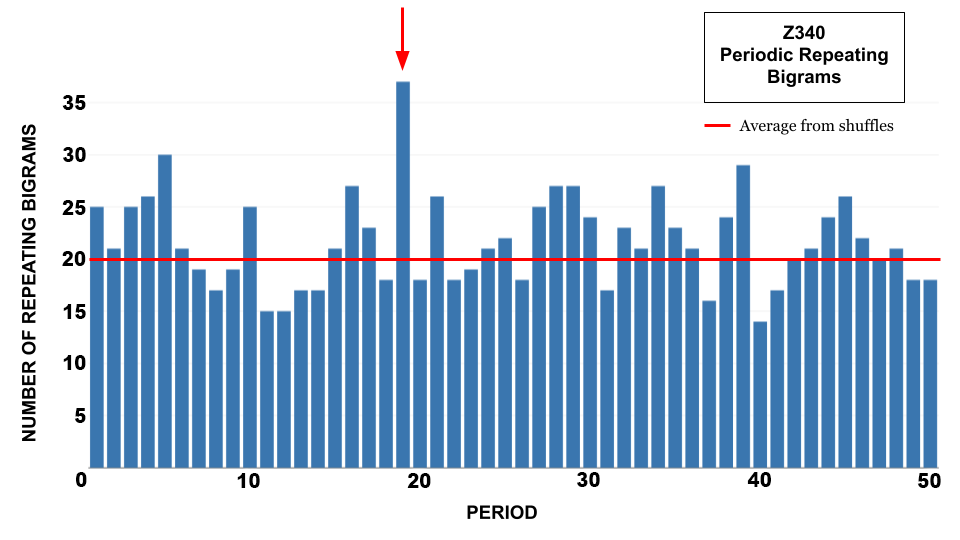} 
  \includegraphics[width=1.0\columnwidth]{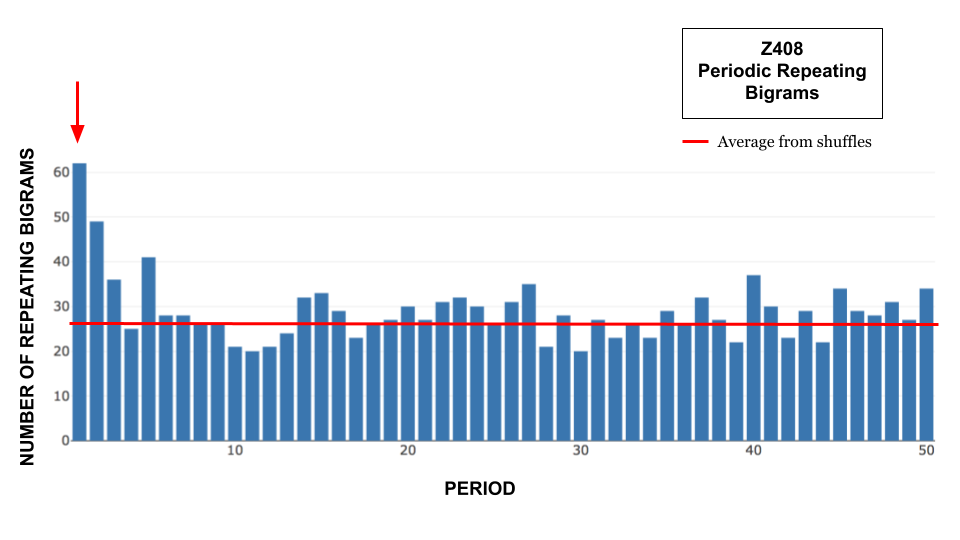} 
  \caption{\textit{Z340} and \textit{Z408} repeating bigram count at different periods.  For \textit{Z340} spike occurs at period 19, with a count of 37.  The average count from random shuffles of \textit{Z340} is 20.}
  \label{fig:Z340_periods}
\end{figure}

It is worth pointing out that if we expand the range of features we seek, then we may encounter them more often, and possibly \textit{post hoc} claim them to be significant.  Some features might be attributable to something like the \textit{look-elsewhere effect} \cite{dorigo2009}, a phenomenon where an observation that seems statistically significant actually arose by chance.  For instance, the aforementioned repeating backwards L-shaped patterns rarely appear during random shuffles of \textit{Z340's} ciphertext.  However, many random shuffles would also likely find other interesting and seemingly rare features, not just the two ``backwards L’s''.  Some of the aforementioned features may be the result of a coincidence-generating effect, which is difficult to measure without simulating the entire range of features that would be considered significant when observed.

\raggedright
\subsection{FBI observations and hypotheses}
\justifying

In 2009, FBI Cryptanalysis and Racketeering Records Unit (CRRU) chief Dan Olson pointed out several observations about \textit{Z340} \cite{olson2009} to Tom Voigt, who runs \textit{zodiackiller.com}:

\begin{itemize}
  \item Lines 1--3 and 11--13 show much higher randomness (fewer repeated characters) than other lines, suggestive of homophonic substitution.
  \item Lines 4, 10, 14, 17 and 18 have many repeated symbols.
  \item Randomness is far greater for rows than columns, seeming to point away from columnar or diagonal transposition.
  \item Randomness tests are roughly similar for \textit{Z408} vs \textit{Z340}.
  \item If \textit{Z340} lacked a message, or was evenly scrambled, then row and column randomness would be nearly identical, which was not observed.
\end{itemize}  

In that same year, Olson was featured on an episode of the History Channel show \textit{MysteryQuest} \cite{mq2009}, in which he describes a theory that \textit{Z340} might need to be split into two equal parts of 10 lines each, where only lines 1--3 and 11--13 are considered to contain a valid message.  These sections must be placed side by side in this scheme, so that line 1 and line 11 are read in a continuous stream, then lines 2 and 12, and finally lines 3 and 13.  The remaining lines may or may not contain a valid message.  

\raggedright
\section{Conjecture: \textit{Z340} is a transposition and a homophonic substitution cipher}
\justifying
In the many years since the November 1969 mailing of \textit{Z340}, attempts to solve \textit{Z340} had repeatedly failed.  Many of those attempts assumed \textit{Z340} was enciphered using a similar system as \textit{Z408}, which was a monoalphabetic, homophonic substitution cipher.\\

The FBI in late 1969 suspected a transposition might have been applied as an additional step in \textit{Z340}'s encipherment.  They referred to this as a ``combination cryptosystem'' and explored linear and route transposition on candidate plaintexts, with negative results \cite{fbivaultgd}.\\

Berg-Kirkpatrick et al. (2013) \cite{berg2013decipherment} argued that \textit{Z340} is not likely strictly homophonic, due to the ease at which \textit{Z340}-like test ciphers can be automatically solved.\\

Yi (2014) \cite{yicryptanalysis} suspected \textit{Z340} used a combination of substitution and columnar transposition, partly from examining previous efforts which had failed to crack \textit{Z340} under the homophonic substitution assumption.\\

Zhong (2016) \cite{zhong2016cryptanalysis} also made an attempt to crack \textit{Z340} using a combination of substitution and transposition, prompted by the conclusion in \cite{berg2013decipherment} that such a combination was likely employed.
In 2019, Juzek's cipher classification method \cite{juzek2019using} put \textit{Z340} in the ``advanced cipher'' or ``pseudo-cipher'' category, which further added to the ever-growing suspicion that \textit{Z340} was not just a substitution cipher.\\

Outside of the academic world, researchers in the online community were using tools such as \textit{ZKDecrypto} \cite{zkdrepo} and \textit{AZdecrypt} \cite{azdrepo} to crack \textit{Z340-like} test ciphers.  These test ciphers were designed to mimic many of \textit{Z340's} qualities and statistics, such as length, \textit{index of coincidence (IC)} \cite{friedman1987index} (a statistical measurement of the likelihood of repeated letters), quantity of homophone cycles, and repeating \textit{n}-grams.  The fact that software tools can solve these test ciphers easily but not \textit{Z340} rejects the idea that \textit{Z340} is a simple homophonic substitution cipher.\\

In 2019, Oranchak presented results of an experiment using a deep learning, multiple-category classification model to classify cipher types \cite{oranchaklcz13}.  This classification model predicted that \textit{Z340} is strongly homophonic but also exhibits qualities of other types of ciphers such as route and columnar transposition, and also pointed strongly against the hypothesis that \textit{Z340} is gibberish.\\

On the other hand, the results of all of the aforementioned efforts and software tools do not conclusively point at the combination cryptosystem hypothesis, because any of the following hypotheses could also have been true:

\begin{itemize}
  \item \textit{Z340} could have been enciphered with a message whose \textit{n}-gram frequency distribution is an outlier when compared to the usual \textit{n}-gram statistics used in many programmatic hill climbers.  For example, the plaintext could have been written without the use of the letter \textit{e}, the most common letter in the English language.  Such a text is known as a \textit{lipogram}, a kind of constrained writing.  Other examples of statistical outliers for the plaintext could be: a list of names, locations, or a long sequence of numbers spelled out in words.  However, in such cases, \textit{n}-gram statistics could be generated for use by a specialized run of a hill climber to attempt to exclude these hypotheses.
  \item \textit{Z340} could have been intended as a legitimate substitution cipher but some unknown error in its encipherment process yielded unrecoverable plaintext.  Specialized experiments are required to exclude this hypothesis, but since the error type and extent are unknown, it is difficult to cover all the possibilities. 
  \item \textit{Z340} could have been a hoaxed or gibberish message, intentionally devised to waste investigators' time, much to the presumed satisfaction of Zodiac.  This hypothesis can be explored by simulating the method of constructing the hoaxed or gibberish cipher, but again the type and extent of such a construction method are unknown and difficult to cover adequately.  Some examples of examinations of hoax hypotheses exist in the literature for other ciphers.  For example, in 1980 Gillogly \cite{gillogly1980beale} examined the seemingly encrypted Beale papers from the 19th century.  Only one has been solved and the others have received attention from literally generations of cryptanalysts.  Gillogly presented statistical evidence to argue that at least one of those ciphers is a hoax.  Mateer (2013) \cite{mateer2013cryptanalysis} also concluded that they are hoaxed, based on his cryptanalysis of the solved Beale cipher and its construction method.
\end{itemize}

Further indications of the possible use of transposition include these unusual qualities observed in \textit{Z340}'s ciphertext:

\begin{itemize} 
  \item Patterns in \textit{Z340}, such as the repeating ``backwards L'' shapes \textit{(Figure \ref{fig:Z340_pivots})}, were suggestive of patterns in underlying plaintext being oriented in multiple directions.
  \item Cyclic homophones in \textit{Z340} are somewhat suppressed \textit{(Figure \ref{fig:Z340_homophones})} when compared to \textit{Z408} \textit{(Figure \ref{fig:Z408_homophones})} and to other constructed test ciphers of length 340.  One explanation is that a transposition step may have disrupted the cycling phenomenon\footnote{However, this may be contradicted by the observation of minimal repeating symbols within windows of length 17 as described in \textit{Section \ref{sec:observe}}, which would be difficult to achieve if substitution were performed prior to transposition.}.
  \item The cipher contains a lower number of repeating bigrams than expected \textit{(Figure \ref{fig:bigrams})}.  Other test homophonic ciphers with known underlying plaintext messages, similar indices of coincidence, and similar cipher symbol distributions usually contain more repeating bigrams than observed in \textit{Z340}.
  \item The repeating bigram count spikes dramatically at period 19 \textit{(Figure \ref{fig:Z340_period19})} (see \textit{Section~\ref{sec:bigrams}}).  Normal bigrams are considered to be at period 1 (that is, taking \textit{Z340} as a one-dimensional linear sequence, the bigram's constituent symbols have a linear distance of exactly one position).  The constituent symbols of a bigram at period 19 have a one-dimensional distance of exactly 19 positions.  Several transposition cipher types, such as columnar transposition, are known to exhibit similar periodic qualities in their ciphertexts.  \textit{Figure \ref{fig:Z340_periods}} shows Z340's repeating bigram count over a range of considered periods, where it peaks at period 19.
\end{itemize}

We further argue that the following qualities pointed away from the gibberish or hoax hypotheses:

\begin{itemize}
  \item The repeating \textit{n}-gram patterns described above were suggestive of repeating patterns in a real underlying plaintext.
  \item Cyclic use of homophones was detected for \textit{Z340}, even though its presence was less significant than it was for \textit{Z408}.
  \item Unigram (single symbol) distribution of \textit{Z340} is consistent with the premise that Zodiac was actively avoiding premature reuse of his cipher symbols.  This ``spread out'' nature of unigrams is statistically significant, and occurs in a horizontal reading direction consistent with a horizontal substitution process \cite{oranchakeoo2012}.
  \item Successful decryption of \textit{Z408} revealed it was designed with a real message, so Zodiac had already proved his ability to create a functioning cipher.
  \item The space of possible transposition schemes, particularly ones that might have been invented by Zodiac, had not yet been explored exhaustively.  Numerous experiments were still possible, since the quantity of hypotheses was so high.
  \item Zodiac corrected a mistaken symbol in the top right region of ciphertext \textit{(Figure \ref{fig:Z340_K})}, suggesting that he cared enough about his creation to correct a possible mistake, but evidently did not want to suffer the tedium of redrafting the entire cipher grid.
\end{itemize} 

\begin{figure}[h]
  \centering
  \includegraphics[width=0.6\columnwidth]{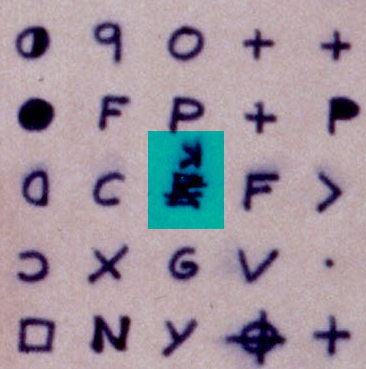} 
  \caption{The forward K symbol was corrected by Zodiac to a backwards K.}
  \label{fig:Z340_K}
\end{figure}

\begin{itemize}
  \item We hypothesize that Zodiac was motivated by attention, and wanted \textit{Z340} solved because it would get his name into the headlines not only after mailing it to the newspapers, but when a solution was found (which is what happened with \textit{Z408} \cite{ntokiv1969} \cite{murdercode1969}).  With a single functioning cipher, he could get his name in the news at least twice.  But a hoaxed cipher would have likely remained perpetually obscure.  Nevertheless, the notoriety of \textit{Z340} grew the longer it remained unsolved.
\end{itemize} 

\raggedright
\section{Attacking \textit{Z340}}
\justifying

\raggedright
\subsection{Crowdsourced research}
\justifying

The Internet has enabled crowdsourcing of research on the Zodiac Killer case for several decades.  The first noteworthy Zodiac-focused website, Tom Voigt's \textit{zodiackiller.com}, has been in operation since it was launched in 1998 \cite{monk2010}.  It became a focal point for case-related information, documents, and discussions.  The popularity of the case gradually led to more websites with their own communities of collaborating Zodiac researchers, such as Michael Butterfield's \textit{zodiackillerfacts.com} launched in 2007 \cite{butterfield2007}, and Mike Morford's \textit{zodiackiller.net} (formerly \textit{zodiackillersite.com} and \textit{zodiackillersite.forummotion.com}) in 2010 \cite{morford2010}.  Such sites hosted many active conversations with research, information, opinions, speculations, stories and debates about the Zodiac case and ciphers.\\

Analysis of \textit{Z340} benefited from collecting factual information about the cipher, its context in the Zodiac case, statistical analyses of the ciphertext, and other cryptanalytic observations.  In those early days of online collaboration on the case, very little research material pertaining to Zodiac's ciphers was available in the academic literature.  Some useful factual details, cryptanalytic observations, and experimental results were found in online Zodiac-themed forums, but these were difficult to identify due to the breadth of conversations.  For example, many of the conversations were speculative and/or argumentative in nature, so useful factual details got lost in the noise.  This motivated Oranchak to collate and summarize factual observations on the website \textit{zodiackillerciphers.com}, as an ``Encyclopedia of Observations'' \cite{oranchakeoo2012}.  As users on online forums discovered interesting details about the ciphers, Oranchak endeavored to preserve them on the site. This allowed other researchers to use the information as a foundational source of factual details to build upon. \\

The online community drew a wide variety of people eager to crack the case and its ciphers, which brought many different approaches and viewpoints.  \textit{Z340}'s notoriety in particular attracted the interest of people from many walks of life and with a wide range of abilities pertaining to codebreaking.  Some people had no codebreaking experience and approached the ciphers with intuition, speculation, pattern-seeking, sometimes flawed analysis, and confirmation bias (a tendency to accept validating information but reject contradicting information).  Some also became aggressively convinced their favored Zodiac suspect was the correct one and the ciphers reflected their suspect's name or other details\footnote{Some people also derived or generated information pointing to their suspect from Zodiac's non-enciphered letters, or from the factual case details themselves such as locations and dates.}.  Others brought more rational and/or technical skills to the investigations. Taken as a whole, the combined efforts of the entire group of enthusiastic investigators led to crowdsourced knowledge about \textit{Z340}, providing many useful details that made the discovery of its solution possible, as long as researchers could navigate around numerous unusable ideas and information.\\

A number of researchers investigating \textit{Z340} suspected it may have used a different encipherment system than \textit{Z408}, or a similar system with additional steps.  In the previous section we summarized observations of \textit{Z340} that led to these suspicions.  In particular, the repeating ``backwards L'' shapes were suggestive of possible use of multiple reading directions \cite{bentley2010}.  But more importantly, periodic qualities were observed when investigating repeating bigram counts.  \\

In the summer of 2015, both Van Eycke \cite{eyckes32015} and a Zodiac forum user using the alias \textit{daikon} \cite{daikon2015} independently made the same observation that bigram counts increased at period 19 (sometimes referred to as ``a distance of 19'').  Both initially thought that columnar transposition was involved in the encipherment process, but experiments based on that hypothesis had negative results.  Additionally, other encipherment schemes were considered, such as zigzag routes, spiral routes, splitting the grid into quadrants or subgrids, diagonal routes, reflections, flips, rotations, interleaving columns, interleaving rows, polyalphabetic schemes, and more.  One hypothesis was that Zodiac may have made some encipherment mistakes during transposition, such as accidentally skipping or duplicating parts of the message during transcription or transposition, which would introduce significant errors in decryption.  None of these hit upon the exact method used to encipher \textit{Z340}, but some of the manipulations of ciphertext resulted in even higher bigram counts than that observed for period 19 bigrams.  The responsiveness of bigram counts to simple manipulations of the ciphertext was an important clue that motivated the authors' further experiments \cite{oranchakeoo2012}.

\raggedright
\subsection{Repeating bigram counts}\label{sec:bigrams}
\justifying

Let us define a \textit{bigram} as an ordered pair of ciphertext elements (symbols) or plaintext elements (letters).  For example, \textit{Z340}'s ciphertext begins with the bigram $(c_1, c_2) = (H, E)$.  \textit{Z408}'s plaintext begins with the bigram $(p_1, p_2) = (I, L)$.  \\

\begin{algorithm}
  \caption{Period p algorithm}\label{period_p}
  \begin{algorithmic}[1]
    \Function{bigrams}{$c,p,n$}\Comment{Generate \textit{B}, the sequence of period \textit{p} bigrams from cipher \textit{c} of length \textit{n}.  Note: \textit{p} must be $\leq n/2$.}
      \State{$B\gets \{\}$}
      \State{$i\gets 1$}
      \While{$i \leq p$}
        \State{$j\gets i$}
        \While{$j \leq n-p$}
          \State{$B\gets B \cup \{(c_j, c_{j+p})\}$} unless $j+p > n$
          \State{$j\gets j+p$}
        \EndWhile\label{w2}
        \State{$i\gets i+1$}
      \EndWhile\label{w1}
      \State \textbf{return} $B$
    \EndFunction
  \end{algorithmic}
\end{algorithm}

Consider the linear distance between elements of the bigram.  In those examples, the distance is one since the bigram elements are adjacent to each other.  So we define those as \textit{period 1} bigrams.  But we also consider bigrams more generally, at any distance or period. \\

A ciphertext \textit{c} consists of a sequence of ciphertext units, $(c_1, c_2, c_3, \ldots , c_n)$, where \textit{n} is the length of the ciphertext.  The full sequence of period 1 bigrams, expressed as a sequence of ordered pairs, is: $\{(c_1, c_2), (c_2, c_3), \ldots , (c_{n-1}, c_n)\}$.  There are several methods to produce a bigram sequence for any period.  One way is given by the \textit{BIGRAMS} function shown in \textit{Algorithm \ref{period_p}}, which successively samples bigrams from the ciphertext at the given period.  We analyze this method in more detail in \textit{Section~\ref{sec:transpo}}, where it is referred to as \textit{pseudo-period} transposition.\\

A bigram repeats if the same ordered pair appears in multiple locations in the sequence.  Let us define the \textit{bigram count} for a bigram $b$ as $c(b)$, and set it to the number of times $b$ appears in the bigram sequence.  Then define the \textit{repeat count} as $r(b) = c(b) - 1$.  For example, if $b$ appears three times in the ciphertext, then $c(b) = 3$, and $r(b) = 2$.\\

We measure the total repeating bigrams for a ciphertext by summing $r(b)$ for every distinct bigram \textit{b} in the sequence.  For \textit{Z408} (period 1), this total is 62, which is 7.52 standard deviations from the mean observed for random shuffles of \textit{Z408}'s ciphertext.  By contrast, \textit{Z340} (period 1) has a total of 25, only 1.36 standard deviations from the mean observed for random shuffles of \textit{Z340}'s ciphertext.  In an experiment with 1,000 artificially constructed homophonic ciphertexts of length 340, with known underlying plaintexts sampled from English-language corpora, the average number of repeating bigrams observed was 34.5 \cite{oranchaklcz2} \textit{(Figure \ref{fig:bigram_shuffles})}.

\begin{figure}[h]
  \centering
  \includegraphics[width=0.95\columnwidth]{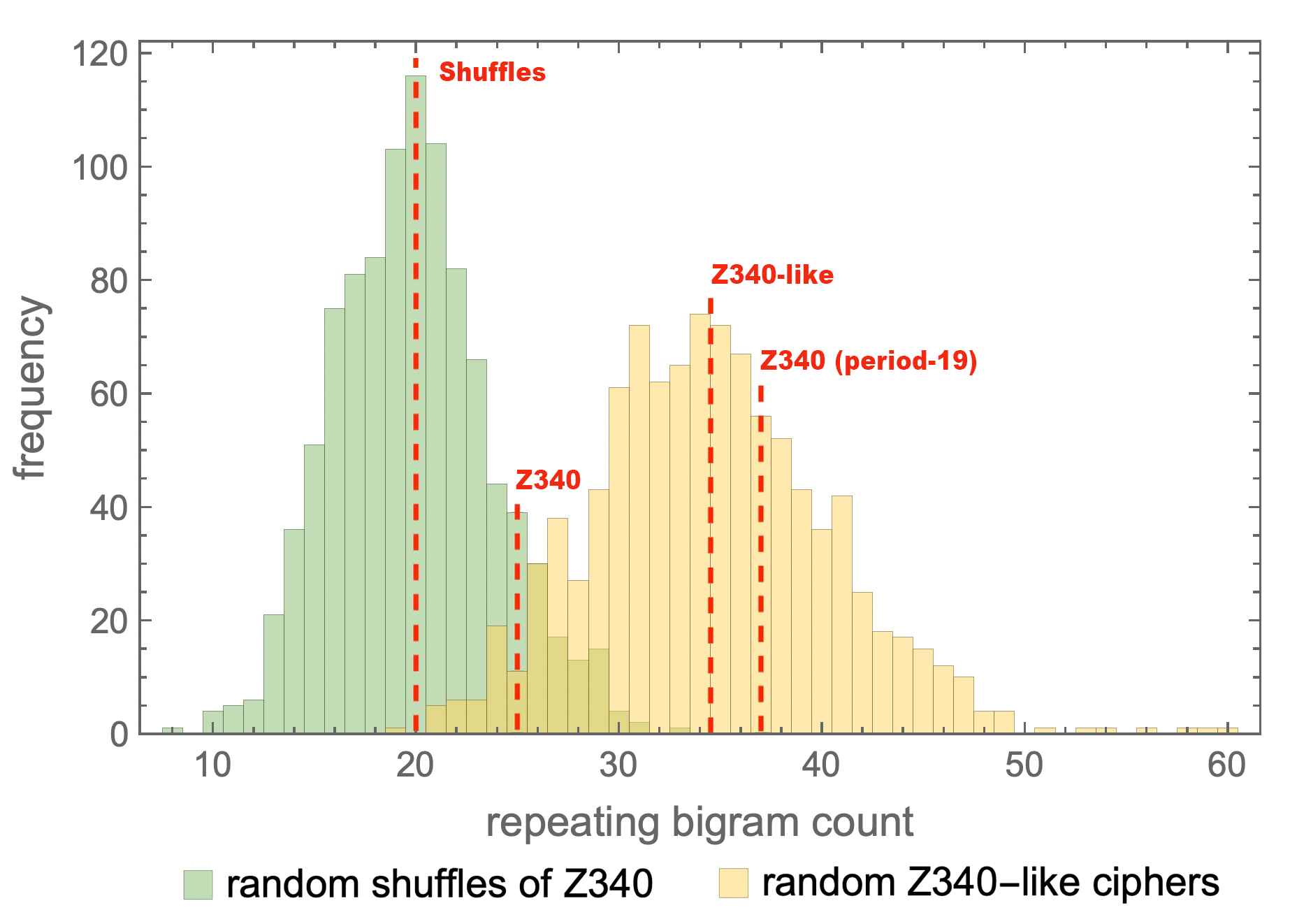} 
  \caption{Repeating bigrams in shuffles of \textit{Z340} (left histogram) vs. \textit{Z340}-like ciphertexts (right histogram).  Dotted lines from left to right mark:  Mean of repeats for shuffles of \textit{Z340} (20), repeats for \textit{Z340} period 1 (25), mean of repeats for \textit{Z340}-like ciphers (34.5), and repeats for \textit{Z340} period 19 (37).}
  \label{fig:bigram_shuffles}
\end{figure}

But for \textit{Z340}, a total of 34 different periods in the range [2, 170] result in repeating bigram counts that meet or exceed that of period 1.  The highest occurs at period 19, for which there are 37 repeating bigrams\footnote{The next highest peaks are 30 at period 5, 29 at period 39, and 28 at period 85.}.  This led to suspicion that a transposition scheme might be involved in the construction of \textit{Z340}.  By contrast, there is no period in the range [2, 204] such that bigram counts for the \textit{Z408} exceed its count for period 1, and \textit{Z408} is known to lack a transposition step.\\

Further transposition variations raise the total repeating bigram counts \cite{oranchakeoo2012}.  For example, horizontally mirroring the ciphertext prior to measuring the period 15 bigram count raises the count from 37 to 41.  Column period 2 combined with linear period 18 produces 44 repeating bigrams.  Various row-wise and column-wise offsets and periodic operations lead to between 41 and 45 bigrams.  Moving the entire last column immediately before the first column raises period 19 repeats from 37 to 45.  These increases in observed repeating bigrams by applying simple manipulations were suspected to be due to getting closer to the true transposition scheme of \textit{Z340}.\\

Other noteworthy ciphers demonstrate periodical behavior due to transposition-related encipherment \cite{oranchakblakeeycke2021}: 

\begin{itemize}
  \item Feynman 1 \cite{pelling2013} (period 5)
  \item Kryptos 3 \cite{bauer2016} (periods 8 and 85)
  \item Langrenus \cite{hope2021} (period 3)
\end{itemize}

Some periodic repeating bigram behavior can be attributed to redundancy in language.  For example, Van Eycke shows that the average plaintext of size 340 will show repeating bigram counts for periods 1, 2 and 3 that are higher than the observed counts for random shuffles of plaintext \cite{oranchakblakeeycke2021}.  \\

The backwards L shapes (pivots) (see \textit{Figure \ref{fig:Z340_pivots}}) contributed to the periodic bigram count, but at period 39 instead of 19, because they consist of bigrams that repeat at period 39. 

\raggedright
\subsection{Transposition investigations}\label{sec:transpo}
\justifying

The periodic nature of \textit{Z340} became the focus of many threads on the Zodiac Killer online forums, particularly Mike Morford's (defunct) site \textit{zodiackillersite.com} \cite{morford2010}.  For several years prior to discovery of the solution, there were many cryptanalytic investigations and experiments conducted by forum participants.  Activities included:

\begin{itemize}
  \item Statistical analysis and cryptanalysis.
  \item Attempts to determine if specific observations were statistically significant.
  \item Rearranging \textit{Z340} using a wide variety of transposition rules.
  \item Feeding the transposed versions of \textit{Z340} into \textit{AZdecrypt} and other solver programs.
  \item Generating test ciphers under various hypotheses, then cryptanalyzing them to determine if \textit{Z340} could fall under the same hypotheses.
  \item Exploring other hypotheses such as:
  \begin{itemize}
    \item Encipherment mistakes or interruptions such as transposition misalignments, skipped symbols (nulls), or extra inserted symbols (filler).
    \item Polyalphabetism
    \item Wildcards (that is, certain cipher symbols may be allowed to stand for any letter, or any selection from a fixed set of letters)
    \item Variations of inscription (how the cipher is written) and transcription (how the cipher is read)
    \item Quadrants or subgrids
  \end{itemize}
  \item Connecting observations of periodic phenomena with other aspects of \textit{Z340}.
  \item Investigating \textit{Z340}'s somewhat suppressed homophonic cycling behavior, and determining if any manipulations could reverse this partially muted effect.
  \item Determining what role the pivots and other unusual features might have in the encipherment scheme.
  \item Incorporating transposition schemes into \textit{AZdecrypt} directly so it can explore a space of transpositions.
  \item Sharing and analyzing solution proposals.
\end{itemize}

On May 17, 2020, Van Eycke proposed a transposition hypothesis for \textit{Z340} \textit{(Figure \ref{fig:jarl_transpo})} \cite{eyckezmtr2020}, which, in hindsight, closely resembled the correct hypothesis. It also significantly increased \textit{Z340's} repeating bigram count from 25 to 42.  However, due to errors in Zodiac's encryption, running this transposed version of \textit{Z340} through \textit{AZdecrypt} at the time did not yield a positive result.  Van Eycke was on the right track with his significant discovery, and had essentially found the needle in the haystack without realizing it.\\

\begin{figure}[h]
  \centering
  \includegraphics[width=0.85\columnwidth]{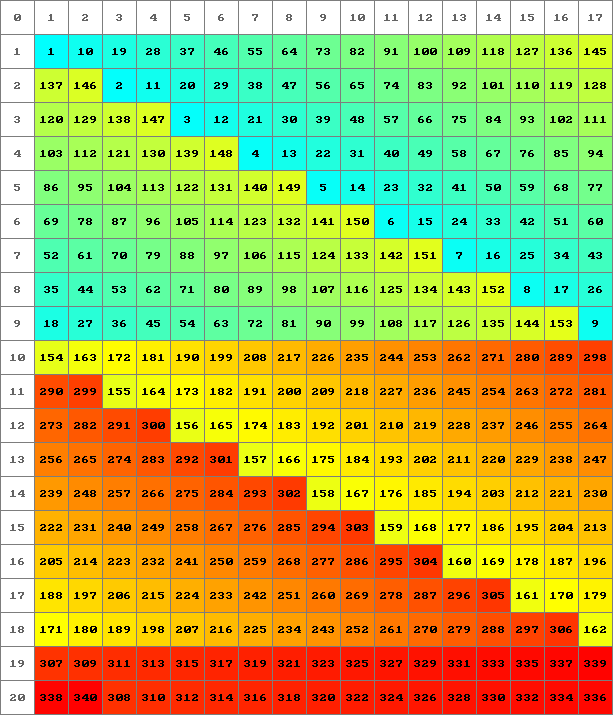} 
  \caption{Van Eycke's proposed transposition matrix for \textit{Z340} \cite{eyckezmtr2020}.  Cell numbers represent the reading order for this transposition.}
  \label{fig:jarl_transpo}
\end{figure}

In January 2019, Australian applied mathematician Sam Blake reached out to David Oranchak after watching his YouTube talk titled \textit{``The Zodiac Ciphers---What do we know, and when do we stop trying to solve them?''} \cite{oranchaktzc2015}. Blake was intrigued by Oranchak's discussion of the statistically unlikely spike in repeating bigrams at period 19, as observed by \textit{daikon} (a \textit{zodiackillersite.com} forum user) \cite{daikon2015} and Jarl Van Eycke \cite{eyckes32015}.  Blake noticed the period 19 discussed was only a \textit{pseudo}-period 19, as it used period 18 when wrapping around vertically. In \textit{Figures \ref{fig:pseudo_period_19} and \ref{fig:period_19}} we illustrate the differences between period 19 and pseudo-period 19 transpositions of the indices of \textit{Z340}. \\

\begin{figure}[h]
  \centering
  \includegraphics[width=0.85\columnwidth]{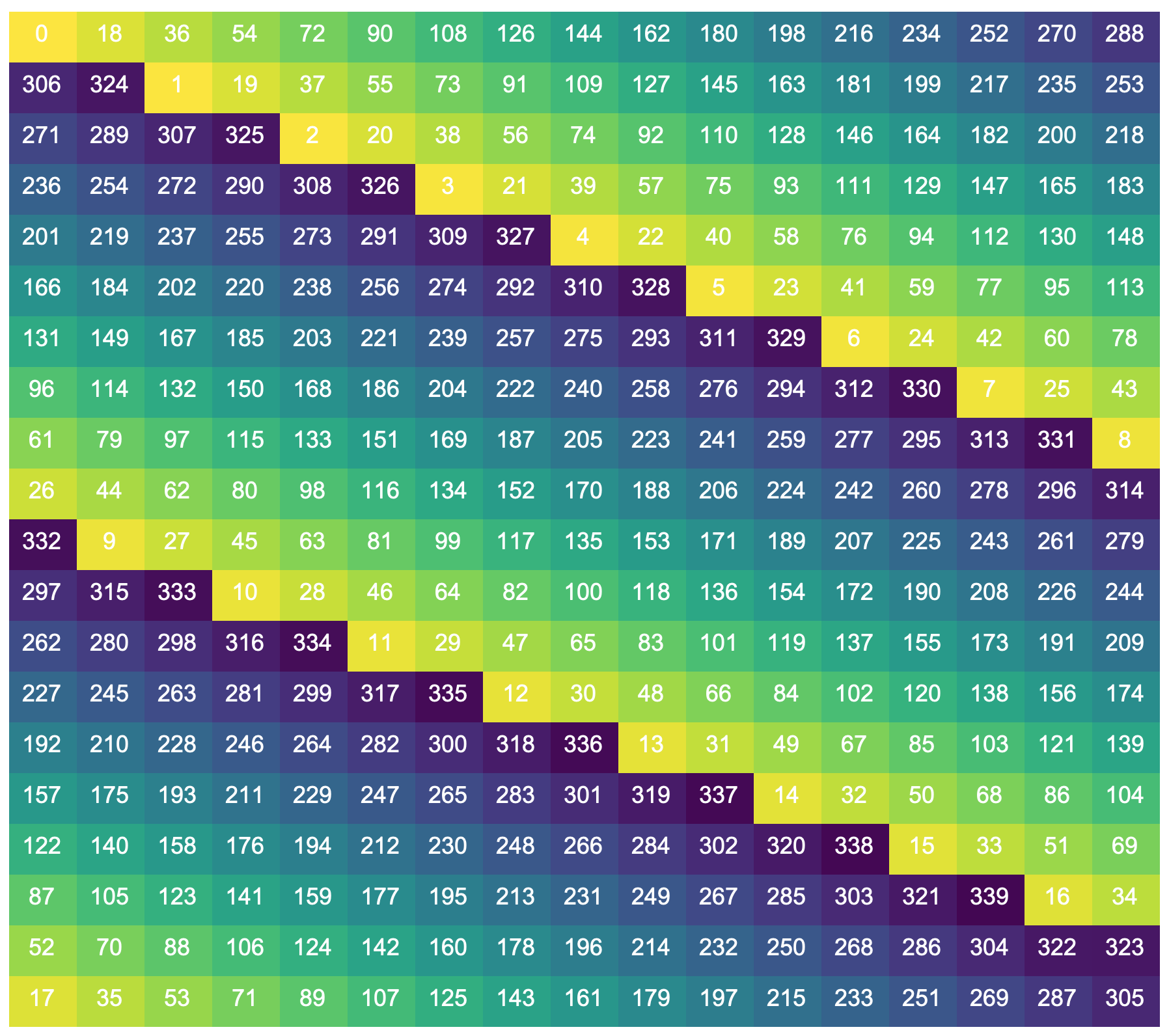} 
  \caption{Pseudo-period 19: period 18 when wrapping around vertically and period 19 otherwise.}
  \label{fig:pseudo_period_19}
\end{figure}

\begin{figure}[h]
  \centering
  \includegraphics[width=0.85\columnwidth]{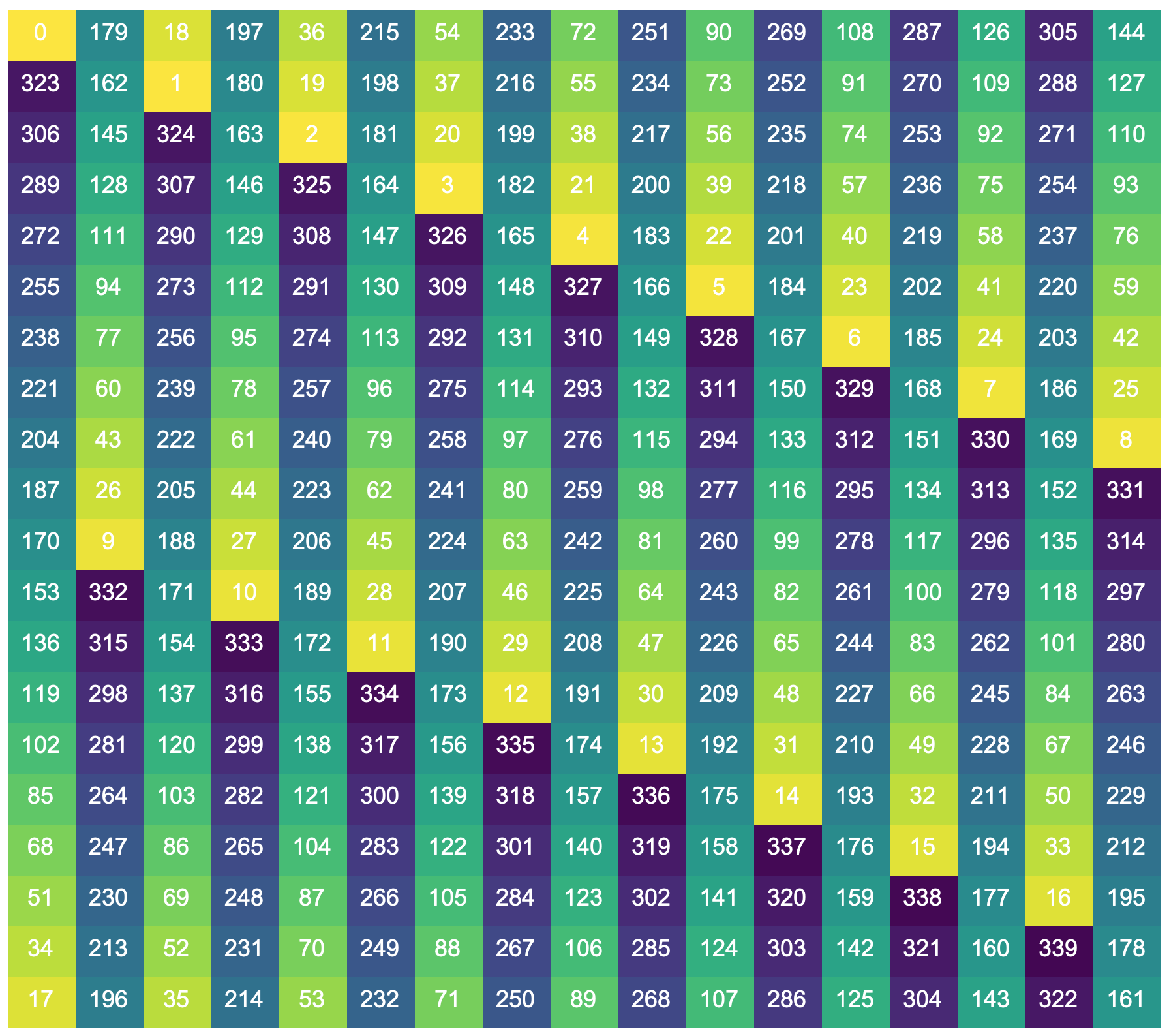} 
  \caption{Period 19: a period 19 decimation of 340.}
  \label{fig:period_19}
\end{figure}

Both of these transpositions take similar \textit{diagonals} through the two-dimensional representation of the cipher when viewed as a 20 by 17 array. Blake wanted to explore similar transpositions; one little-known way to enumerate a two-dimensional array is via a doubly-periodic \textit{proper decimation}. That is, a proper $(n,m)$-decimation of an $N$ by $M$ array, \textbf{C} is given by
\begin{gather*}
  \text{dec}_{n,m}\left(\textbf{C}_k\right) = \textbf{C}_{n\,k \text{ mod } N, m\,k \text{ mod } M}\\
  k \in [0, N \times M - 1],
\end{gather*}

providing $n$ is coprime to $N$ and $m$ is coprime to $M$. (If these conditions are not satisfied, then the decimation is improper and some indices will be missed.) While this may appear complex, geometrically this is just $n$-down, $m$-across, and wrapping around periodically both horizontally and vertically. \\

The $(1,2)$-decimation of a 20 by 17 array takes similar \textit{diagonals} through the cipher as the pseudo-period 19 and period 19 transpositions \textit{(Figure \ref{fig:12_decimation})}. \\

\begin{figure}[h]
  \centering
  \includegraphics[width=0.85\columnwidth]{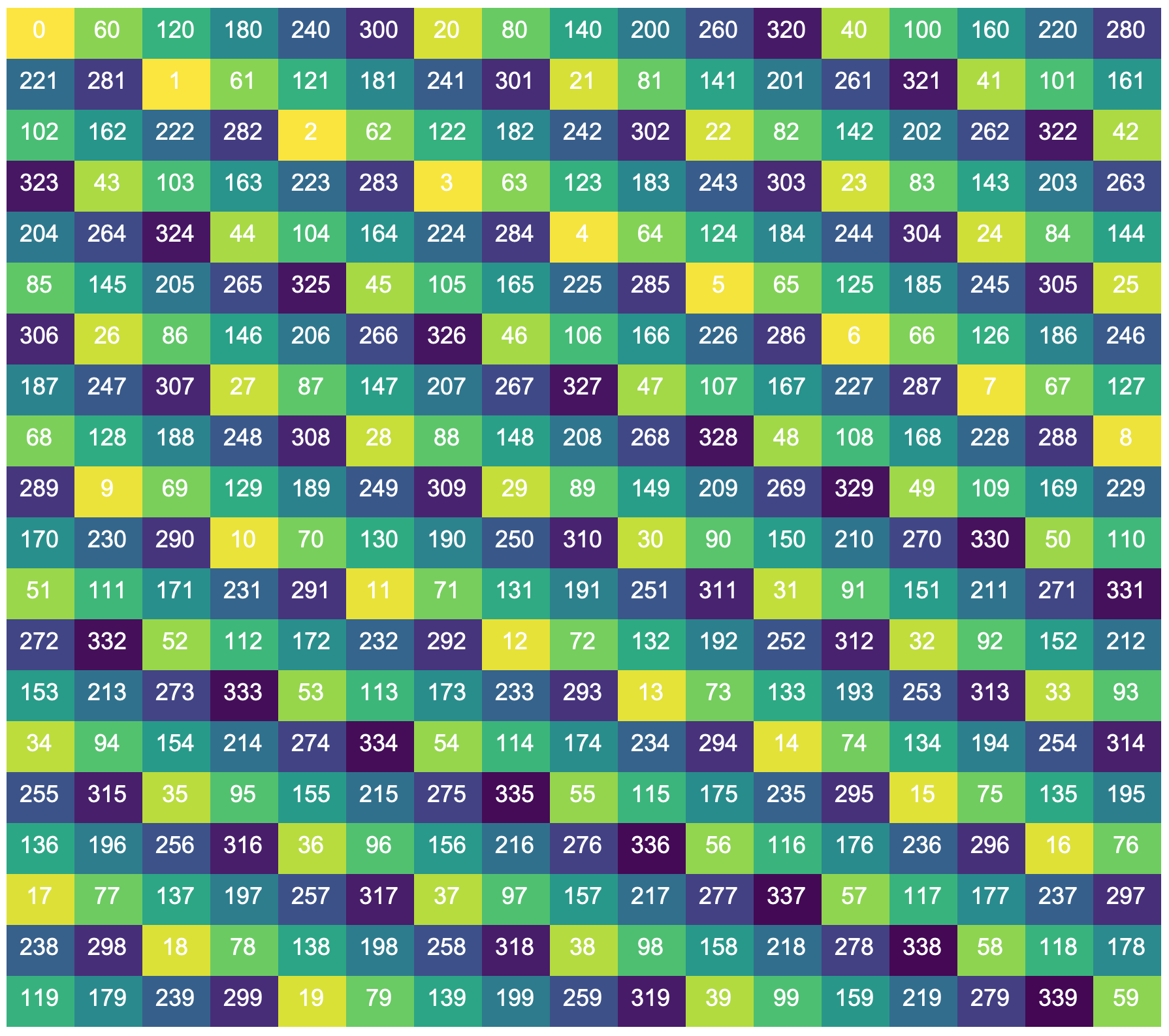} 
  \caption{$(1,2)$-decimation of a 20 by 17 array.}
  \label{fig:12_decimation}
\end{figure}

The pseudo-period 19, period 19, and the $(1,2)$-decimation of \textit{Z340} each possess 37 repeating bigrams. Blake passed these observations on to Oranchak by email in early 2019. \\

Blake then enumerated billions of candidate transposition schemes and applied each scheme in an iterative way to produce a set of transposition variants. These variants were sieved based on thresholding the number of repeating bigrams. \\

Using these variants, during 2019 and 2020, Oranchak and Blake conducted experiments on \textit{Z340} with these steps:

\begin{itemize}
  \item Blake produced batches of transposition variations of \textit{Z340}, under certain fixed transposition schemes.
  \item Oranchak processed each variation in \textit{AZdecrypt} which would produce a batch of candidate plaintexts.
  \item Blake processed each variation in \textit{ZKDecrypto} using the \textit{Spartan} supercomputer \cite{spartan} at The University of Melbourne. 
  \item Oranchak analyzed the plaintext results to determine if any of them warranted further scrutiny.
  \begin{itemize}
    \item \textit{AZdecrypt} ranks each plaintext result based on a composite score that includes language \textit{n}-gram statistics, and the entropy (or measure of unpredictability) of the plaintext\footnote{The entropy is needed to keep the solver from converging to solutions that only use a few of the most common letters in the English alphabet (``\textit{n}-gram spaghetti''). \cite{eyckekeynote2021}}.  Plaintexts that more closely resemble English language text score higher than plaintexts that resemble gibberish.
  \end{itemize}
\end{itemize}

\textit{Z340}'s cipher alphabet contains 63 distinct symbols, which lowers the cipher's index of coincidence compared to substitution ciphers with smaller alphabets.  This results in a much larger space of possible substitution keys.  Therefore, there was a greater probability of finding readable words and phrases when automatic solvers such as \textit{AZdecrypt} searched for high-scoring substitution keys.  But among such results, none appeared to be on the right track.\\

Blake and Oranchak expanded their search to include the hypothesis that Zodiac may have split his cipher into multiple sections. Starting with one horizontal split, Blake and Oranchak considered all positive $a$, $b$ such that $a + b = 17$ \textit{(Figure \ref{fig:split_horizontal_1})}. \\

\begin{figure}[h]
  \centering
  \includegraphics[width=0.55\columnwidth]{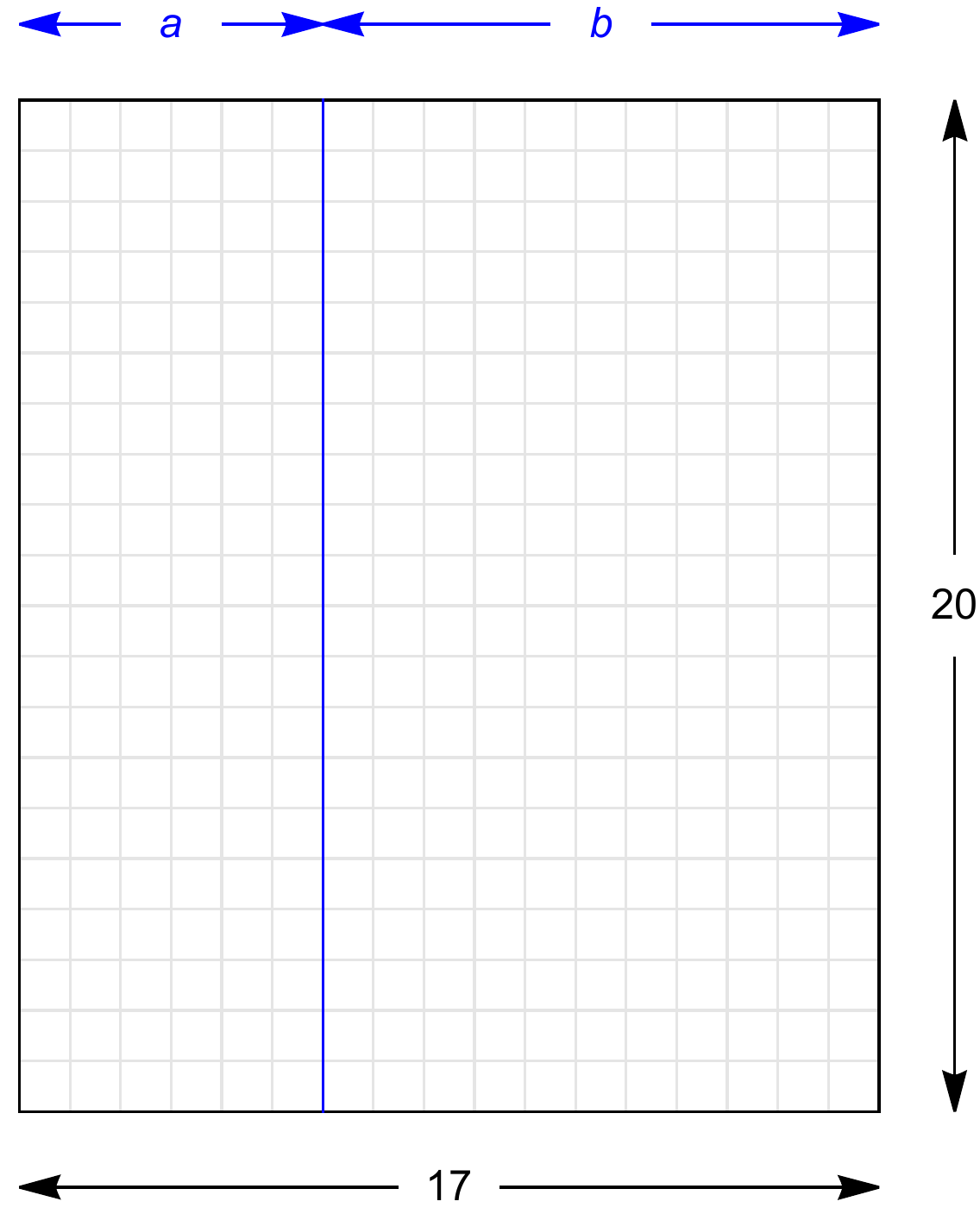} 
  \caption{Splitting the cipher horizontally into two vertical segments.}
  \label{fig:split_horizontal_1}
\end{figure}

Then one vertical split, considering all positive $a$, $b$ such that $a + b = 20$ \textit{(Figure \ref{fig:split_vertical_1})}. \\

\begin{figure}[h]
  \centering
  \includegraphics[width=0.55\columnwidth]{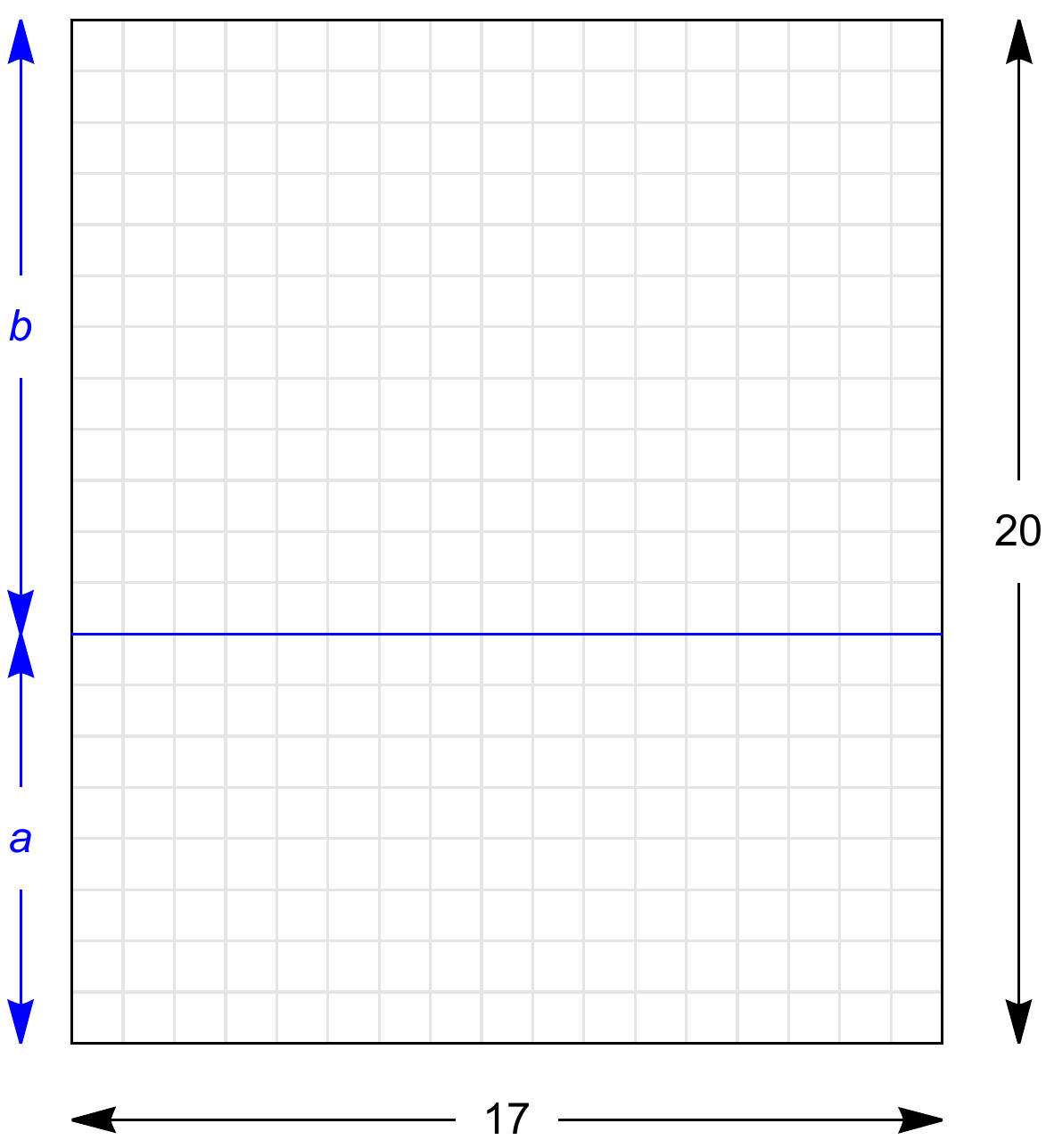} 
  \caption{Splitting the cipher vertically into two horizontal segments.}
  \label{fig:split_vertical_1}
\end{figure}

This was extended to two and three horizontal and three vertical splits, where Blake and Oranchak considered all positive $a$, $b$, $c$, $d$ such that $a + b + c + d = 17$ and $a + b + c + d = 20$ respectively. \\

Blake and Oranchak had wanted to test all candidate transpositions for each section; however, the number of candidates proved to be intractable. So they initially considered using the same candidate transposition on each section. Each was processed in \textit{AZdecrypt} and \textit{ZKDecrypto} with seemingly negative results.

\raggedright
\subsection{Solution breakthrough, first 9 rows}
\justifying

Because of the sheer number of transposition variations, and the fear that a positive result may have been lost in the pile, Blake and Oranchak decided to reprocess them all in \textit{AZdecrypt} with a more careful experimental approach, paying closer attention to the \textit{AZdecrypt} scores and their relationship to cipher length (because many of the transpositions had length less than 340 due to omitted portions that accounted for nulls and skips in the ciphertext).  \textit{AZdecrypt} scoring was changed from 5-gram stats (the default) to 6-gram stats.  It was hoped the increase in \textit{n}-gram stats would improve the accuracy of the solver.  At this stage, the total number of transposition variations to be reprocessed was 655,088.  Also, during the course of previous experiments, Van Eycke released newer versions of \textit{AZdecrypt} with more features, bug fixes, and performance improvements.  A new feature that turned out to be instrumental was the software’s ability to add whitespace characters automatically to candidate plaintexts. Both \textit{Z408} and \textit{Z340} lack word divisions, so decrypted plaintexts also lack them.  But \textit{AZdecrypt}'s new feature could automatically insert them where they likely belonged, based on language \textit{n}-gram statistics.  This made partial solutions much more comprehensible.\\

On December 2, 2020, Blake and Oranchak began running the batch of 655,088 transposition variations of \textit{Z340} in \textit{AZdecrypt}.  By the next day, after only processing a relatively small portion of the variations, \textit{AZdecrypt} reported the following result for one of them, which included some noteworthy words and phrases (marked in all-caps) that attracted their interest:

\begin{adjustwidth}{0.05\textwidth}{0.05\textwidth}
  \begin{tcolorbox}[colback=white,colframe=black!50!white,boxrule=1pt,arc=4pt]
    \texttt{e \textbf{[HOPE YOU ARE]} he sing ist torra enn \textbf{[TRYING TO CATCH ME]} th aftaint mt on the ts \textbf{[SHOT WHICH BRINGS UP]} als in tabs it me name of ar heed \textbf{[OR THE GAS CHAMBER]} beca ate it wild vent me roler a dice ai i the vs shen because too wha seen tight deserts wors ros me there everyoneed \textbf{[HE HAS NOTHING THEN THEY]} he ach paradict is they alreare and norder ther ameo earre and becauite is yot tv hat mr newe itle never ind baeyn neia at a hoe cdr pet}
  \end{tcolorbox}
\end{adjustwidth}

Those phrases constituted only 23\% of the ciphertext.  The remaining 73\% still appeared to be mostly incomprehensible gibberish.  It was not unusual for candidate plaintexts to feature assortments of comprehensible phrases amidst noise or gibberish, but in this particular candidate, the subject matter of the legible phrases was striking, and evidently consistent with the tone of writing associated with Zodiac's other correspondences (\textit{Figure \ref{fig:similar_phrases}}).  For example, ``HOPE YOU ARE'' is reminiscent of Zodiac's phrases from other letters:  ``I hope you have fun'' \cite{voigtbomb2001}, ``I hope you do not think'' \cite{voigtmyname2001}, and ``I hope you enjoy your selves'' \cite{voigtdragon2004}.  The phrase ``TRYING TO CATCH ME'' resembled a theme that appeared in other Zodiac letters:  ``The police shall never catch me'' \cite{voigtbusbomb2004}, and ``If the Blue Meannies [sic] are evere [sic] going to catch me'' \cite{voigtlatimes2001}.  These factors compelled us to pay closer attention to this particular solution candidate.\\

\begin{figure}[h]
  \centering
  \includegraphics[width=\columnwidth]{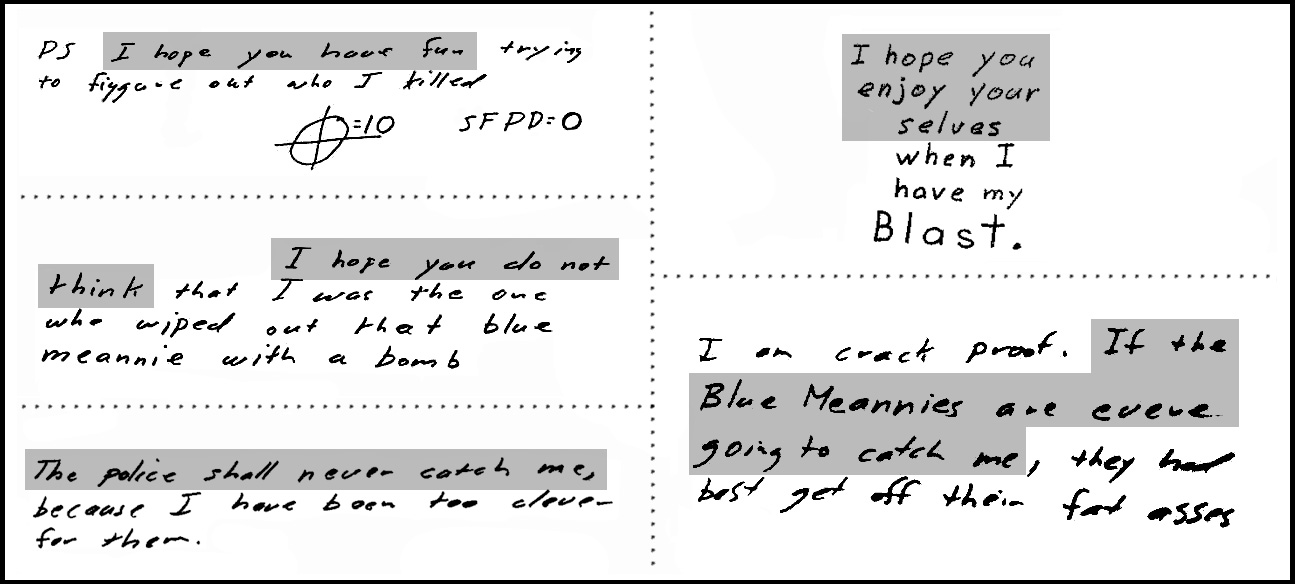} 
  \caption{The phrases ``HOPE YOU ARE'' and ``TRYING TO CATCH ME'' resemble these phrases in Zodiac's other correspondences.}
  \label{fig:similar_phrases}
\end{figure}

The transposition variation that led to this solution was one that split the original 20 rows of \textit{Z340}'s ciphertext into four sections:

\begin{itemize}
  \item \textit{Section 1}: The first 9 rows
  \item \textit{Section 2}: The second 9 rows
  \item \textit{Section 3}: Row 19
  \item \textit{Section 4}: Row 20
\end{itemize}

The same transposition scheme, a $(1,2)$-decimation had been applied to each section \textit{(Figures \ref{fig:decimation_matrix} and \ref{fig:decimation})}.  This scheme is formally defined in \textit{Section~\ref{sec:transpo}}, but can also be described with the following procedure:

\begin{enumerate}
  \item Start at position 1.  This is considered the current position.
  \item Write out the cipher symbol at the current position.
  \item Move down one position, wrapping around to the first row if the bottommost boundary of the grid is reached.
  \item Move right two positions, wrapping around to the first column if the rightmost boundary of the grid is reached.
  \item Continue from step 2 until all positions are visited.
\end{enumerate}

\begin{figure}[h]
  \centering
  \includegraphics[width=0.8\columnwidth]{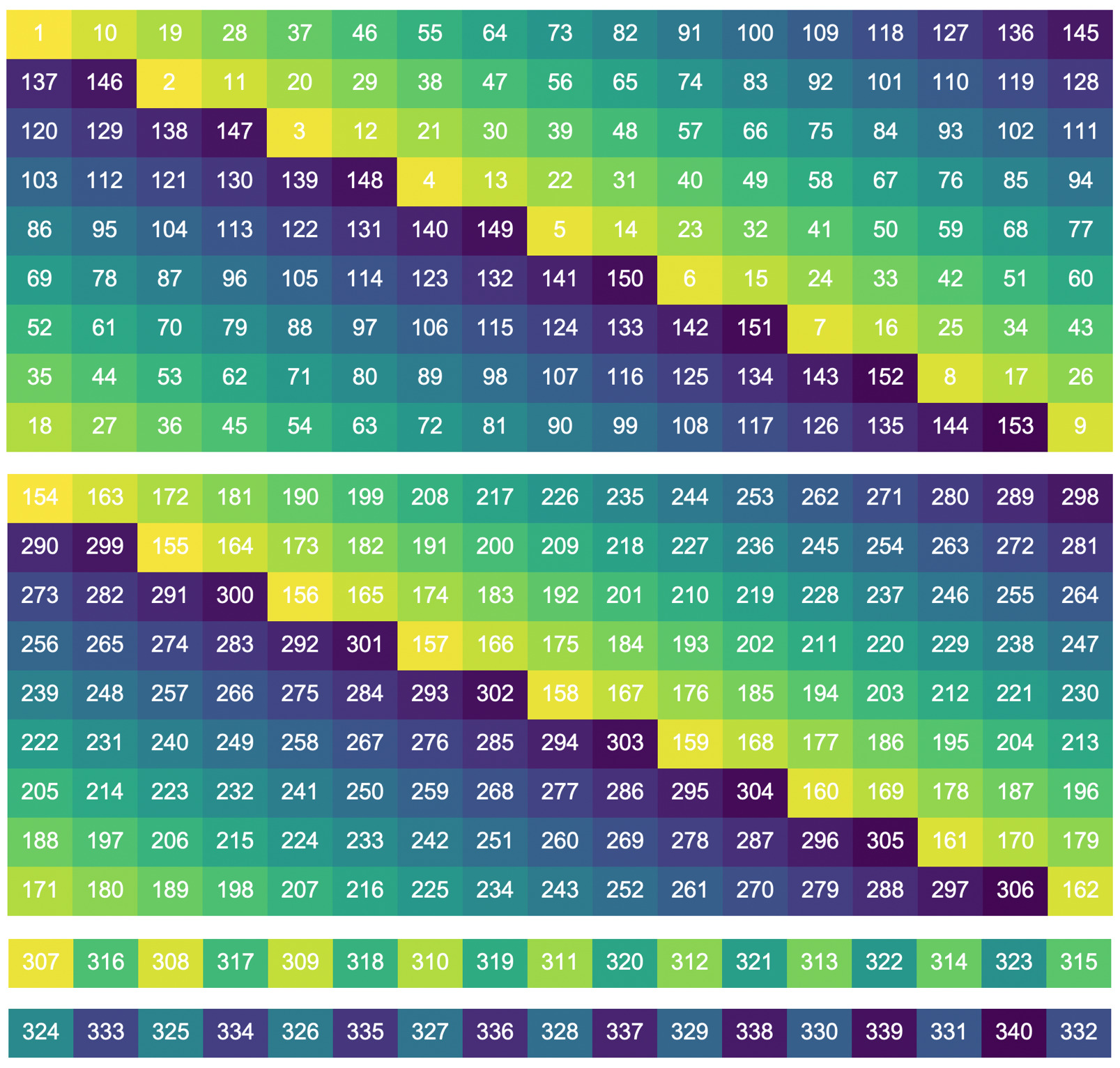} 
  \caption{A $(1,2)$-decimation matrix for all four sections of \textit{Z340}.}
  \label{fig:decimation_matrix}
\end{figure}

\begin{figure}[h]
  \centering
  \includegraphics[width=1.0\columnwidth]{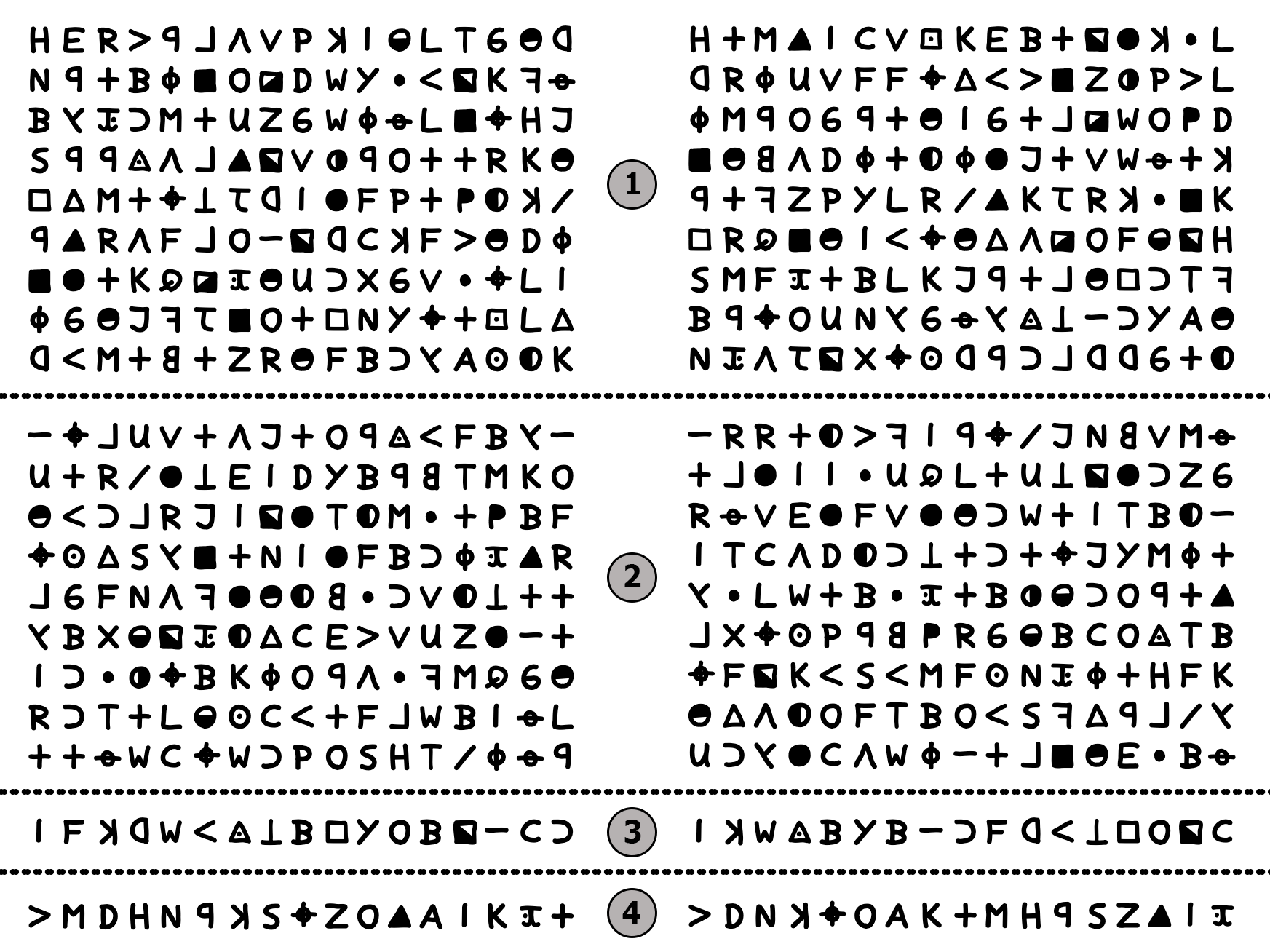} 
  \caption{The $(1,2)$-decimation from \textit{Figure \ref{fig:decimation_matrix}} applied to all four sections.  Left: Original \textit{Z340} ciphertext; Right: Results after applying decimation to each section.}
  \label{fig:decimation}
\end{figure}

Because the solution was far from complete, and the transposition was applied to individual sections, Oranchak decided to run Section 1, consisting of the first 9 rows of \textit{Z340}, in \textit{AZdecrypt}.  This produced a negative result \textit{(Figure \ref{fig:azdecrypt_negative})}, likely due to the high \textit{multiplicity}.  Multiplicity \cite{king1993framework} is the ratio of the key length to the cipher length.  When considering only the first 9 rows, this ratio is thus $63/153$ or about $0.41$.  By contrast, the multiplicity of \textit{Z408} is only $0.13$.  Automatic solvers can find solutions for low multiplicity ciphers more easily than they can for high multiplicity ciphers, due to the smaller solution space of the former\footnote{We later found that \textit{AZdecrypt} can solve the first 9 rows without using cribs if \textit{n}-grams of length 7 or higher are used.}.\\

\begin{figure}[h]
  \centering
    \begin{tcolorbox}[colback=white,colframe=black!50!white,boxrule=1pt,arc=4pt]
      \texttt{to eight and mous the first lled think there say song yours again year ours tots pot so en their indith indiving tenderal but welcoment sound thems easily place thanized use of stuff you}
    \end{tcolorbox}
  \caption{AZdecrypt solution for first 9 rows of transposed \textit{Z340}, no cribs applied.  Negative result.}
  \label{fig:azdecrypt_negative}
\end{figure}

To overcome this limitation, Oranchak used the cribbing feature in \textit{AZdecrypt}.  \textit{Cribbing} is the insertion of known or suspected text into a candidate plaintext, to try to coax more legible plaintext out of the non-cribbed regions.  Oranchak placed the phrases HOPE YOU ARE, TRYING TO CATCH ME, and THE GAS CHAMBER into the locations where they appeared in the plaintext that had resulted from the batch run.  Placing the cribs in their corresponding positions resulted in substitutions for 27 of the 63 unique symbols of \textit{Z340's} cipher alphabet (or about 43\% of its key), yielding 83 out of 153 characters (or 54\%) of the entire plaintext for the first 9 rows \textit{(Figure \ref{fig:azdecrypt_cribs})}.\\

\begin{figure}[h]
  \centering
  \includegraphics[width=1.0\columnwidth]{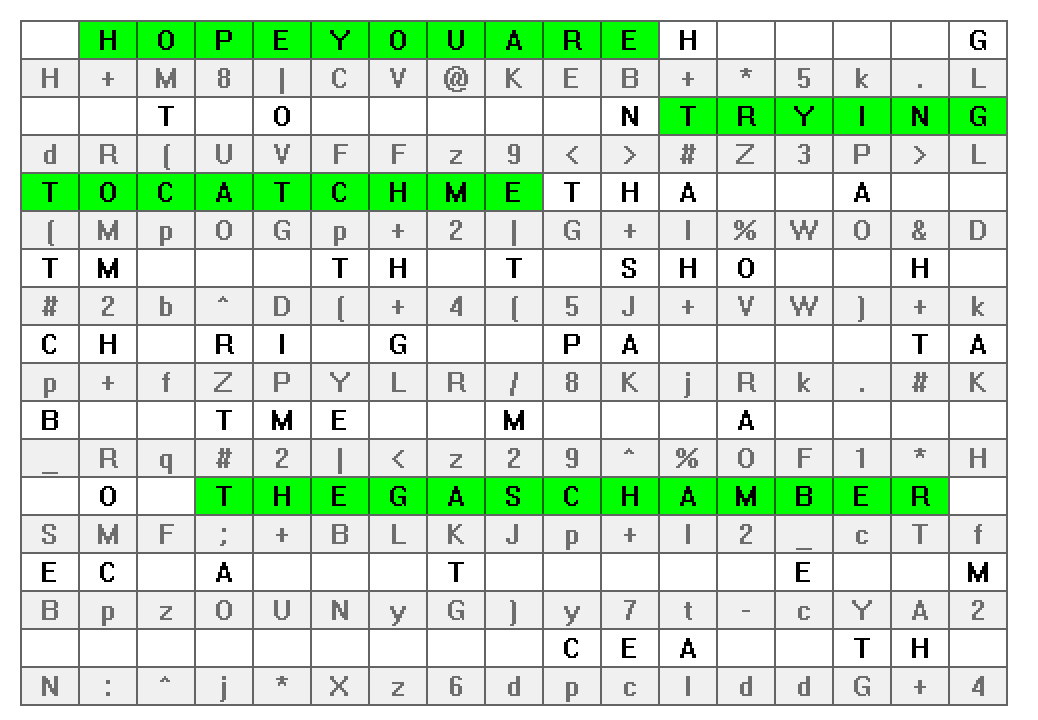} 
  \caption{\textit{AZdecrypt} cribs for Section 1 of \textit{Z340}'s ciphertext.  }
  \label{fig:azdecrypt_cribs}
\end{figure}

\textit{AZdecrypt}'s key search was then initiated to recover the remaining 46\% of plaintext, looking through a much smaller search space due to the large amount of fixed partial plaintext.  It quickly converged on the plaintext shown in \textit{Figure \ref{fig:cribbed}}.

\begin{figure}[h]
  \centering
  \begin{tcolorbox}[colback=white,colframe=black!50!white,boxrule=1pt,arc=4pt]
    \texttt{i \textbf{[HOPE YOU ARE]} having lots of fun in \textbf{[TRYING TO CATCH ME]} that wasnt me on the tv show which bring or pa point about me i um not afraid of \textbf{[THE GAS CHAMBER]} becuase it will send me to pay unlce all the}
  \end{tcolorbox}
  \caption{Initial \textit{AZdecrypt} solution for cribbed Section 1.  Cribs are marked.}
  \label{fig:cribbed}
\end{figure}

The three of us believed there was a very high chance this message was close to the true message intended by Zodiac, because of these factors:

\begin{itemize}
  \item Despite some garbled misspellings, the message was intuitively comprehensible.
  \item Comprehensible portions of previous decryption attempts in \textit{AZdecrypt} were usually very short by comparison (isolated to individual words and small fragments).
  \item The $(1,2)$-decimation transposition scheme was orderly, and not random rearrangements (anagramming) such as those employed by other misguided solution attempts which can generate numerous non-unique, comprehensible and partially comprehensible solutions (e.g. \cite{graysmith2007zodiac3} \cite{fbivault5} \cite{oranchaklcz1}).
  \item The decryption process did not force predetermined words or phrases into the plaintext, apart from the phrases used in the cribbing, which was only prompted by their spontaneous appearance in the initial batch run of systematic transposition variations of \textit{Z340}.  
  \item The decryption made significant contemporary references that lined up with known events in the Zodiac case:
  \begin{itemize}
    \item Zodiac mailed \textit{Z340} on November 8, 1969.
    \item Seventeen days prior, on October 22, 1969, someone claiming to be Zodiac had repeatedly called into Jim Dunbar's live call-in TV program which also featured prominent attorney Melvin Belli \cite{ztvcf1969}.
    \item The decryption said, ``That wasn't me on the TV show.''  
    \item During the program, the same caller said, ``I need help.  I'm sick.  I don't want to go to \textbf{the gas chamber}.'' \cite{duston1969} (emphasis ours)
    \item \textit{Z340's} Section 1 decryption said, ``I am not afraid of the gas chamber.''
  \end{itemize}
\end{itemize}
\raggedright
\subsection{Solution breakthrough, remaining sections}
\justifying

\begin{figure}[h]
  \centering
  \includegraphics[width=0.95\columnwidth]{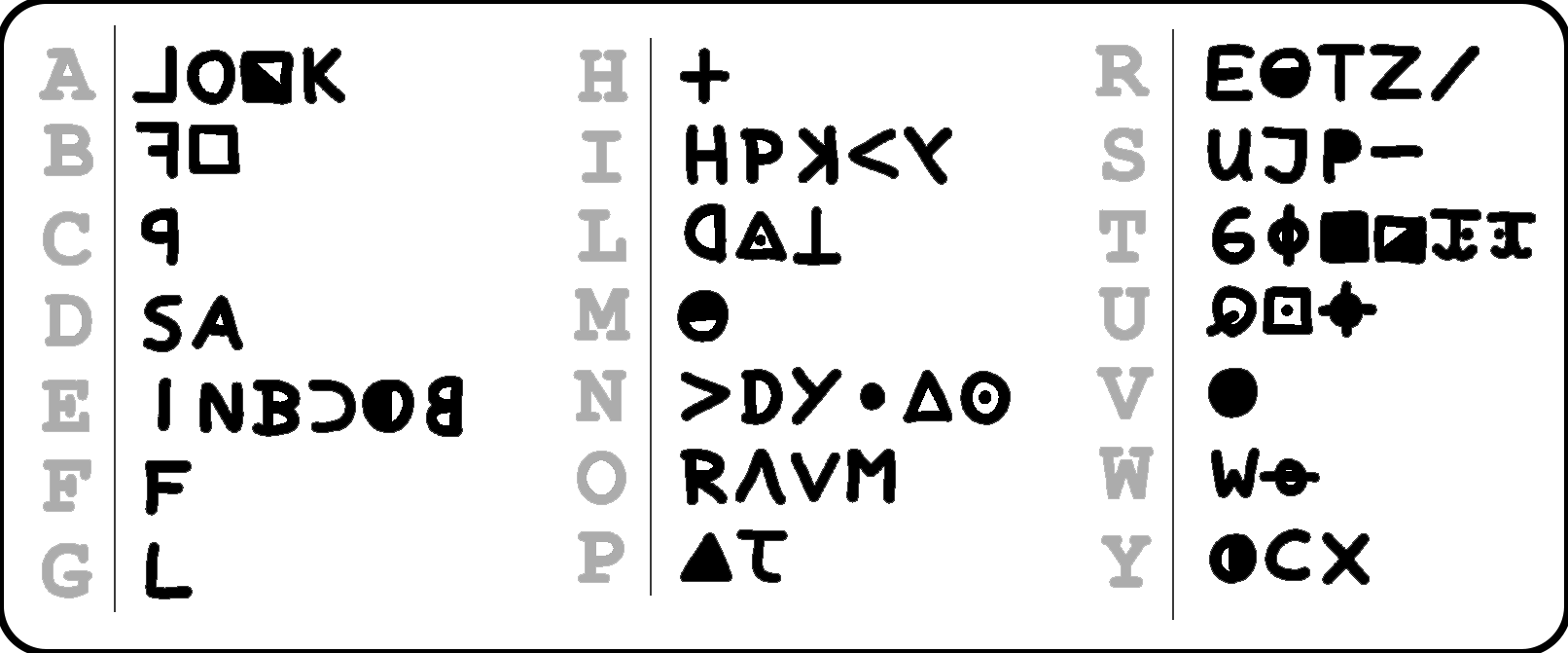} 
  \caption{Homophonic substitution key for \textit{Z340} derived from Section 1 breakthrough.  Note: There are no substitutions for the letters J, K, Q, X, and Z since none appear in the plaintext.}
  \label{fig:Z340_key}
\end{figure}

All symbols of Zodiac's cipher alphabet for \textit{Z340} appear in the first nine rows of the transposed cipher.  Therefore, the entire cipher key was already determined \textit{(Figure \ref{fig:Z340_key})} and could be applied to the remaining sections of the cipher, resulting in the following 340-character plaintext:

\raggedright
\begin{itemize}
  \item \textit{\underline{Section 1}:} \texttt{I HOPE YOU ARE HAVING LOTS OF FUN IN TRYING TO CATCH ME THAT WASNT ME ON THE TV SHOW WHICH BRING OR PA POINT ABOUT ME I UM NOT AFRAID OF THE GAS CHAMBER BECUASE IT WILL SEND ME TO PAY UNLCE ALL THE}
  \item \textit{\underline{Section 2}:} \texttt{SOO HEN BE CURSEE OOW HAVE ENSUGH SLAVER TO WOR V FOV ME WHERE ESERYONE EL HEH US NOTHING WHEN THEY REACH PAY UNICE SO TREY ALREU FAA I FI OF NET TH IF AM NO EA FREA IF BNC ARISE IV YO WT SHAT MR NEW}
  \item \textit{\underline{Section 3}:} \texttt{EIW LENESE FLIL BAAY}
  \item \textit{\underline{Section 4}:} \texttt{NNE I UADAHO I CFR PET}
\end{itemize}
\justifying

The second section had a high English language \textit{n}-gram score compared to random text or gibberish, but still appeared garbled.  For example, Section 2's top-scoring 6-grams based on English language statistics were as follows: 

\begin{itemize}
  \item \texttt{NOTHIN, WHENTH, HENTHE, SNOTHI, MEWHER, YONEEL, THEYRE, OTHING, INGWHE, ERYONE, HEYREA, THINGW, EWHERE, RYONEE, NTHEYR, ENTHEY, WHEREE, HAVEEN, NGWHEN, HINGWH, SLAVER, GWHENT, EYALRE, EYREAC, OWHAVE, RTOWOR, HENBEC, ERTOWO, YREACH, HEREES}
\end{itemize}

The 6-grams match many common words and phrases in English. (e.g., \textit{nothing, when the, they're, thing when, everyone,} etc.)  Furthermore, when applying this cipher key to the original unmodified \textit{Z340} ciphertext, we noticed some additional clear text \textit{(Figure \ref{fig:Z340_death})}:

\begin{itemize}
  \item The word DEATH appeared at the very end of the last line.
  \item LIFE IS appeared in the upper right portion of the 2nd section.
\end{itemize}

\begin{figure}[h]
  \centering
  \begin{tcolorbox}[colback=white,colframe=black!50!white,boxrule=1pt,arc=4pt]
    \includegraphics[width=0.95\columnwidth]{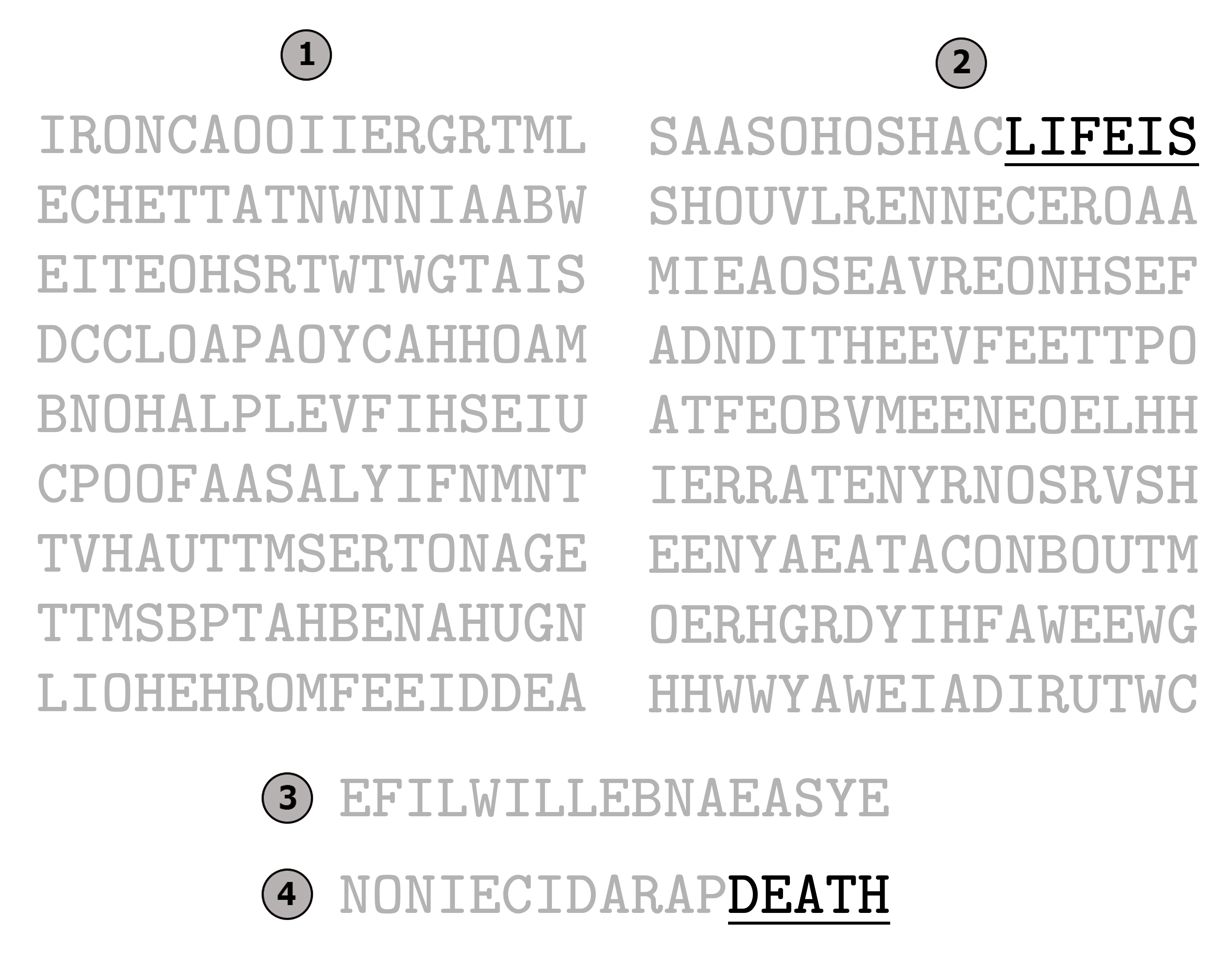} 
  \end{tcolorbox}
  \caption{``DEATH'' and ``LIFE IS'' appear in plaintext prior to transpositions of sections.}
  \label{fig:Z340_death}
\end{figure}

At that point, Oranchak, Blake, and Van Eycke teamed up to determine how to resolve the remaining garbled sections of ciphertext.  Thinking that Zodiac changed his strategy for Section 2 of the cipher, we briefly explored other variations of transpositions that might lead to a better decryption, but reached negative results.  However, we noticed reversed words within the last two lines of the candidate plaintext \textit{(Figure \ref{fig:Z340_last})}.\\

\begin{figure}[h]
\begin{adjustwidth}{0.05\textwidth}{0.05\textwidth}
\centering
\texttt{{\color{black!30}EFIL} \textbf{WILL} {\color{black!30}EB NA} \textbf{EASY}
\vspace*{0.015in}
{\color{black!30}ENO NI ECIDARAP} \textbf{DEATH}}
\vspace*{0.05in}
\rule{0.8\columnwidth}{0.5pt} 
\vspace*{0.035in}
\texttt{\textbf{LIFE WILL BE AN EASY}
\vspace*{-0.015in}
\textbf{ONE IN PARADICE DEATH}}
\end{adjustwidth}  
\caption{Last two lines of \textit{Z340} plaintext.  Top: Normal and reversed words were observed.  Bottom: Result of reversals.}
\label{fig:Z340_last}
\end{figure}

Because the 2nd section seemed only partially garbled, we tried to interpret it using the closest matching words and phrases.  One partial interpretation was:

\begin{itemize}
  \item \texttt{BECAUSE ? HAVE ENOUGH SLAVES TO WORK FOR ME WHERE EVERYONE ELSE HAS NOTHING WHEN THEY REACH PARADICE SO THEY ARE AFRAID OF LETTING ? BECAUSE I (VOW?)...}
\end{itemize}

We soon noticed that errors were systematically appearing on line 15 (the 6th line of Section 2) \textit{(Figure \ref{fig:Z340_garble})}.  Van Eycke determined the cause: there was a symbol skipped on that line.  Much of the garbled text was corrected by circularly right-shifting the segment \texttt{RATENYRNOSRVSH} within the line \texttt{IERRATENYRNOSRVSH}, to read \texttt{IERHRATENYRNOSRVS}\footnote{The \textit{H} at the end wraps around to the fourth position in the line after right-shifting.}.  He also determined that the portion reading LIFE IS needed to be ignored during the transposition step in the 2nd section.  Once those two items were taken into account, transposition of the 2nd section results in comprehensible plaintext.\\

\begin{figure}[h]
  \centering
  \includegraphics[width=0.8\columnwidth]{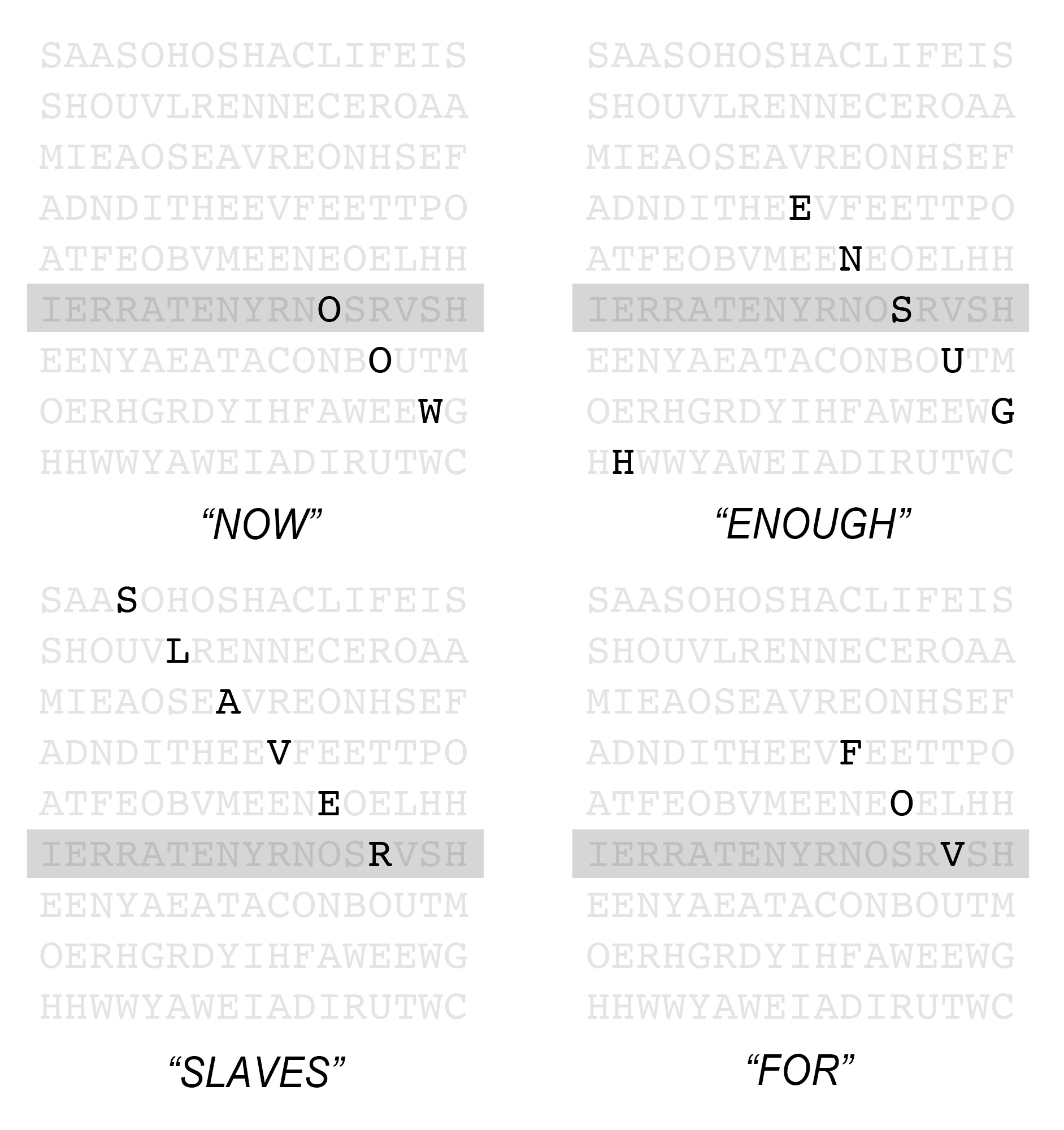} 
  \caption{Examples of four words that were garbled on the same line of Section 2 during transposition.}
  \label{fig:Z340_garble}
\end{figure}

With these corrections, the transposition used in the enciphering of \textit{Z340} is given in \textit{Figure \ref{fig:Z340_final_transposition}}. \\

\begin{figure}[h]
  \centering
  \includegraphics[width=0.85\columnwidth]{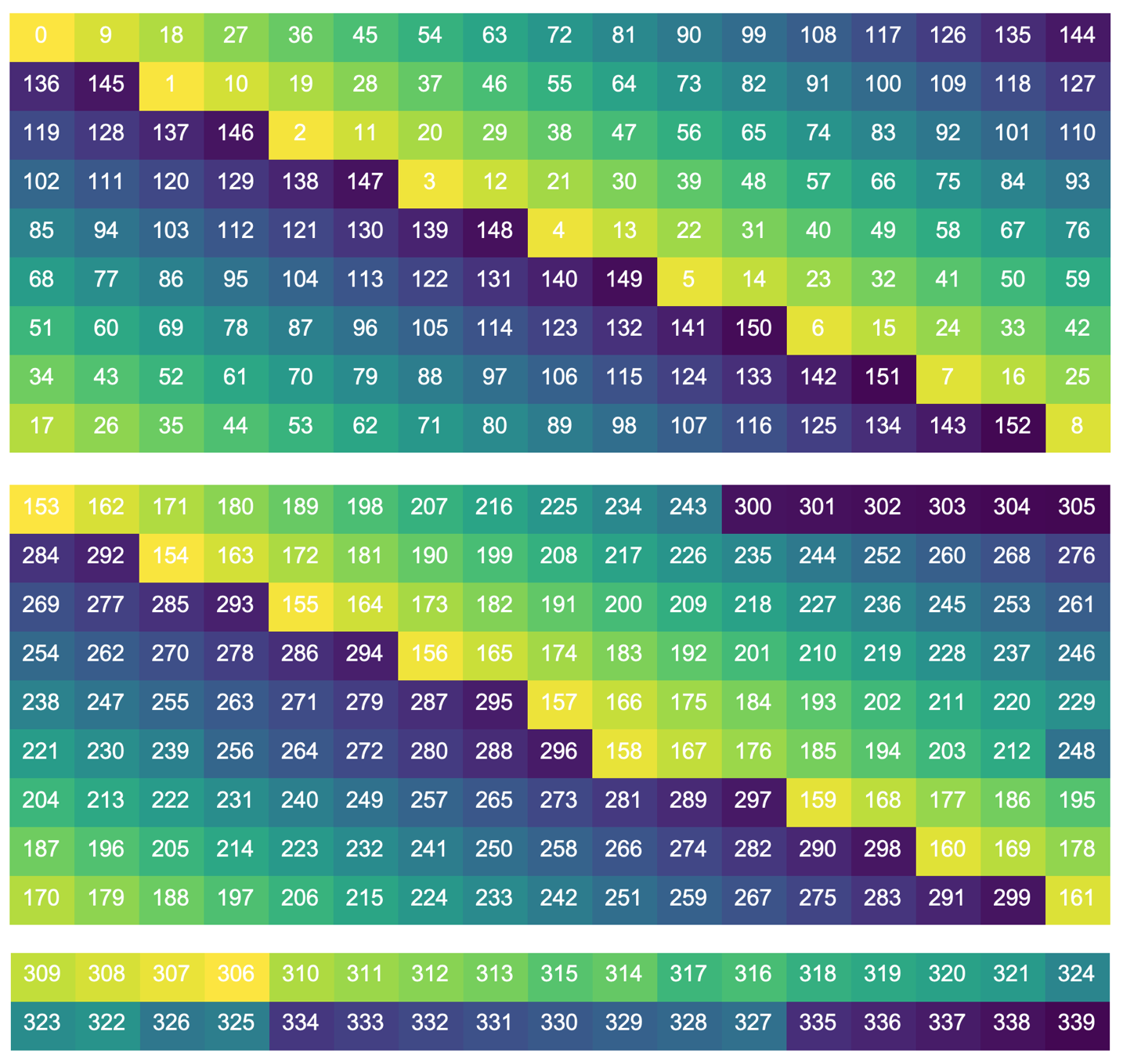} 
  \caption{The transposition used to encipher \textit{Z340}.}
  \label{fig:Z340_final_transposition}
\end{figure}

The full plaintext, including remaining misspellings (in brackets, with assumed corrections)\footnote{We placed the ``LIFE IS'' portion, which was ignored during transposition, near the end, to read ``LIFE IS DEATH''.  The precise placement of ``LIFE IS'' remains subjective, and other possible interpretations are documented at \cite{oranchak340solution}.}, is as follows:

\begin{itemize}
  \item \texttt{I HOPE YOU ARE HAVING LOTS OF [FAN: FUN] IN TRYING TO CATCH ME THAT WASN'T ME ON THE TV SHOW WHICH [BRINGO: BRINGS] UP A POINT ABOUT ME I AM NOT AFRAID OF THE GAS CHAMBER [BECAASE: BECAUSE] IT WILL SEND ME TO [PARADLCE: PARADISE] ALL THE [SOOHER: SOONER] BECAUSE [E: I] NOW HAVE ENOUGH SLAVES TO [WORV: WORK] FOR ME WHERE EVERYONE ELSE HAS NOTHING WHEN THEY REACH [PARADICE: PARADISE] SO THEY ARE AFRAID OF DEATH I AM NOT AFRAID BECAUSE I [VNOW: KNOW] THAT MY NEW LIFE WILL BE AN EASY ONE IN [PARADICE: PARADISE] LIFE IS DEATH}
\end{itemize}

The final substitution key is given in \textit{Figure \ref{fig:Z340_key_final}}.  Final transpositions of cipher and plaintext are given in \textit{Figures \ref{fig:Z340_ba_transpo_1}} and \textit{Figures \ref{fig:Z340_ba_transpo_2}}.\\

\begin{figure}[h]
  \centering
  \includegraphics[width=0.95\columnwidth]{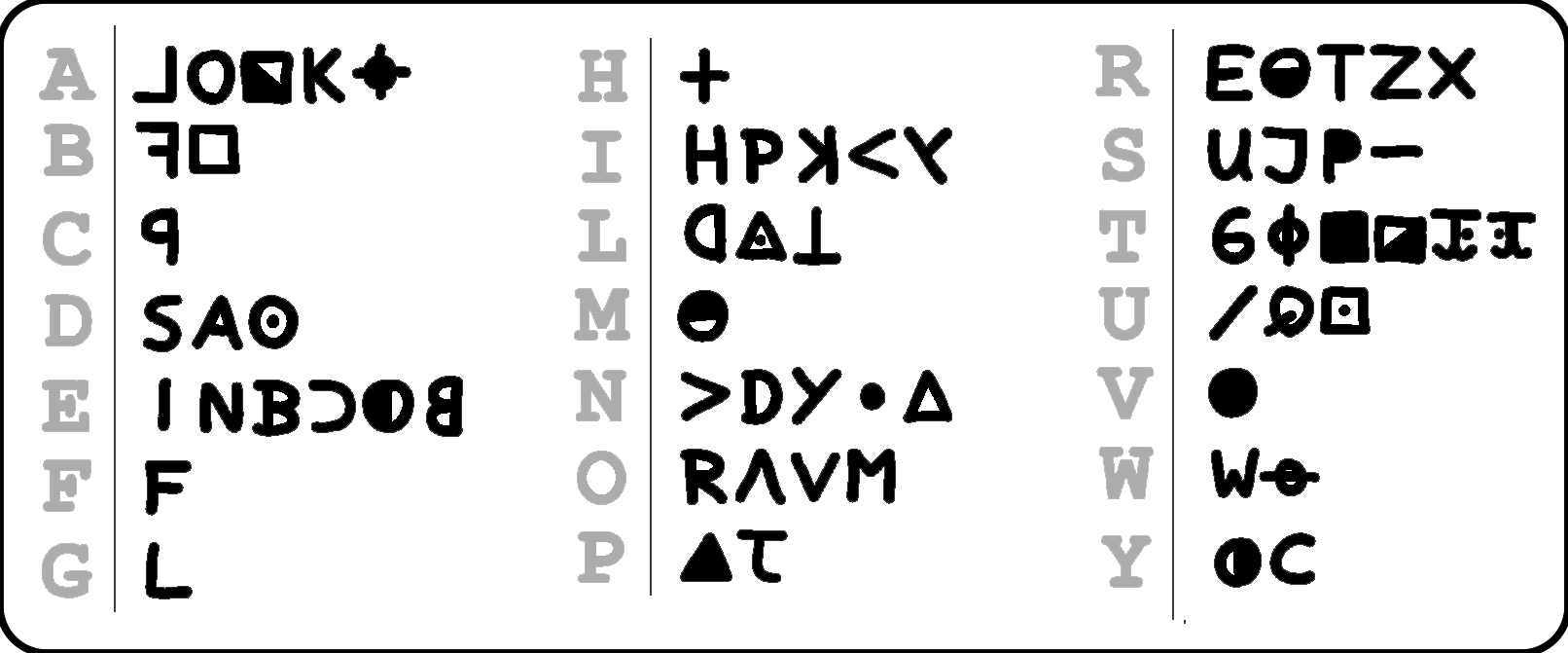} 
  \caption{Refined homophonic substitution key to decipher entire \textit{Z340} ciphertext.  Note: There are no substitutions for the letters J, K, Q, X, and Z since none appear in the plaintext.}
  \label{fig:Z340_key_final}
\end{figure}

On December 5, 2020\footnote{In a bizarre coincidence, this was the 340th day of the year 2020.  We did not plan this!}, we submitted a report summarizing our findings to the FBI Cryptanalysis and Racketeering Records Unit (\textit{CRRU}).  On December 11, after sufficient time was given to conduct victim notifications, we were cleared to release our solution to the public, which was done via a video posted to YouTube \cite{oranchaklcz5}. The FBI released two official statements on December 11, 2020 confirming the solution.  The first is attributed to the San Francisco FBI field office and is a more general announcement.  The second is from the FBI as a whole and gives more detail from the perspective of CRRU, the FBI Laboratory unit that includes cryptanalysts.  The statements are as follows:

\begin{itemize}
  \item FBI, San Francisco Field Office \cite{fagan2020}: ``The FBI is aware that a cipher attributed to the Zodiac Killer was recently solved by private citizens. The Zodiac Killer case remains an ongoing investigation for the FBI San Francisco division and our local law enforcement partners. The Zodiac Killer terrorized multiple communities across Northern California and even though decades have gone by, we continue to seek justice for the victims of these brutal crimes. Due to the ongoing nature of the investigation, and out of respect for the victims and their families, we will not be providing further comment at this time.''
  \item Statement from FBI \cite{butterfield2020}:  ``The FBI has a team of cryptanalysis experts that decipher coded messages, symbols, and records from criminals known as the Cryptanalysis and Racketeering Records Unit.  \textit{CRRU} regularly works with the cryptologic research community to solve ciphers.  On December 5, 2020, the FBI received the solution to a cipher popularly known as \textit{Z340} from a cryptologic researcher and independently verified the decryption.  Cipher \textit{Z340} is one of four ciphers attributed to the Zodiac Killer.  This cipher was first submitted to the FBI Laboratory on November 13, 1969, but not successfully decrypted.  Over the past 51 years \textit{CRRU} has reviewed numerous proposed solutions from the public---none of which had merit.  The cipher was recently solved by a team of three private citizens.''
\end{itemize}

\raggedright
\section{von zur Gathen's estimate of \textit{Z340}'s unicity distance}
\justifying

In the previous section, we outlined several items that support our claim that the true solution has been found, such as contemporary references to events involving Zodiac, the lack of forced words and phrases, and the application of methodical transposition methods rather than random rearrangement.  \\

We supplement the evidence with von zur Gathen's paper \cite{von2023unicity}.  He estimated the \textit{unicity distance} for \textit{Z340}'s cipher system.  \textit{Unicity distance}, defined in Claude Shannon's 1949 paper \cite{shannon1949communication}, is the minimum length of ciphertext needed to guarantee that a legible decipherment is the only one that can be found for the given cipher system.  In other words, if the ciphertext meets this minimum length, then it is practically impossible for another key to exist that produces an equally plausible (but different) decryption.  The calculation of unicity distance is based on the entropy (measurement of uncertainty or unpredictability) of the cipher key space, and the redundancy of the underlying plaintext language.  Von zur Gathen performed a calculation of unicity distance based on the information theoretic contributions of individual components of \textit{Z340}'s encipherment system:

\begin{enumerate}
  \item Homophonic substitution
  \item Sectioned plaintext
  \item Transpositions 
  \item Irregular substitutions (misspellings and ``dummies'' or nulls)
\end{enumerate}

He then considered the entropy (or unpredictability) of the underlying plaintext language by calculating it under three different assumptions:

\begin{enumerate}
  \item The base language is standard English.
  \item The base language is represented solely by \textit{Z340's} plaintext.
  \item The base language is the entirety of Zodiac's corpus, consisting of all of his correspondences.
\end{enumerate}

However, the varying quantitative range found for base language entropies in those three scenarios did not have much effect on the final unicity calculation, which von zur Gathen determined to be at most 152.  Since the discovered \textit{Z340} message has length 340, it is well beyond this limit, and can be concluded to be the only plausible decryption of the ciphertext under \textit{Z340}'s encipherment system, and subject to the assumptions stated in von zur Gathen's paper.

\raggedright
\section{Discussion and open questions}
\justifying

\raggedright
\subsection{Speculations on construction method}
\justifying

The ``backwards L'' patterns (pivots), as discussed in Section \ref{sec:observe} \textit{(Figure \ref{fig:Z340_pivots})}, do not seem to have a known connection to the transposition scheme that Zodiac used.  This remains an open question: Did Zodiac purposefully create those patterns, upon noticing that the transposed plaintext had similar patterns, by carefully assigning symbols to preserve the patterns?  Or were they a random unintended result of the encipherment process?\\

It is not clear what method Zodiac used to apply the transposition scheme to Section 1 and Section 2.  Presumably, he performed the transposition by hand, using only pen and paper.  We are aware of the following two hypotheses for construction methods.\\

\paragraph{Knight's move:}

In chess, the knight piece can move on the board two squares vertically and one square horizontally, or two squares horizontally and one square vertically.  The transposition operation needed to read the decoded plaintext in the untransposed \textit{Z340} cipher resembles this move, where the knight piece always moves right two squares and down one square, wrapping around the boundaries of the section as needed.  It is possible Zodiac constructed the plaintext grid in this fashion, starting his message in the upper left corner and then writing each letter out in turn following this ``two right, one down'' rule.\\

\paragraph{Triangular manipulations:}

An equivalent transposition, discovered by CRRU forensic cryptanalyst Jeanne Anderson, can be achieved using a simple mechanical method involving triangular divisions of the ciphertext grid for sections 1 and 2.  The steps applied to Section 1 are as follows\footnote{A video that shows a demonstration of these steps can be viewed at \cite{oranchaklcz6}, starting at the 10:09 timestamp.}:

\begin{enumerate}
  \item Write out the plaintext for the section in the normal reading direction, in a grid containing 9 rows of 17 characters \textit{(Figure \ref{fig:Z340_step1})}.
  \item Rewrite the plaintext into a grid of the same dimensions, but vertically by columns.
  \item Divide the grid into triangular sections, noted in \textit{Figure \ref{fig:Z340_step2}}.
  \item Move the triangular sections in a mirroring fashion, as shown in \textit{Figure \ref{fig:Z340_step2}}.
  \item The resulting grid of plaintext is the final arrangement of the plaintext for that section prior to substitution with symbols.
\end{enumerate}

A similar process can be followed for Section 2.  However, the complications of ``LIFE IS'' remaining untransposed, and the error made on the 6th line of the section, have to be taken into account.\\

Software tools such as \textit{AZdecrypt} and Kopal's \textit{Z340} transposition reverser \cite{kopal340rev} can be used on the original ciphertext to reverse the transposition process and yield the normal reading order of the encrypted plaintext.\\

\begin{figure}[h]
  \centering
  \includegraphics[width=0.95\columnwidth]{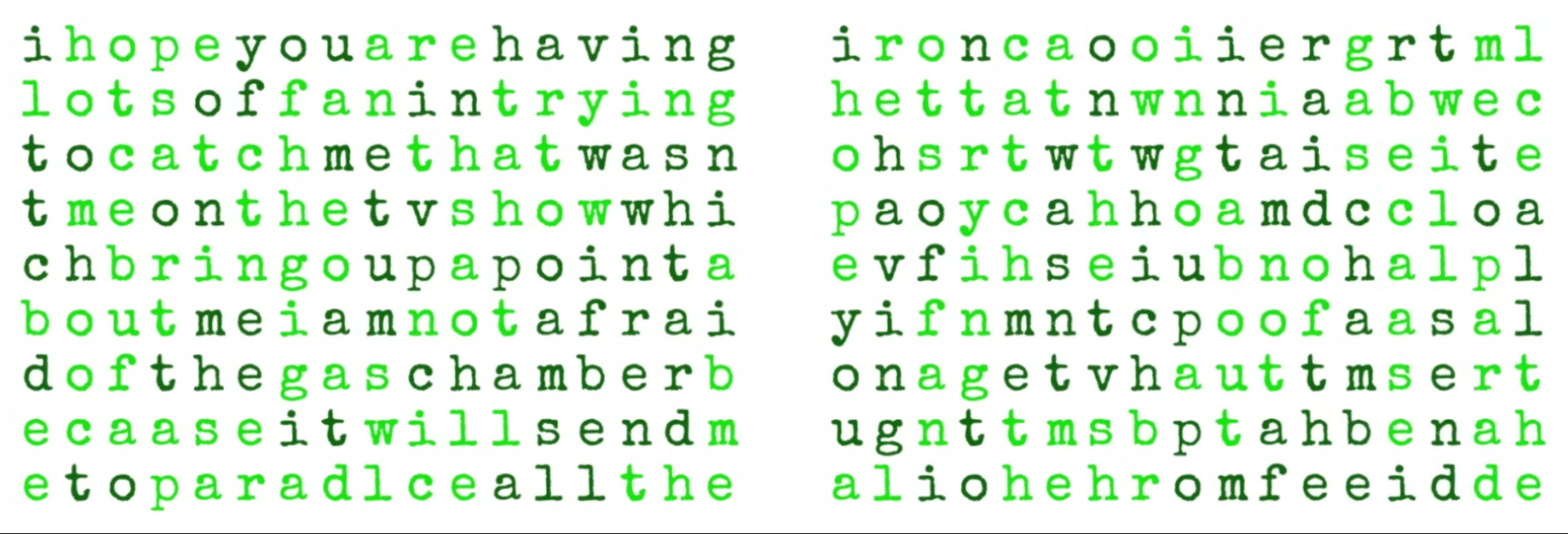} 
  \caption{Rewriting Section 1 plaintext by columns.  On the left is the starting plaintext.  On the right is the plaintext rewritten vertically by columns.  Colors alternate at word breaks.}
  \label{fig:Z340_step1}
\end{figure}

\begin{figure}[h]
  \centering
  \includegraphics[width=0.95\columnwidth]{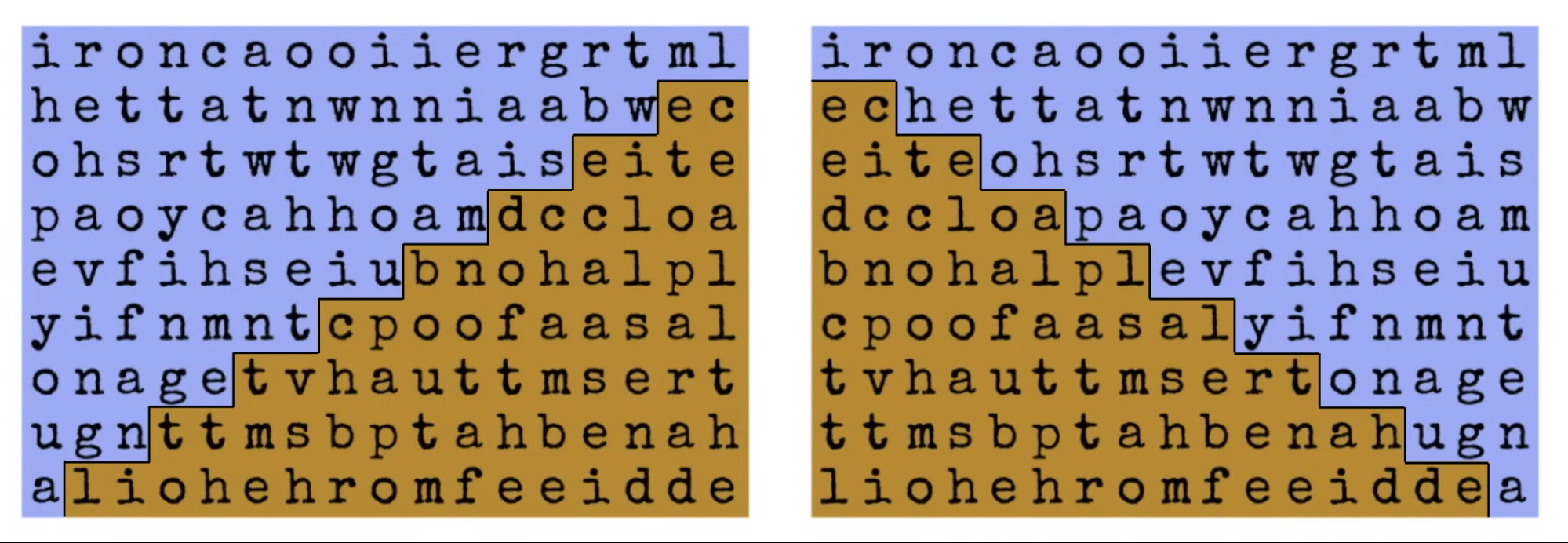} 
  \caption{Rewriting Section 1 plaintext by triangles.  On the left is the plaintext rewritten vertically by columns.  On the right is the plaintext after triangular portions are rewritten in a mirrored fashion.}
  \label{fig:Z340_step2}
\end{figure}

\paragraph{Graph paper:}

An observation made on the Zodiac forums in 2011 \cite{graph2011} noted that graph paper containing half-inch squares (two squares per inch) relates to the combined dimensions of \textit{Z408} and \textit{Z340}.  The standard size for letter paper in the United States is 8.5" x 11".  Thus two such sheets of half-inch ruled graph paper together contain 748 squares, which exactly equals the sum of the lengths of \textit{Z408} and \textit{Z340}.  Whether or not Zodiac intended this remains a subject of speculation.

\paragraph{Transposition clues:}

Another open question is:  Did Zodiac leave hints about how to decrypt \textit{Z340}?  \\

Starting with the card \textit{Z340} was mailed in, Zodiac sent three communications in a row, and a Halloween card, with diagonal writing on the accompanying envelopes \textit{(Figure \ref{fig:diagonal_envelopes})}, something he did not do in his other communications.  We can speculate that these may have hinted at the diagonal reading direction of \textit{Z340's} transposition scheme.  Further, the Halloween card bears a distinctive and unexplained symbol shown in \textit{Figure \ref{fig:halloween}}.  It is made up of diagonal line segments, which could be interpreted as diagonal reading directions, and the segments could be interpreted as splitting a region into three sections, as was required to decrypt \textit{Z340}.  However, there is not yet a way to confirm these speculations.

\begin{figure}[h]
  \centering
  \includegraphics[width=0.95\columnwidth]{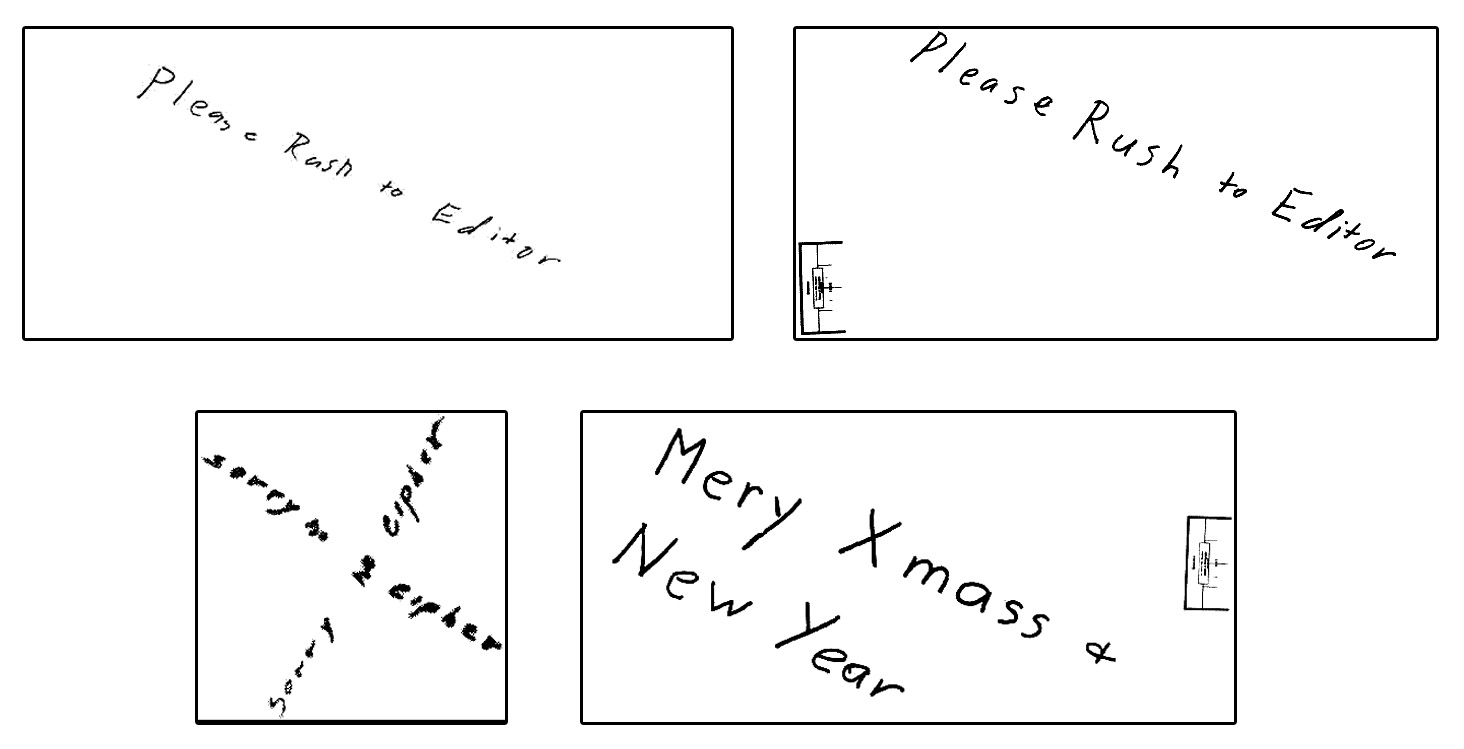} 
  \caption{Four envelopes mailed by Zodiac on which he had written text in diagonal directions \cite{voigtletters}.  Clockwise from top left:  \textit{Z340} envelope (November 8, 1969), ``Bus Bomb'' letter (November 9, 1969), Melvin Belli letter (December 20, 1969), and Halloween card (October 27, 1970).}
  \label{fig:diagonal_envelopes}
\end{figure}

\raggedright
\subsection{Security through obscurity}
\justifying

It is clear that the complications introduced by Zodiac led to much delay in its solution.  If the cipher were simply one of the well-understood classical systems, such as homophonic substitution, transposition, or polyalphabetic substitution (e.g., Vigenère), then traditional cryptanalysis and application of powerful cipher solving tools would have broken it.  But several factors that needed to be discovered confounded cryptanalysis for many years:

\begin{itemize}
  \item The presence of sectional divisions of the cipher grid.
  \item The quantity and dimensions of the sections.
  \item The types of transpositions applied to each section.
  \item The disruptions to the transposition procedures.
  \item The presence and quantity of misspellings and encipherment errors.
  \item The substitution key required to uncover the plaintext.
\end{itemize}

Zodiac succeeded in applying the principle of \textit{security through obscurity}, whereby ignorance of the presence and components of a method of secrecy is enough to protect the secret.  We are not aware of any cryptanalytic tools that could have aided in such a combination of difficulties except for the manual and iterative process we went through.  And if any of the confounding factors had been more extreme (e.g., additional sections, different combinations of transpositions, additional encipherment errors, more unique symbols, etc.), the cipher might still remain unsolved.  

\raggedright
\subsection{Zodiac's cryptographic ability}
\justifying

Can we make any conclusions about Zodiac's skill pertaining to cryptography, and where he may have learned those skills?  The plaintext itself does not seem to reveal any additional information about the perpetrator's identity or his crimes.  We can only speculate on some possibilities.\\

The first is that police and amateur investigators alike have speculated that Zodiac may be linked to the military \cite{reiterman1978}. This speculation is based on observations such as his crew cut hairstyle, close proximity of his crimes to a Navy base on Mare Island in Vallejo, and the perceived expertise in ciphers, explosives, bomb making, orienteering and firearms conveyed by his correspondences.  Also, military-style boot prints were found at one of the crime scenes \cite{napa1969}.  If he served in the military, perhaps he received cryptologic training there.\\

The second possibility is that he was self-taught about cryptography from contemporary sources.  Many books about making and breaking codes were available in his time \cite{squid2015}, including David Kahn's comprehensive history of cryptography, \textit{The Codebreakers} \cite{kahn1996codebreakers5}, first published in 1967, two years before Zodiac created his cryptograms.  Detective fiction stories in periodicals and newspapers in the early 20th century sometimes involved codes and ciphers\footnote{Examples include \cite{keller1920} \cite{unclesam1919} \cite{walk1916} and \cite{white1934}.}.  These periodicals also published many articles about cryptography, such as the 1920s series ``Solving Cipher Secrets'' by M. E. Ohaver \cite{toebes2}.  Ohaver's articles and the contemporary popularity of cryptographic puzzles led to the founding of the American Cryptogram Association in 1930 \cite{acahist2016}.  Comic books in Zodiac's time were also full of codes, ciphers, and articles about cryptography \cite{oranchaklcz12}.  Zodiac's correspondences included several references to pop culture, such as \textit{The Most Dangerous Game} \cite{horton2020}, Gilbert and Sullivan's \textit{The Mikado} \cite{voigtlist}, \textit{The Exorcist} \cite{voigtel2001}, and \textit{Badlands} \cite{voigtbad}.  Thus we might speculate he could have obtained cryptography knowledge from other pop culture sources such as comic books and detective magazines.

\raggedright
\section{Future directions}
\justifying

\raggedright
\subsection{Zodiac's identity}
\justifying

Much of the effort behind attempts to crack \textit{Z340} and Zodiac's other cryptograms came from curiosity about the hidden messages and the possibility that Zodiac may have indeed given up his name or other identifying information, or perhaps additional information about his crimes.  In a letter that he mailed on July 31, 1969 with one segment of \textit{Z408} \cite{ccim1969}, Zodiac claimed: ``In this cipher is my identity.''  Four days later he mailed a letter to the \textit{San Francisco Chronicle} teasing authorities with: ``By the way, are the police having a good time with the code?  If not, tell them to cheer up; when they do crack it they will have me \cite{cknl1969}.''  But within days, \textit{Z408} was solved and revealed Zodiac's admission: ``I will not give you my name because you will try to slow down or stop my collecting of slaves for my afterlife \cite{stbcovm1969}.''\\

In later messages, Zodiac continued his game of teasing that he might reveal his name.  Referring to \textit{Z340} which was still unsolved at the time, he began his April 20, 1970 letter \cite{zsnlct1969} with a similar question:  ``By the way have you cracked the last cipher I sent you?  My name is \rule[0pt]{30pt}{0.25pt}''.  The blank line was immediately followed by his third cryptogram, \textit{Z13}.  Short enough to suggest it might conceal a name, \textit{Z13} is an intriguing target for sleuths and codebreakers.  Many attempts have been made to extract names and other identifying information from the short sequence of symbols \cite{oranchakz13sol2023}.  Given his previous unfulfilled claim to reveal his identity in the puzzle, there is no reason to believe he would simply give up his name in it, unless he understood the difficulty of verifying any solution that might fit.  Nevertheless, people interested in the Zodiac case continue to try to extract names and other information from Zodiac's ciphers and correspondences, on the chance that he really did hide his personal details somewhere.\\

The aforementioned Professor Donald C. B. Marsh \textit{(see Section~\ref{sec:historical})} was again interviewed about the Zodiac ciphers.  In a January 6, 1970 article in the \textit{Vallejo Evening News Chronicle} \cite{belittle1970}, Marsh said he suspected Zodiac would likely have not put his name in \textit{Z340}, since he failed to put any identifying information in \textit{Z408}. Marsh also said he and fellow members of the American Cryptogram association ``have not wasted much time or effort'' on \textit{Z340}, because ``I would imagine… the contents of this second cipher are as valueless as those of the first---consisting merely of boasts or a rehash of his warped philosophical beliefs.''  Now that we know the hidden message in \textit{Z340}, his suspicion was accurate and arguably prophetic.\\

Short of a specific name, we can speculate on some elements of Zodiac's profile based on what we have learned from the two solved ciphers: 

\paragraph{\textit{Z408}:}
\begin{itemize}
  \item He knew enough about cryptography to apply a proper homophonic substitution.
  \item He was sloppy and made some encipherment mistakes and misspellings (or perhaps intentionally made them).
  \item His construction of the cipher grid was overall orderly, and his application of homophones during substitution was organized.
  \item The plaintext message revealed not his identity but a list of his purported motivations for his killings, including his running theme of producing ``slaves for [his] afterlife.''
\end{itemize}

\paragraph{\textit{Z340}:}
\begin{itemize}
  \item He likely made \textit{Z340} more complicated to solve because of how quickly \textit{Z408} was solved.
  \item He added transposition features, which were not present in \textit{Z408}, and reflect a somewhat deeper understanding of classical cryptography. 
  \item The plaintext message again revealed nothing specific about Zodiac's identity apart from denying he was the same ``Zodiac'' that appeared on the Jim Dunbar program with Melvin Belli \cite{ztvcf1969}.  The message continued Zodiac's theme of killing to produce slaves for his afterlife which he referred to as ``paradice [sic]'', a word he also used in other correspondences.
  \item He was concerned with setting the record straight, or at least his perception of the record.  He denied calling in to the Jim Dunbar TV program, writing he had ``grown rather angry with the police for their telling lies'' about him \cite{voigtbusbomb2004}, and denied responsibility \cite{voigtmyname2001} for a bomb that killed San Francisco police officer Brian McDonnell \cite{branning1970}.
  \item He possibly expected \textit{Z340} to be solved relatively quickly, otherwise the reference to Jim Dunbar's TV program would become increasingly dated the longer the cipher remained unsolved.  Perhaps he was better at creating ciphers than solving them or estimating the difficulty of solving them.
\end{itemize}

Readers may research suspects in the Zodiac case by visiting online communities, such as Tom Voigt's Zodiac Killer site (\url{https://zodiackiller.com}) and Mike Morford's Zodiac forum (\url{https://forum.zodiackillerciphers.com/community}).

\raggedright
\subsection{Remaining Zodiac ciphers}
\justifying

\raggedright
\subsubsection{\textit{Z13} and \textit{Z32}}
\justifying

In \textit{Section~\ref{sec:remaining}} we discussed Zodiac's remaining ciphers: \textit{Z13} and \textit{Z32}.  At the time of publication they remain unsolved.  Under minimal assumptions, such as the hypothesis that they employ simple substitution, they do not meet the unicity distance requirements to guarantee unique solutions.  Many plausible plaintexts can be generated under this hypothesis.  For example, thousands of solutions can be generated for \textit{Z13} and are collected at \cite{oranchakz13sol2023}.  And it is almost trivial to generate plausible plaintexts for \textit{Z32} because 29 of its 32 symbols are unique, giving significant freedom in producing substitution keys that yield legible plaintexts.  Therefore, for both cryptograms, there is no known test that can scientifically falsify or validate candidate solutions, apart from the biases and preferences of whoever produces them, or the remote chance that more directly confirming Zodiac materials can be found such as his notebooks or a confession. Now that we know \textit{Z340} was complicated by unexpected factors that were added to simple homophonic substitution, it is possible that Zodiac may have added similar complications to the encipherment of \textit{Z13} and \textit{Z32}, greatly expanding the space of possibilities.  Thus, extraordinary evidence is required to confirm solutions for such short ciphers\footnote{Some possibilities include: \begin{enumerate}
  \item Solid connections are discovered between the solved ciphers and \textit{Z13} or \textit{Z32}.
  \item The ciphers are discovered to have been taken from some other source for which a verified solution is available.
  \item Zodiac's worksheets are discovered and they show how the cryptograms were made.
\end{enumerate}}.

\raggedright
\subsubsection{\textit{Z18}}
\justifying

In \textit{Section~\ref{sec:z408}} we discuss \textit{Z408} and its last 18 symbols, which we denote \textit{Z18}.  The substitution key for \textit{Z408} produces a readable message for all but the last 18 letters, which read:  \texttt{EBEORIETEMETHHPITI}\footnote{Slight variations in this sequence are possible due to some cipher symbols being associated with more than one plaintext letter.  For more details, see \cite{oranchak408key}.}.  Two possible explanations for this sequence are:

\begin{itemize}
  \item It is filler added by Zodiac, ostensibly to fill out the 3rd section of ciphertext so that it matches the dimensions of the previous two sections (\textit{Z408} was split into thirds and mailed to three different newspapers).  
  \item It contains a still unknown secondary message, requiring a different key and/or procedure than that used to decrypt the first 390 characters.
\end{itemize}

If \textit{Z18} does indeed contain a still hidden message, then discovering it faces the same steep challenge presented by \textit{Z13} and \textit{Z32}.  Fifteen of its 18 symbols are unique, making it easy to generate many plausible and scientifically unfalsifiable plaintexts.  By contrast, the 390-character solved portion of \textit{Z408} uses an alphabet of only 54 unique symbols, which highly constrains the space of possible legible plaintexts.

\raggedright
\subsubsection{Miscellaneous symbology}
\justifying

Several mysterious elements, not necessarily cryptographic in nature, still remain in Zodiac's correspondences, and there is much speculation on whether or not they have any meaning.

\paragraph{The Halloween card symbol:}
\begin{figure}[h]
  \centering
  \includegraphics[width=0.95\columnwidth]{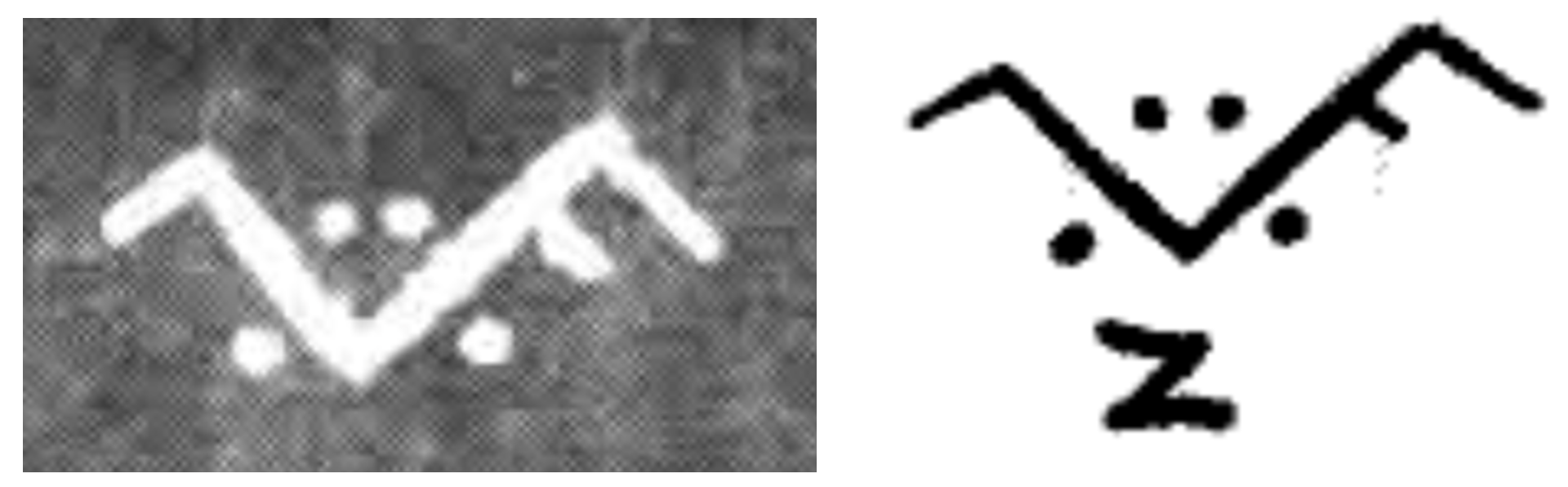} 
  \caption{Symbol found on the Halloween card (left) and its envelope (right) \cite{voigthc1999}.}
  \label{fig:halloween}
\end{figure}

Zodiac mailed a Halloween card to \textit{San Francisco Chronicle} reporter Paul Avery on October 20, 1970 \cite{voigthc1999}.  The card contained handwritten threatening remarks and symbols, including the unique and explained mark shown in \textit{Figure \ref{fig:halloween}} which appeared both inside the card and on the envelope the card was mailed in.  A wide variety of interpretations of the symbol have been suggested, many of which are tracked at \cite{oranchakhc2023}.

\paragraph{The Exorcist letter markings:}
\begin{figure}[h]
  \centering
  \includegraphics[width=0.8\columnwidth]{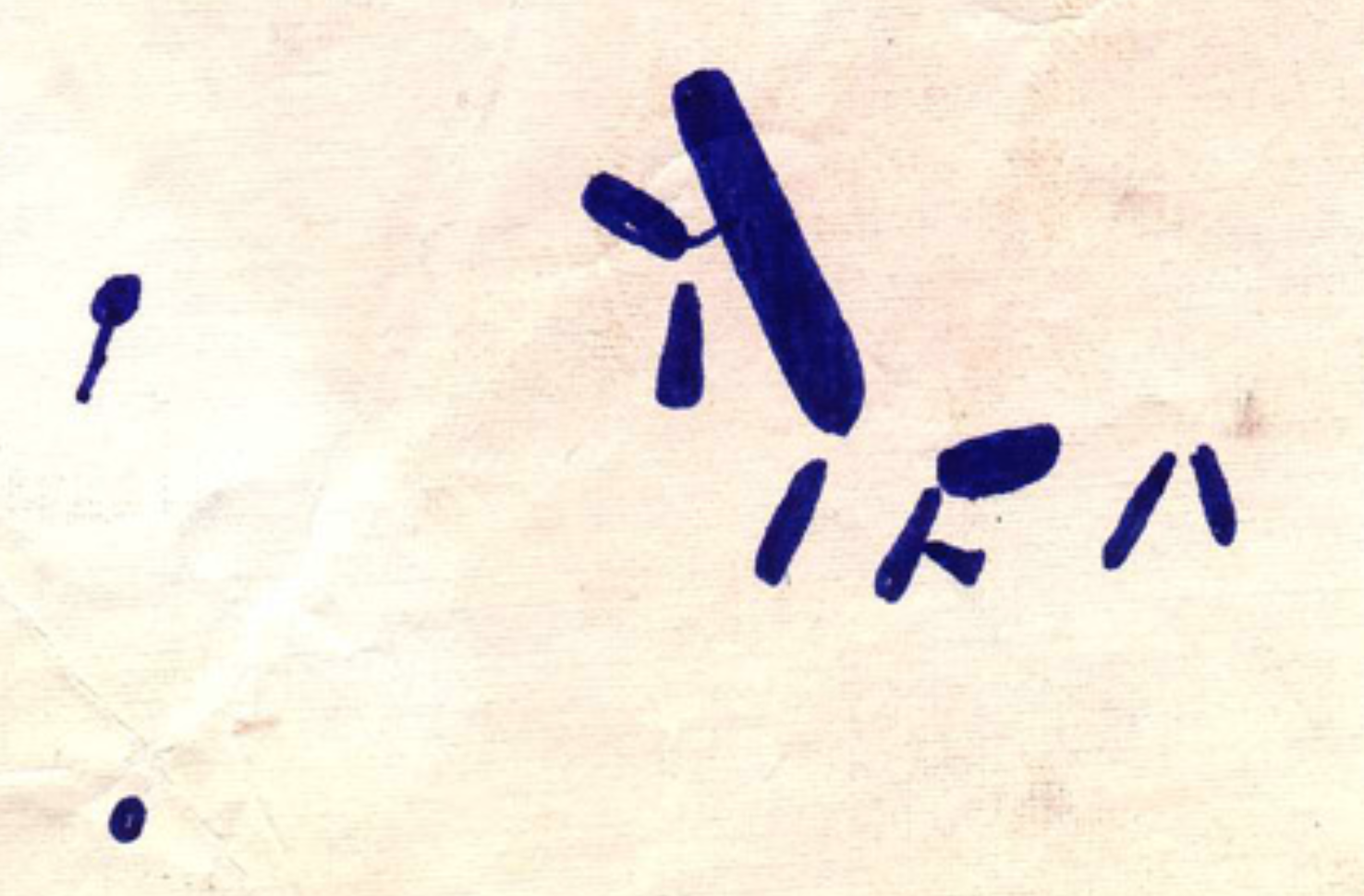} 
  \caption{Exorcist letter symbol \cite{voigtel2001}.}
  \label{fig:exorcist}
\end{figure}

On January 29, 1974, after being silent for almost three years, Zodiac mailed another letter to the \textit{San Francisco Chronicle} \cite{voigtel2001}.  The letter mentioned the film \textit{The Exorcist} and included mysterious markings at the bottom \textit{(Figure \ref{fig:exorcist})} resembling fragments of letters of the alphabet.  Various speculative interpretations of these markings are tracked at \cite{oranchakelm2023}.

\raggedright
\subsection{Improvements to cryptanalytic tools}
\justifying

The resilience of \textit{Z340} to cryptanalysis shows that classical encryption methods can be adjusted in ways that make them overcome their significant security limitations.  Hybrid cryptographic schemes, unpredictable layouts, encipherment mistakes, other errors, ad hoc manipulations, and other factors are often not sufficiently considered when attacking an unknown cipher.  Had \textit{Z340} been constructed with more of these kinds of complications, it might still remain unsolved.\\

Many unsolved ciphers considered to be ``classical'' or pen-and-paper oriented remain unsolved \cite{decryptproject} \cite{unsolved}, which may be due to these kinds of challenges.  Cryptanalysis tools and codebreaking software often look at individual classical cryptographic schemes rather than hybrid or modified schemes such as how \textit{Z340} was constructed.  Modifying these tools to incorporate what we learned from \textit{Z340} will be challenging but possibly essential to breaking at least some of the remaining unsolved ciphers.  Investigating \textit{Z340} was a years-long process of experimental iteration, resulting in the organic growth of skills and software tools required to tackle the problem.  Ultimately, an exhaustive and laborious experimental process, involving both manually and automatically executed steps, was required to unlock the solution.  
One small step that can be made with existing tools is to generalize the exploration of suspected segments of an unknown ciphertext.  A possible avenue of future research would be to add such a feature to a solver such as \textit{AZdecrypt}, such that the dimensions of the segments would be considered part of candidate keys.  Perhaps then the solver can discover partially accurate keys to \textit{Z340} and other unknown ciphertexts without being given any specific information about how they are constructed.\\

\raggedright
\subsection{Cipher type classification}
\justifying

One of the most important steps in the cryptanalysis of an unknown cipher is the identification of the encipherment method.  Over time, cryptographers have developed methods for identifying unique signatures or markers that help reduce the space of possible cryptographic schemes.  Historically this has been performed manually by skilled cryptanalysts but now computers are relied upon to aid the task.  The cipher type identification problem has been studied\footnote{Examples include \cite{abd2019classification} \cite{kopal2020ciphers} \cite{leierzopf2021massive} \cite{leierzopf2021detection} and \cite{nuhn2014cipher}.}, and identification systems have a variety of reported accuracies.  Some systems incorporate \textit{artificial intelligence} concepts such as neural networks, and train classifiers to identify cipher types using large sets of examples of known ciphertexts.  It is conceivable that greater classification accuracies can be achieved with larger training sets and more hardware, such as those used by LLM (\textit{large language model}) applications, like ChatGPT \cite{openai2022}.  Such large training sets could be supplemented with examples of hybrid and ad hoc encipherment systems to improve the generalization power of classifiers.  Improved ability to identify encipherment systems will likely result in breakthroughs in solving more unsolved ciphers.

\raggedright
\section{Summary}
\justifying

After 51 years, the exceedingly large target on the back of the \textit{Z340} has finally been taken down. If not for the previously detailed fragments ``HOPEYOUARE'', ``TRYINGTOCATCHME'', and ``GASCHAMBER'' from the original partial break of the cipher, the \textit{Z340} might still remain unsolved. \\

We have detailed the historical significance of the \textit{Z340}, prior efforts to solve the cipher, our solution, and the validation of the solution. \\

The solution of this cipher was the result of a large, multi-decade group effort, and we ultimately stood on the shoulders of many others' excellent cryptanalytic contributions. \\

We dedicate our efforts to the victims of the Zodiac Killer, their families and descendants.  We hope that one day justice will prevail.

\raggedright
\section{Acknowledgments}
\justifying

We would like to thank Klaus Schmeh, Bill Briere, Elonka Dunin, and Louie Helm, who all generously reviewed and copy edited early drafts of this manuscript.\\

Coincidentally, Elonka Dunin and Klaus Schmeh had just released their book ``Codebreaking: A Practical Guide'' on December 10, the day immediately before the solution to \textit{Z340} was announced.  The book had multiple references to \textit{Z340} as a famous unsolved code so, somewhat tongue-in-cheek, we informed them that now they were going to have to rewrite the book! (which they did, releasing an expanded edition with that and other updates on September 19, 2023 \cite{schmehduninuide}).

\subsection{David Oranchak}

Thanks to Sam, Jarl, and the rest of the Zodiac research community for all their hard work that led to this success.  Mike Morford, Michael Butterfield and Tom Voigt for their Zodiac sites and forums, full of useful information, collections of case materials (a source of many of the images in this paper), and lively discussions.  Brax Cisco, Wesley Hopper, Michael Eaton and David Campbell for developing the influential solver \textit{ZKDecrypto} which inspired Jarl's \textit{AZdecrypt}.  Geofrey LaTurner who contributed many fascinating cipher experiments on the forums in his investigations of the period 19 phenomena.  Thanks to Heiko Kalista for his many contributions on the forums, excellent Zodiac cipher font, and codebreaking tools.  And to Dan Olson, Jeanne Anderson, and Scott Hull at FBI CRRU for their valuable support.\\

And many thanks to my family for their support and putting up with this long quest.

\subsection{Sam Blake}

Many of the experiments we conducted as part of our research into candidate transpositions of \textit{Z340} were conceived because we had access to the \textit{Spartan} supercomputer \cite{spartan} at The University of Melbourne. We thank The University of Melbourne’s Research Computing Services and the Petascale Campus Initiative.\\

This research was conducted while I was a research fellow at The University of Melbourne. \\

I wish to thank my former PhD Advisor, Andrew Tirkel, who was a constant sounding board for my research in this area. \\

Lastly, I wish to thank David and Jarl for their persistence, brilliance, and amazing teamwork. 

\subsection{Jarl Van Eycke}

I wish to make special mentions to:

\begin{itemize}
  \item David Oranchak, for everything, including getting this adventure started for me.
  \item Sam Blake for applying his expert mathematical skills to the problem.
  \item Geoffrey LaTurner (aka ``smokie treats'') for working with me so actively throughout the years on just about everything.
  \item Louie Helm for helping me out with developing AZdecrypt.
  \item All the people at the Zodiac forums, to name a few by their handles: \textit{daikon}, \textit{doranchak}, \textit{glurk}, \textit{Largo}, \textit{Marclean}, \textit{Mr lowe}, \textit{smokie treats}, and \textit{f.reichmann}.
  \item All the people at the FreeBASIC forum, to name a few by their handles: \textit{counting\_pine}, \textit{dodicat}, \textit{fxm}, \textit{jj2007}, \textit{Lothar Schirm}, \textit{MichaelW}, \textit{MrSwiss}, \textit{paul doe}, \textit{PaulSquires}, \textit{SARG}, \textit{srvaldez}, \textit{St\_W}, and \textit{Richard}.
  \item And all the people I forgot to include who should be on this list!
\end{itemize}

\raggedright
\section{About the authors}
\justifying

\subsection{David Oranchak}

Dave is a software developer and cryptologist with over 25 years of experience working on challenging problems.  He has a Master of Science degree in Computer Science from NTU/Walden, and a Bachelor of Science degree in Computer Engineering from Virginia Tech.  He has worked for many areas of industry and levels of government, including the Department of Defense, Defense Intelligence Agency, Department of Justice, and the codebreaking unit of the FBI.  His years of research into the Zodiac ciphers led to the YouTube series \textit{Let's Crack Zodiac} and guest appearances on a variety of TV docuseries \cite{oranchakimdb}.

\subsection{Sam Blake}
Sam has a PhD in mathematics from Monash University in Melbourne, Australia. His PhD concerned families of sequences with good autocorrelation and pairwise cross-correlation. He subsequently used these families of arrays in a Python package for encrypted spread spectrum steganography (watermarking) of audio, imagery, and video. \\

While Sam had known about the unsolved Zodiac cipher for many years, he was inspired to look into \textit{Z340} after watching two lectures given by David Oranchak in 2015 \cite{oranchaktzc2015} and 2018 \cite{oranchakwiz2018}. \\

Sam is currently experimenting with the final unsolved Kryptos cipher \cite{duninkryptos}. 

\subsection{Jarl Van Eycke}

Jarl Van Eycke is a cipher expert from Flanders, Belgium. Originally schooled as a graphic designer, he now works as a warehouse operator for a logistics provider, mostly handling parts for the automotive industry. He started out with cryptography in 2014 trying to solve the unsolved \textit{Z340} cryptogram and it has been a hobby ever since.  Starting in late 2014, Jarl has been the ongoing author of \textit{AZdecrypt}, a fast and powerful cipher solver \cite{schmeh2019}.  Besides \textit{Z340}, he cracked the 350-year old \textit{Code of Langren} \cite{hope2021} \cite{langrenschmeh} in 2021, a 122-year old encrypted newspaper ad \cite{jarlpaperad} in 2022, and holds a world record along with Louie Helm for a bigram substitution challenge \cite{schmeh2019}.

\end{twocolumn}

\onecolumn

\raggedbottom
\listoffigures
\appendix
\section{Appendix}\label{the_appendix}
\subsection*{FBI summary of \textit{Z340} analysis}\label{FBI_340_analysis}
\begin{figure}[H]
  \centering
  \includegraphics[width=0.8\columnwidth]{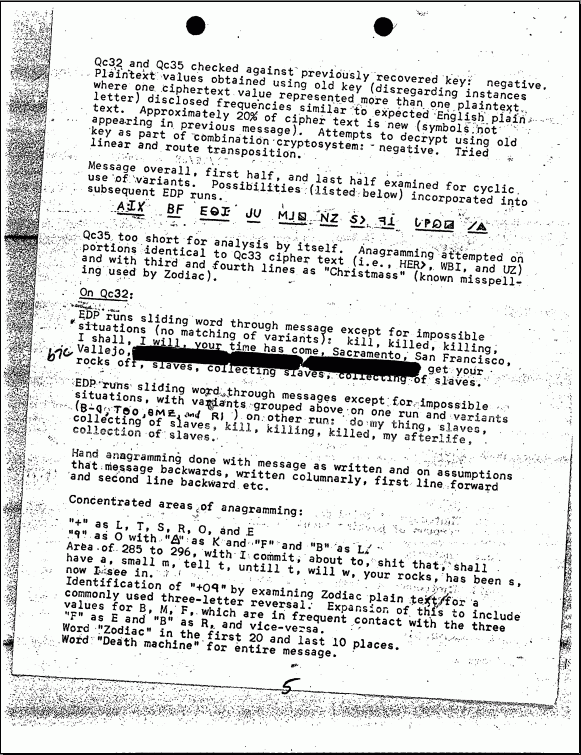} 
  \caption{FBI summary of \textit{Z340} analysis \cite{fbivaultgd}.}
  \label{fig:FBI_340_analysis}
\end{figure}
\subsection*{November 8, 1969 dripping pen card}\label{Z340_card_1}
\begin{figure}[H]
  \centering
  \includegraphics[width=0.8\columnwidth]{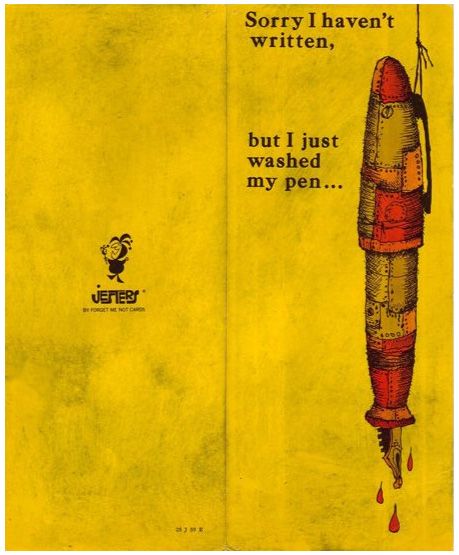} 
  \caption{November 8, 1969 dripping pen card (Back and front) \cite{butterfieldz340}.}
  \label{fig:Z340_card_1}
\end{figure}
\subsection*{November 8, 1969 dripping pen card}\label{Z340_card_2}
\begin{figure}[H]
  \centering
  \includegraphics[width=0.8\columnwidth]{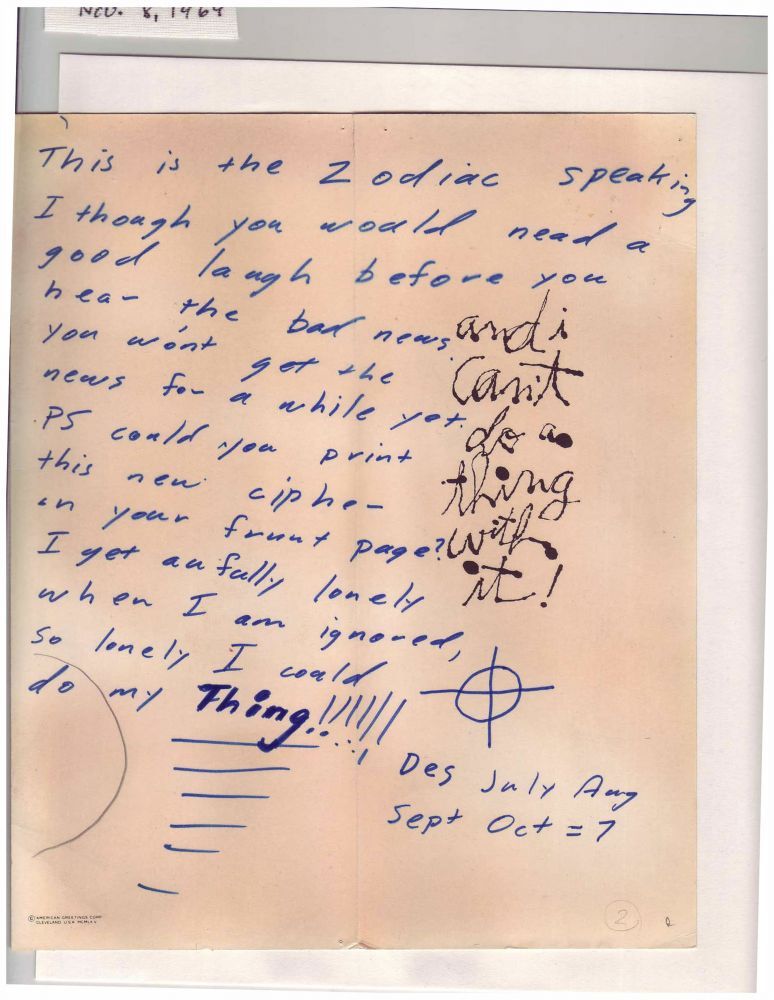} 
  \caption{November 8, 1969 dripping pen card (Inside) \cite{butterfieldz340}.}
  \label{fig:Z340_card_2}
\end{figure}
\subsection*{Final \textit{Z340} ciphertext transpositions.}\label{sec:Z340_ba_transpo_1}
\begin{figure}[H]
  \centering
  \includegraphics[width=0.8\columnwidth]{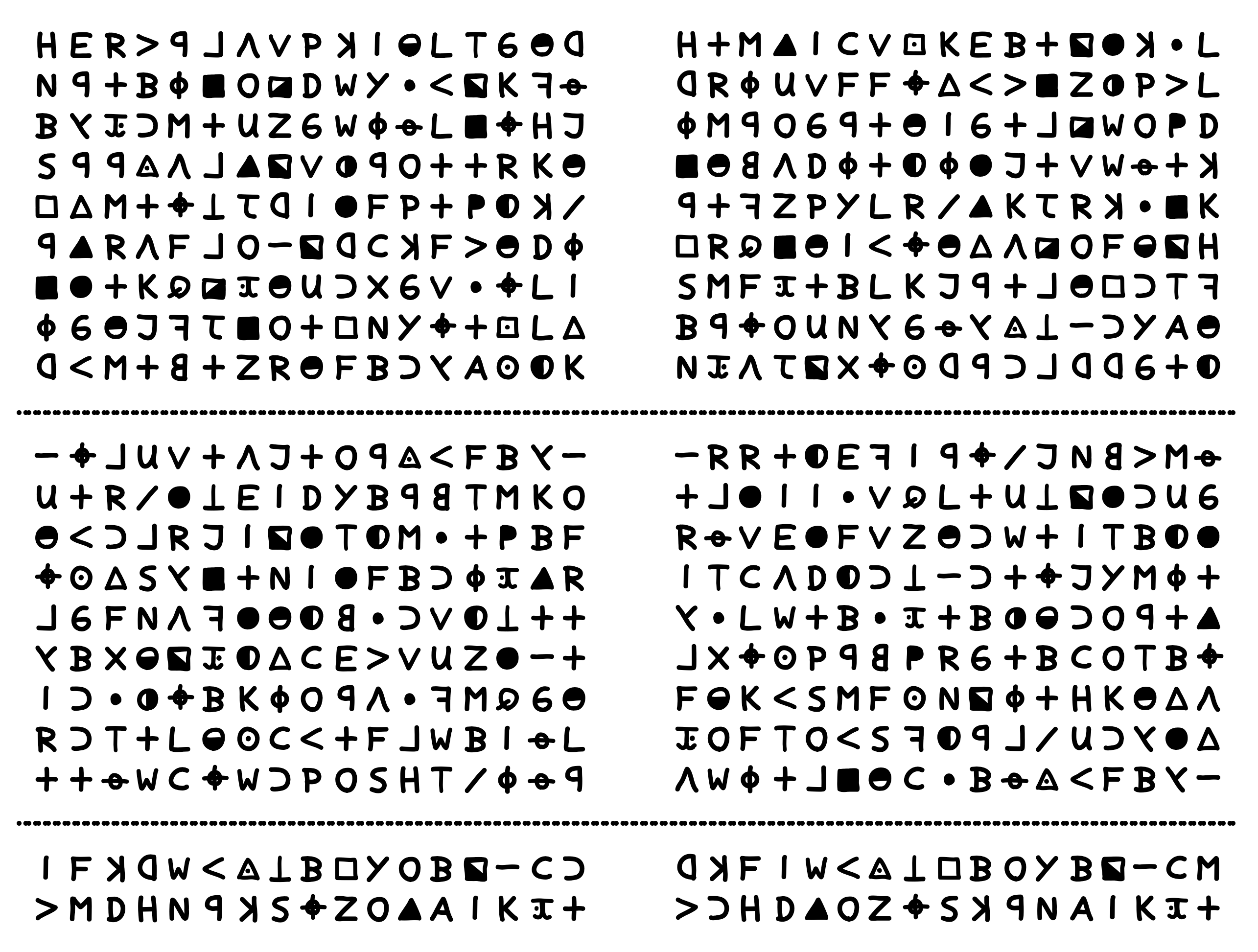} 
  \caption{\textit{Z340} ciphertext sections before \textit{(left)} and after \textit{(right)} transpositions.}
  \label{fig:Z340_ba_transpo_1}
\end{figure}
\subsection*{Final \textit{Z340} plaintext transpositions.}\label{sec:Z340_ba_transpo_2}
\begin{figure}[H]
  \centering
  \includegraphics[width=0.8\columnwidth]{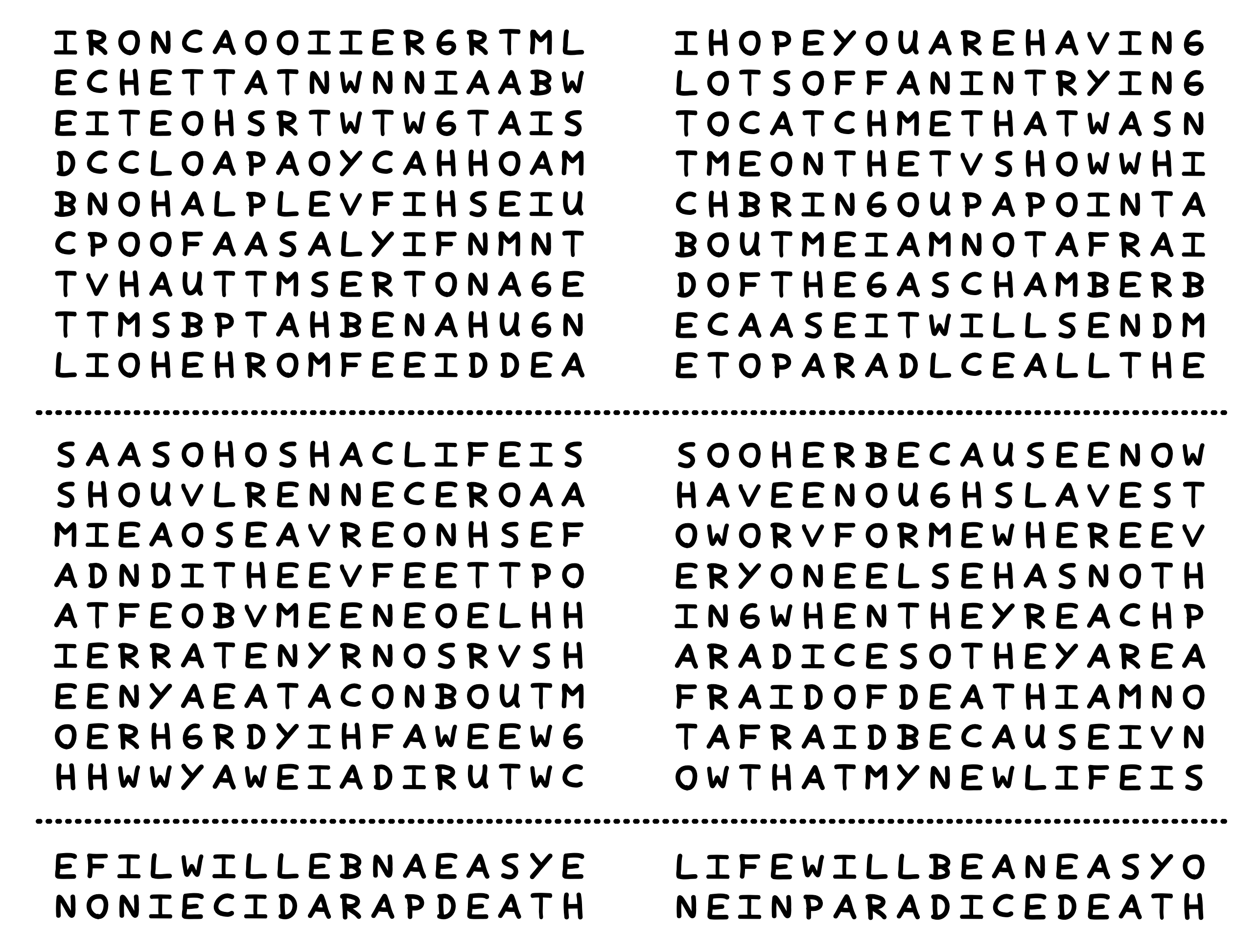} 
  \caption{\textit{Z340} plaintext sections before \textit{(left)} and after \textit{(right)} transpositions.}
  \label{fig:Z340_ba_transpo_2}
\end{figure}
\subsection*{Alleged attempt by NSA to solve \textit{Z340}}\label{sec:nsa_z340}
\begin{figure}[H]
  \centering
  \includegraphics[width=0.8\columnwidth]{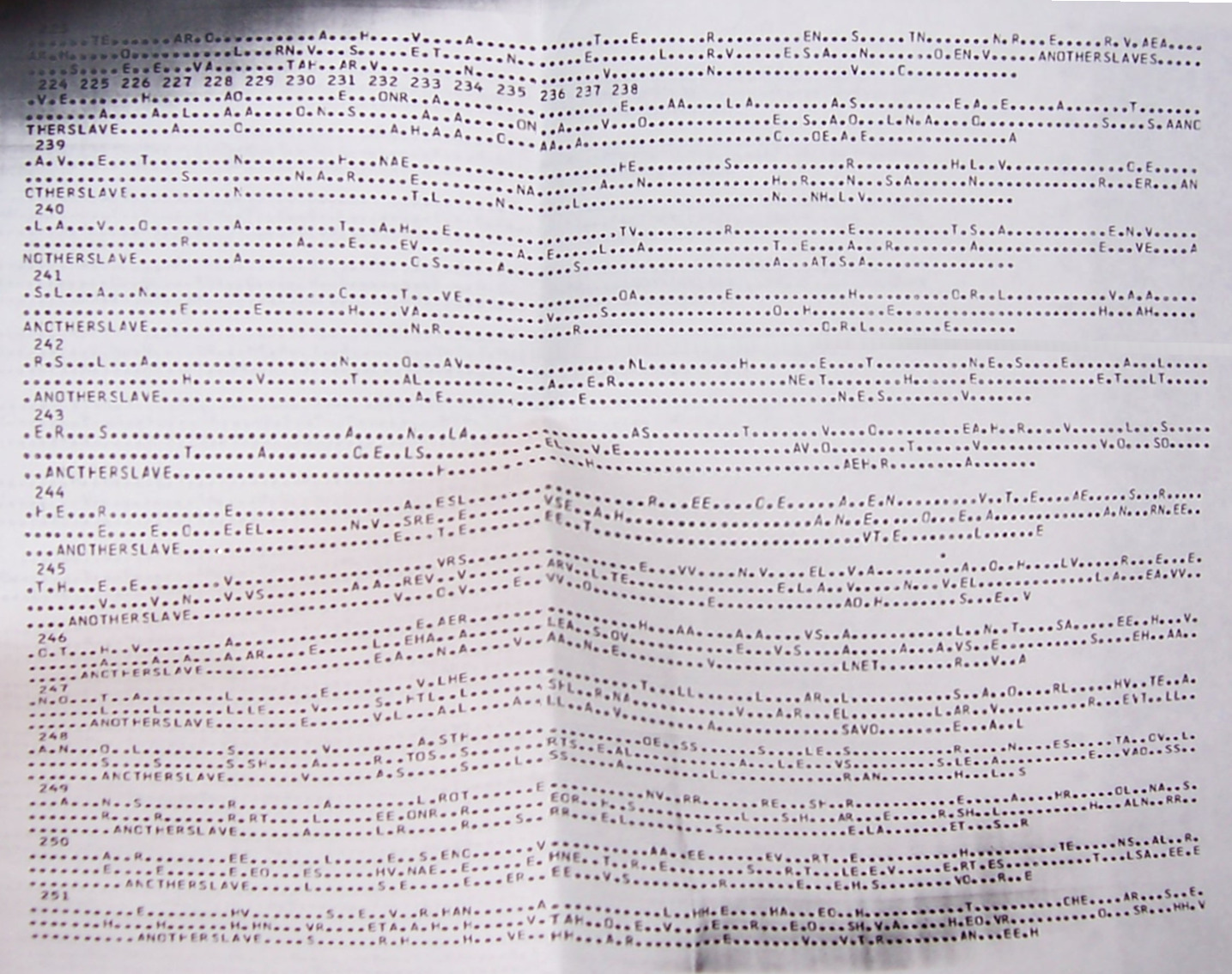} 
  \caption{An alleged attempt by NSA to solve \textit{Z340} by sliding the crib ``ANOTHERSLAVE'' to multiple positions \cite{voigtnsa}.}
  \label{fig:NSA_z340}
\end{figure}
\subsection*{\textit{Z32}: Letter and cipher}\label{sec:z32_1}
\begin{figure}[H]
  \centering
  \includegraphics[width=0.8\columnwidth]{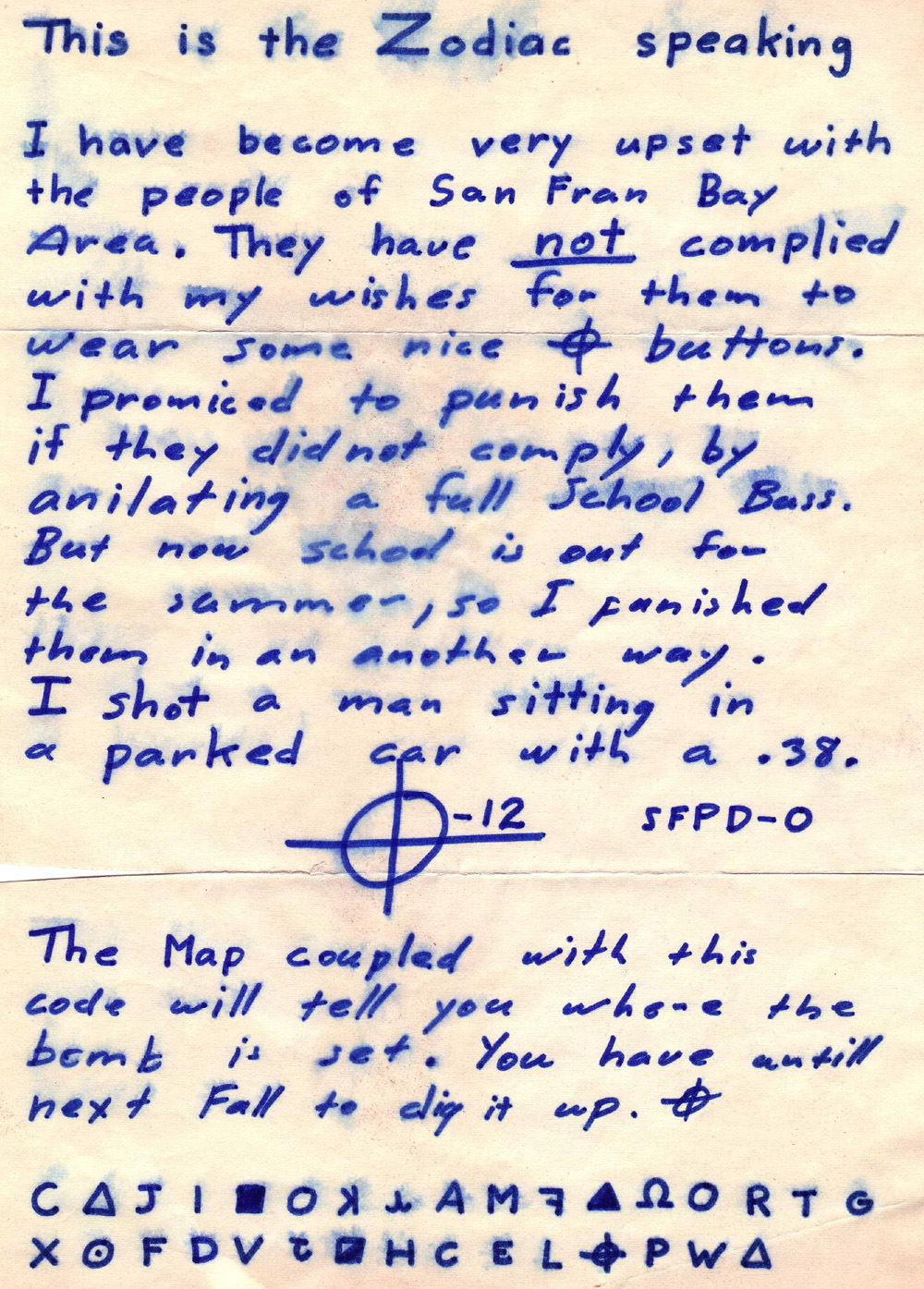} 
  \caption{June 26, 1970 Button Letter and Cipher (\textit{Z32}) \cite{voigt32}}
  \label{fig:z32_1}
\end{figure}
\subsection*{\textit{Z32}: Map}\label{sec:z32_2}
\begin{figure}[H]
  \centering
  \includegraphics[width=0.8\columnwidth]{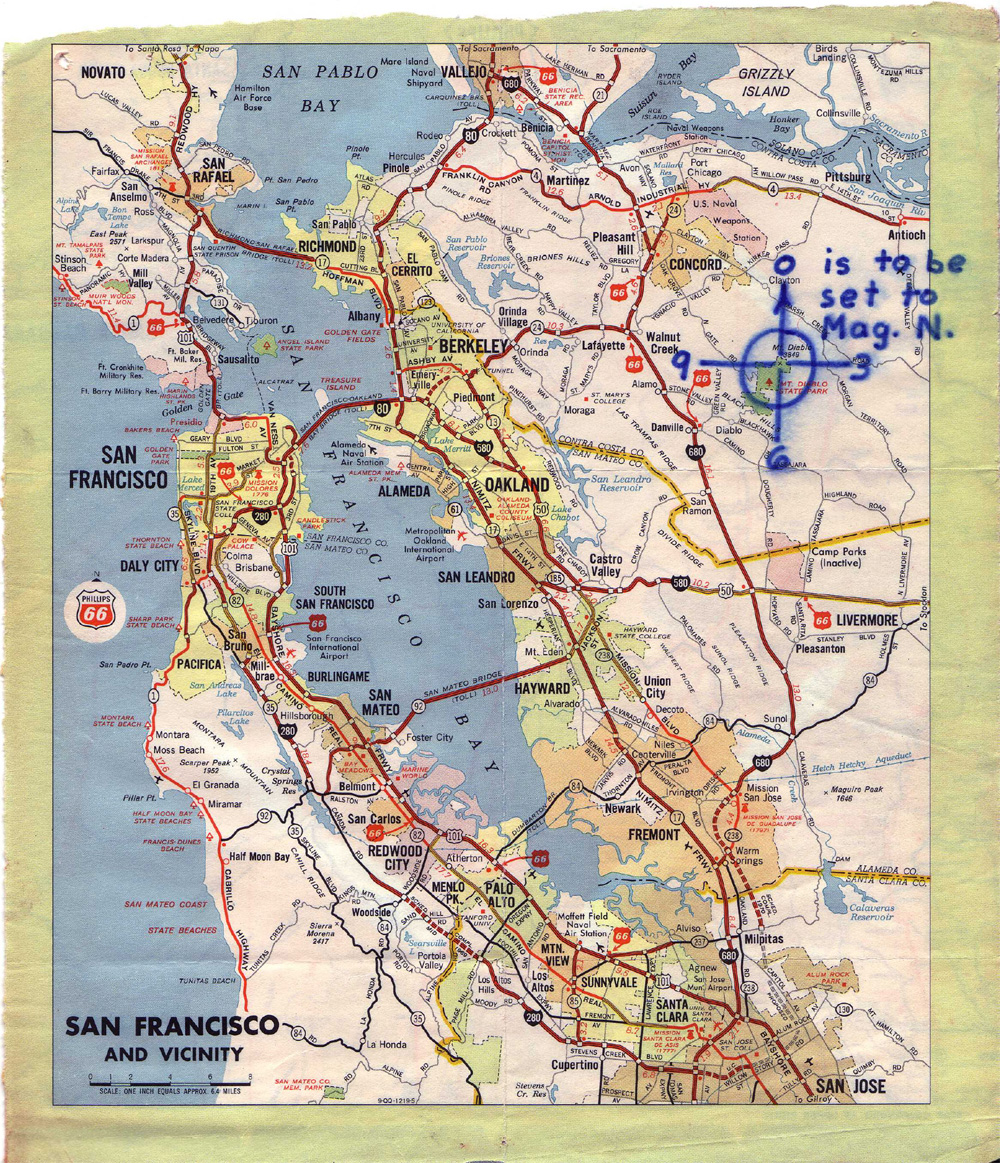} 
  \caption{June 26, 1970 Button Letter Map \cite{voigt32}}
  \label{fig:z32_2}
\end{figure}

\section{References}
\raggedright
\printbibliography[type=book,title={Books}]
\printbibliography[keyword={papers},title={Papers}]
\printbibliography[keyword={news},title={News/Periodicals}]
\printbibliography[keyword={videos},title={Videos}]
\printbibliography[keyword={software},title={Software}]
\printbibliography[keyword={forums},title={Forums}]
\printbibliography[keyword={web},title={Web}]
\justifying

\end{document}